# Efficient Solution Algorithms for Factored MDPs

**Carlos Guestrin** GUESTRIN@CS.STANFORD.EDU
*Computer Science Dept., Stanford University*

**Daphne Koller** KOLLER@CS.STANFORD.EDU
*Computer Science Dept., Stanford University*

**Ronald Parr** PARR@CS.DUKE.EDU
*Computer Science Dept., Duke University*

**Shobha Venkataraman** SHOBHA@CS.CMU.EDU
*Computer Science Dept., Carnegie Mellon University*

## Abstract

This paper addresses the problem of planning under uncertainty in large Markov Decision Processes (MDPs). *Factored MDPs* represent a complex state space using state variables and the transition model using a dynamic Bayesian network. This representation often allows an exponential reduction in the representation size of structured MDPs, but the complexity of exact solution algorithms for such MDPs can grow exponentially in the representation size. In this paper, we present two approximate solution algorithms that exploit structure in factored MDPs. Both use an approximate value function represented as a linear combination of basis functions, where each basis function involves only a small subset of the domain variables. A key contribution of this paper is that it shows how the basic operations of both algorithms can be performed efficiently in closed form, by exploiting both *additive* and *context-specific* structure in a factored MDP. A central element of our algorithms is a novel linear program decomposition technique, analogous to variable elimination in Bayesian networks, which reduces an exponentially large LP to a provably equivalent, polynomial-sized one. One algorithm uses approximate linear programming, and the second approximate dynamic programming. Our dynamic programming algorithm is novel in that it uses an approximation based on max-norm, a technique that more directly minimizes the terms that appear in error bounds for approximate MDP algorithms. We provide experimental results on problems with over $10^{40}$ states, demonstrating a promising indication of the scalability of our approach, and compare our algorithm to an existing state-of-the-art approach, showing, in some problems, exponential gains in computation time.

## 1. Introduction

Over the last few years, *Markov Decision Processes (MDPs)* have been used as the basic semantics for optimal planning for decision theoretic agents in stochastic environments. In the MDP framework, the system is modeled via a set of states which evolve stochastically. The main problem with this representation is that, in virtually any real-life domain, the state space is quite large. However, many large MDPs have significant internal structure, and can be modeled compactly if the structure is exploited in the representation.

*Factored MDPs* (Boutilier, Dearden, & Goldszmidt, 2000) are one approach to representing large, structured MDPs compactly. In this framework, a state is implicitly described by an assignment to some set of *state variables*. A *dynamic Bayesian network (DBN)* (Dean & Kanazawa, 1989) can then allow a compact representation of the transition model, by exploiting the fact that the transition of a variable often depends only on a small number of other variables. Furthermore, the momentary rewards can often also be decomposed as a sum of rewards related to individual variables or small clusters of variables.

There are two main types of structure that can simultaneously be exploited in factored MDPs: additive and context-specific structure. *Additive* structure captures the fact that typical large-scale systems can often be decomposed into a combination of locally interacting components. For example, consider the management of a large factory with many production cells. Of course, in the long run, if a cell positioned early in the production line generates faulty parts, then the whole factory may be affected. However, the quality of the parts a cell generates depends *directly* only on the state of this cell and the quality of the parts it receives from neighboring cells. Such additive structure can also be present in the reward function. For example, the cost of running the factory depends, among other things, on the *sum* of the costs of maintaining each local cell.

*Context-specific* structure encodes a different type of locality of influence: Although a part of a large system may, in general, be influenced by the state of every other part of this system, at any given point in time only a small number of parts may influence it directly. In our factory example, a cell responsible for anodization may receive parts directly from any other cell in the factory. However, a work order for a cylindrical part may restrict this dependency only to cells that have a lathe. Thus, *in the context* of producing cylindrical parts, the quality of the anodized parts depends directly *only* on the state of cells with a lathe.

Even when a large MDP can be represented compactly, for example, by using a factored representation, solving it exactly may still be intractable: Typical exact MDP solution algorithms require the manipulation of a value function, whose representation is linear in the number of states, which is exponential in the number of state variables. One approach is to approximate the solution using an approximate value function with a compact representation. A common choice is the use of *linear* value functions as an approximation — value functions that are a linear combination of potentially non-linear basis functions (Bellman, Kalaba, & Kotkin, 1963; Sutton, 1988; Tsitsiklis & Van Roy, 1996b). Our work builds on the ideas of Koller and Parr (1999, 2000), by using *factored (linear) value functions*, where each basis function is restricted to some small subset of the domain variables.

This paper presents two new algorithms for computing linear value function approximations for factored MDPs: one that uses approximate dynamic programming and another that uses approximate linear programming. Both algorithms are based on the use of factored linear value functions, a highly expressive function approximation method. This representation allows the algorithms to take advantage of *both* additive and context-specific structure, in order to produce high-quality approximate solutions very efficiently. The capability to exploit both types of structure distinguishes these algorithms differ from earlier approaches (Boutilier *et al.*, 2000), which only exploit context-specific structure. We provide a more detailed discussion of the differences in Section 10.

We show that, for a factored MDP and factored value functions, various critical operations for our planning algorithms can be implemented in closed form without necessarily enumerating the entire state space. In particular, both our new algorithms build upon a novel linear programming decomposition technique. This technique reduces structured LPs with exponentially many constraints to equivalent, polynomially-sized ones. This decomposition follows a procedure analogous to variable elimination that applies both to additively structured value functions (Bertele & Brioschi, 1972) and to value functions that also exploit context-specific structure (Zhang & Poole, 1999). Using these basic operations, our planning algorithms can be implemented efficiently, even though the size of the state space grows exponentially in the number of variables.

Our first method is based on the approximate linear programming algorithm (Schweitzer & Seidmann, 1985). This algorithm generates a linear, approximate value function by solving a single linear program. Unfortunately, the number of constraints in the LP proposed by Schweitzer and Seidmann grows exponentially in the number of variables. Using our LP decomposition technique, we exploit structure in factored MDPs to represent exactly the same optimization problem with exponentially fewer constraints.

In terms of approximate dynamic programming, this paper makes a twofold contribution. First, we provide a new approach for approximately solving MDPs using a linear value function. Previous approaches to linear function approximation typically have utilized a least squares ($\mathcal{L}_2$-norm) approximation to the value function. Least squares approximations are incompatible with most convergence analyses for MDPs, which are based on max-norm. We provide the first MDP solution algorithms — both value iteration and policy iteration — that use a linear max-norm projection to approximate the value function, thereby directly optimizing the quantity that appears in our provided error bounds. Second, we show how to exploit the structure of the problem to apply this technique to factored MDPs, by again leveraging on our LP decomposition technique.

Although approximate dynamic programming currently possesses stronger theoretical guarantees, our experimental results suggest that approximate linear programming is a good alternative. Whereas the former tends to generate better policies for the same set of basis functions, due to the simplicity and computational advantages of approximate linear programming, we can add more basis functions, obtaining a better policy and still requiring less computation than the approximate dynamic programming approach.

Finally, we present experimental results comparing our approach to the work of Boutilier *et al.* (2000), illustrating some of the tradeoffs between the two methods. In particular, for problems with significant context-specific structure *in the value function*, their approach can be faster due to their efficient handling of their value function representation. However, there are cases with significant context-specific structure in the problem, rather than in the value function, in which their algorithm requires an exponentially large value function representation. In such classes of problems, we demonstrate that by using a value function that exploits both additive and context-specific structure, our algorithm can obtain a polynomial-time near-optimal approximation of the true value function.

This paper starts with a presentation of factored MDPs and approximate solution algorithms for MDPs. In Section 4, we describe the basic operations used in our algorithms, including our LP decomposition technique. In Section 5, we present the first of our two algorithms: the approximate linear programming algorithm for factored MDPs. The second algorithm, approximate policy iteration with max-norm projection, is presented in Section 6. Section 7 describes an approach for efficiently computing bounds on policy quality based on the Bellman error. Section 8 shows how to extend our methods to deal with context-specific structure. Our paper concludes with an empirical evaluation in Section 9 and a discussion of related work in Section 10.

This paper is a greatly expanded version of work that was published before in Guestrin *et al.* (2001a), and some of the work presented in Guestrin *et al.* (2001b, 2002).

## 2. Factored Markov Decision Processes

A Markov decision process (MDP) is a mathematical framework for sequential decision problems in stochastic domains. It thus provides an underlying semantics for the task of planning under uncertainty. We begin with a concise overview of the MDP framework, and then describe the representation of *factored MDPs*.

### 2.1 Markov Decision Processes

We briefly review the MDP framework, referring the reader to the books by Bertsekas and Tsitsiklis (1996) or Puterman (1994) for a more in-depth review. A *Markov Decision Process (MDP)* is defined as a 4-tuple $(\mathbf{X}, A, R, P)$ where: $\mathbf{X}$ is a finite set of $|\mathbf{X}| = N$ states; $A$ is a finite set of actions; $R$ is a *reward function* $R : \mathbf{X} \times A \mapsto \mathbb{R}$, such that $R(\mathbf{x}, a)$ represents the reward obtained by the agent in state $\mathbf{x}$ after taking action $a$; and $P$ is a *Markovian transition model* where $P(\mathbf{x}' \mid \mathbf{x}, a)$ represents the probability of going from state $\mathbf{x}$ to state $\mathbf{x}'$ with action $a$. We assume that the rewards are bounded, that is, there exists $R_{max}$ such that $R_{max} \geq |R(\mathbf{x}, a)|, \forall \mathbf{x}, a$.

**Example 2.1** *Consider the problem of optimizing the behavior of a system administrator (SysAdmin) maintaining a network of $m$ computers. In this network, each machine is connected to some subset of the other machines. Various possible network topologies can be defined in this manner (see Figure 1 for some examples). In one simple network, we might connect the machines in a ring, with machine $i$ connected to machines $i+1$ and $i-1$. (In this example, we assume addition and subtraction are performed modulo $m$.)*

*Each machine is associated with a binary random variable $X_i$, representing whether it is working or has failed. At every time step, the SysAdmin receives a certain amount of money (reward) for each working machine. The job of the SysAdmin is to decide which machine to reboot; thus, there are $m+1$ possible actions at each time step: reboot one of the $m$ machines or do nothing (only one machine can be rebooted per time step). If a machine is rebooted, it will be working with high probability at the next time step. Every machine has a small probability of failing at each time step. However, if a neighboring machine fails, this probability increases dramatically. These failure probabilities define the transition model $P(\mathbf{x}' \mid \mathbf{x}, a)$, where $\mathbf{x}$ is a particular assignment describing which machines are working or have failed in the current time step, $a$ is the SysAdmin's choice of machine to reboot and $\mathbf{x}'$ is the resulting state in the next time step.* $\square$

We assume that the MDP has an infinite horizon and that future rewards are discounted exponentially with a discount factor $\gamma \in [0, 1)$. A *stationary policy* $\pi$ for an MDP is a mapping $\pi : \mathbf{X} \mapsto A$, where $\pi(\mathbf{x})$ is the action the agent takes at state $\mathbf{x}$. In the computer network problem, for each possible configuration of working and failing machines, the policy would tell the SysAdmin which machine to reboot. Each policy is associated with a *value function* $\mathcal{V}_\pi \in \mathbb{R}^N$, where $\mathcal{V}_\pi(\mathbf{x})$ is the discounted cumulative value that the agent gets if it starts at state $\mathbf{x}$ and follows policy $\pi$. More precisely, the value $\mathcal{V}_\pi$ of a state $\mathbf{x}$ under

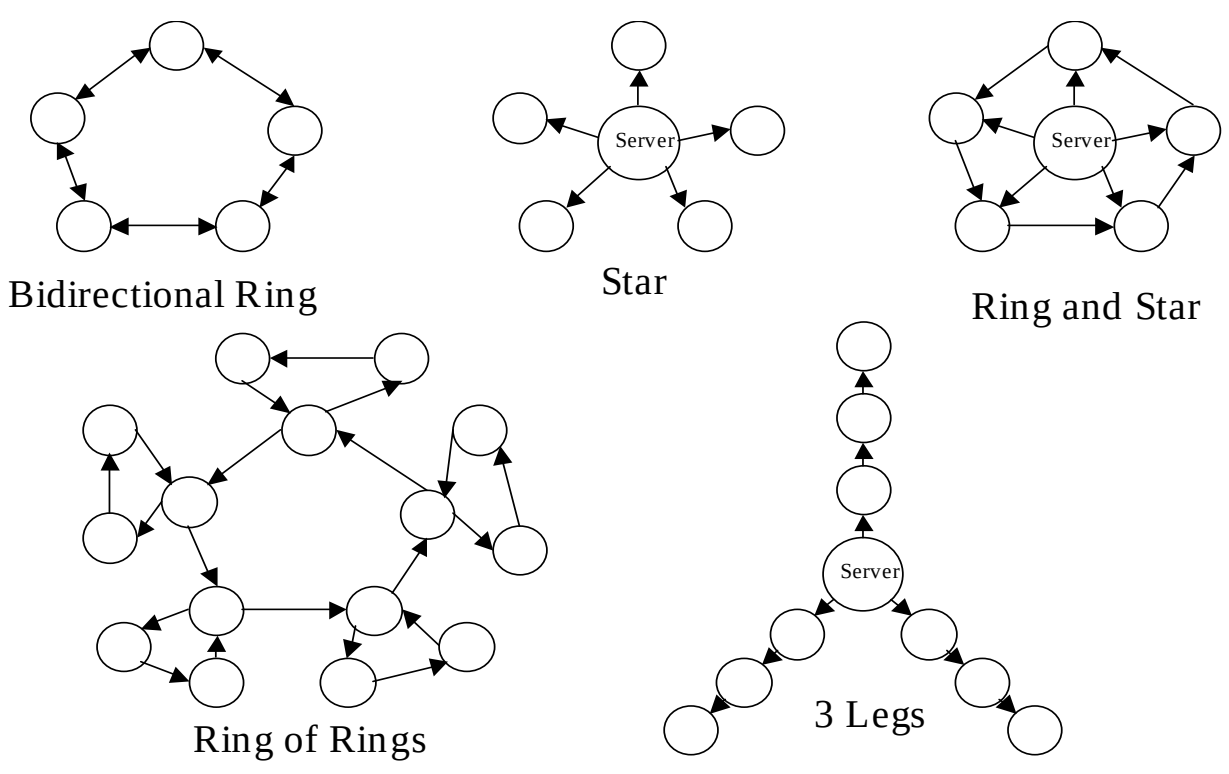

Figure 1: Network topologies tested; the status of a machine is influence by the status of its parent in the network.

policy $\pi$ is given by:

$$\mathcal{V}_\pi(\mathbf{x}) = E_\pi\left[\sum_{t=0}^{\infty}\gamma^t R\left(\mathbf{X}^{(t)}, \pi(\mathbf{X}^{(t)})\right)\middle|\, \mathbf{X}^{(0)} = \mathbf{x}\right],$$

where $\mathbf{X}^{(t)}$ is a random variable representing the state of the system after $t$ steps. In our running example, the value function represents how much money the SysAdmin expects to collect if she starts acting according to $\pi$ when the network is at state $\mathbf{x}$. The value function for a fixed policy is the fixed point of a set of linear equations that define the value of a state in terms of the value of its possible successor states. More formally, we define:

**Definition 2.2** *The* DP operator*, $\mathcal{T}_\pi$, for a stationary policy $\pi$ is:*

$$\mathcal{T}_\pi\mathcal{V}(\mathbf{x}) = R(\mathbf{x}, \pi(\mathbf{x})) + \gamma\sum_{\mathbf{x}'} P(\mathbf{x}' \mid \mathbf{x}, \pi(\mathbf{x}))\mathcal{V}(\mathbf{x}').$$

*The value function of policy $\pi$, $\mathcal{V}_\pi$, is the fixed point of the $\mathcal{T}_\pi$ operator: $\mathcal{V}_\pi = \mathcal{T}_\pi\mathcal{V}_\pi$.* □

The optimal value function $\mathcal{V}^*$ describes the optimal value the agent can achieve for each starting state. $\mathcal{V}^*$ is also defined by a set of non-linear equations. In this case, the value of a state must be the maximal expected value achievable by any policy starting at that state. More precisely, we define:

**Definition 2.3** *The* Bellman operator*, $\mathcal{T}^*$, is:*

$$\mathcal{T}^*\mathcal{V}(\mathbf{x}) = \max_a[R(\mathbf{x}, a) + \gamma\sum_{\mathbf{x}'} P(\mathbf{x}' \mid \mathbf{x}, a)\mathcal{V}(\mathbf{x}')].$$

*The optimal value function $\mathcal{V}^*$ is the fixed point of $\mathcal{T}^*$: $\mathcal{V}^* = \mathcal{T}^*\mathcal{V}^*$.* □

For any value function $\mathcal{V}$, we can define the policy obtained by acting greedily relative to $\mathcal{V}$. In other words, at each state, the agent takes the action that maximizes the one-step utility, assuming that $\mathcal{V}$ represents our long-term utility achieved at the next state. More precisely, we define:

$$Greedy(\mathcal{V})(\mathbf{x}) = \arg\max_a [R(\mathbf{x}, a) + \gamma \sum_{\mathbf{x}'} P(\mathbf{x}' \mid \mathbf{x}, a)\mathcal{V}(\mathbf{x}')]. \quad (1)$$

The greedy policy relative to the optimal value function $\mathcal{V}^*$ is the optimal policy $\pi^* = Greedy(\mathcal{V}^*)$.

## 2.2 Factored MDPs

*Factored MDPs* are a representation language that allows us to exploit problem structure to represent exponentially large MDPs very compactly. The idea of representing a large MDP using a factored model was first proposed by Boutilier *et al.* (1995).

In a factored MDP, the set of states is described via a set of random variables $\mathbf{X} = \{X_1, \dots, X_n\}$, where each $X_i$ takes on values in some finite domain $\mathrm{Dom}(X_i)$. A state $\mathbf{x}$ defines a value $x_i \in \mathrm{Dom}(X_i)$ for each variable $X_i$. In general, we use upper case letters (*e.g.*, $X$) to denote random variables, and lower case (*e.g.*, $x$) to denote their values. We use boldface to denote vectors of variables (*e.g.*, $\mathbf{X}$) or their values ($\mathbf{x}$). For an instantiation $\mathbf{y} \in \mathrm{Dom}(\mathbf{Y})$ and a subset of these variables $\mathbf{Z} \subset \mathbf{Y}$, we use $\mathbf{y}[\mathbf{Z}]$ to denote the value of the variables $\mathbf{Z}$ in the instantiation $\mathbf{y}$.

In a factored MDP, we define a state transition model $\tau$ using a *dynamic Bayesian network (DBN)* (Dean & Kanazawa, 1989). Let $X_i$ denote the variable $X_i$ at the current time and $X'_i$, the same variable at the next step. The *transition graph* of a DBN is a two-layer directed acyclic graph $G_\tau$ whose nodes are $\{X_1, \dots, X_n, X'_1, \dots, X'_n\}$. We denote the parents of $X'_i$ in the graph by $Parents_\tau(X'_i)$. For simplicity of exposition, we assume that $Parents_\tau(X'_i) \subseteq \mathbf{X}$; thus, all arcs in the DBN are between variables in consecutive time slices. (This assumption is used for expository purposes only; intra-time slice arcs are handled by a small modification presented in Section 4.1.) Each node $X'_i$ is associated with a *conditional probability distribution (CPD)* $P_\tau(X'_i \mid Parents_\tau(X'_i))$. The transition probability $P_\tau(\mathbf{x}' \mid \mathbf{x})$ is then defined to be:

$$P_\tau(\mathbf{x}' \mid \mathbf{x}) = \prod_i P_\tau(x'_i \mid \mathbf{u}_i) \ ,$$

where $\mathbf{u}_i$ is the value in $\mathbf{x}$ of the variables in $Parents_\tau(X'_i)$.

**Example 2.4** *Consider an instance of the SysAdmin problem with four computers, labelled $M_1, \dots, M_4$, in an unidirectional ring topology as shown in Figure 2(a). Our first task in modeling this problem as a factored MDP is to define the state space $\mathbf{X}$. Each machine is associated with a binary random variable $X_i$, representing whether it is working or has failed. Thus, our state space is represented by four random variables: $\{X_1, X_2, X_3, X_4\}$. The next task is to define the transition model, represented as a DBN. The parents of the next time step variables $X'_i$ depend on the network topology. Specifically, the probability that machine i will fail at the next time step depends on whether it is working at the current time step and on the status of its direct neighbors (parents in the topology) in the network at the current time step. As shown in Figure 2(b), the parents of $X'_i$ in this example are $X_i$ and $X_{i-1}$. The CPD of $X'_i$ is such that if $X_i = false$, then $X'_i = false$ with high probability;*

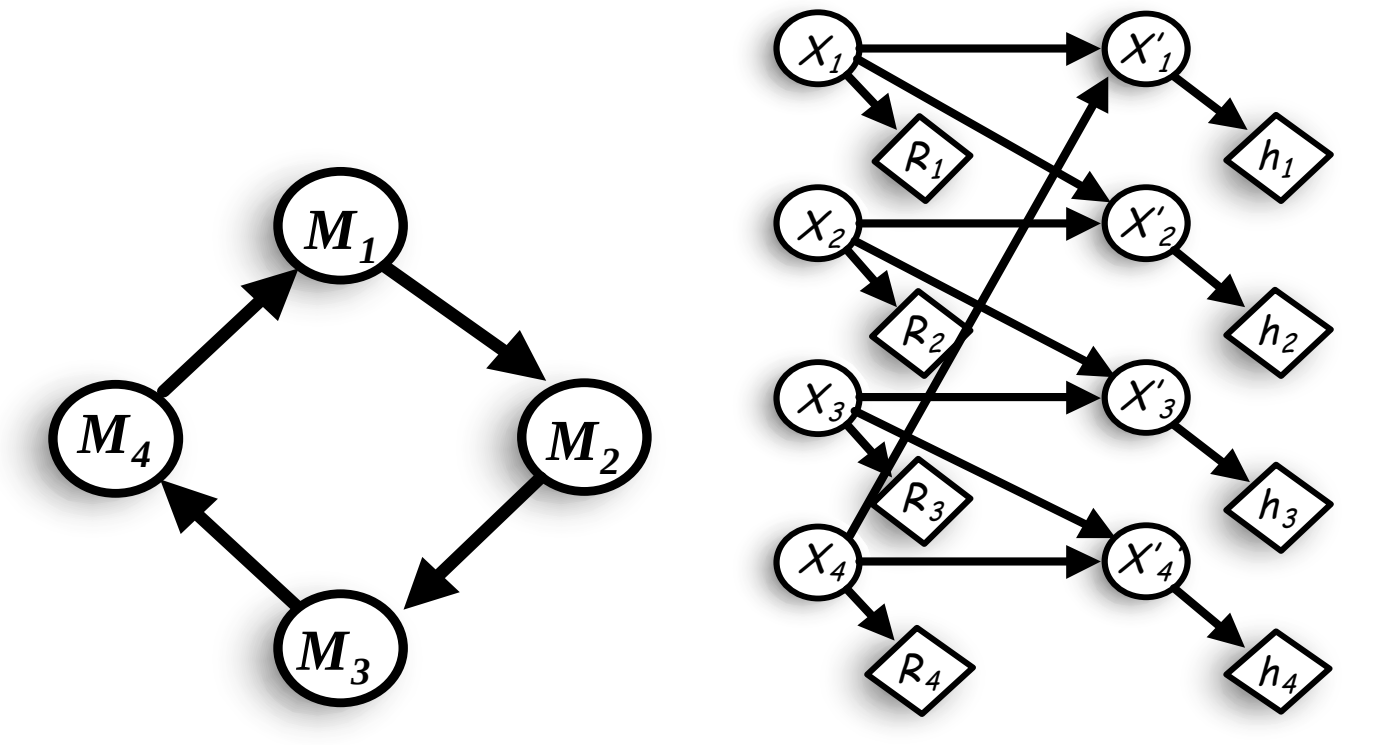

(a) (b)

$P(X_i' = t \mid X_i, X_{i-1}, A)$:

| | Action is reboot: machine $i$ | Action is reboot: other machine |
|---|---|---|
| $X_{i-1} = f \wedge X_i = f$ | 1 | 0.0238 |
| $X_{i-1} = f \wedge X_i = t$ | 1 | 0.475 |
| $X_{i-1} = t \wedge X_i = f$ | 1 | 0.0475 |
| $X_{i-1} = t \wedge X_i = t$ | 1 | 0.95 |

(c)

Figure 2: Factored MDP example: from a network topology (a) we obtain the factored MDP representation (b) with the CPDs described in (c).

*that is, failures tend to persist. If $X_i = true$, then $X_i'$ is a* noisy or *of its other parents (in the unidirectional ring topology $X_i'$ has only one other parent $X_{i-1}$); that is, a failure in any of its neighbors can independently cause machine $i$ to fail.* □

We have described how to represent factored the Markovian transition dynamics arising from an MDP as a DBN, but we have not directly addressed the representation of actions. Generally, we can define the transition dynamics of an MDP by defining a separate DBN model $\tau_a = \langle G_a, P_a \rangle$ for each action $a$.

**Example 2.5** *In our system administrator example, we have an action $a_i$ for rebooting each one of the machines, and a default action $d$ for doing nothing. The transition model described above corresponds to the "do nothing" action. The transition model for $a_i$ is different from $d$ only in the transition model for the variable $X_i'$, which is now $X_i' = true$ with probability one, regardless of the status of the neighboring machines. Figure 2(c) shows the actual CPD for $P(X_i' = Working \mid X_i, X_{i-1}, A)$, with one entry for each assignment to the state variables $X_i$ and $X_{i-1}$, and to the action $A$.* □

To fully specify an MDP, we also need to provide a compact representation of the reward function. We assume that the reward function is factored additively into a set of localized reward functions, each of which only depends on a small set of variables. In our example, we might have a reward function associated with each machine $i$, which depends on $X_i$. That is, the SysAdmin is paid on a per-machine basis: at every time step, she receives money for machine $i$ only if it is working. We can formalize this concept of localized functions:

**Definition 2.6** *A function $f$ has a* scope $\text{Scope}[f] = \mathbf{C} \subseteq \mathbf{X}$ *if $f : \text{Dom}(\mathbf{C}) \mapsto \mathbb{R}$.* □

If $f$ has scope $\mathbf{Y}$ and $\mathbf{Y} \subset \mathbf{Z}$, we use $f(\mathbf{z})$ as shorthand for $f(\mathbf{y})$ where $\mathbf{y}$ is the part of the instantiation $\mathbf{z}$ that corresponds to variables in $\mathbf{Y}$.

We can now characterize the concept of local rewards. Let $R_1^a, \ldots, R_r^a$ be a set of functions, where the scope of each $R_i^a$ is restricted to variable cluster $\mathbf{U}_i^a \subset \{X_1, \ldots, X_n\}$. The reward for taking action $a$ at state $\mathbf{x}$ is defined to be $R^a(\mathbf{x}) = \sum_{i=1}^r R_i^a(\mathbf{U}_i^a) \in \mathbb{R}$. In our example, we have a reward function $R_i$ associated with each machine $i$, which depends only $X_i$, and does not depend on the action choice. These local rewards are represented by the diamonds in Figure 2(b), in the usual notation for influence diagrams (Howard & Matheson, 1984).

## 3. Approximate Solution Algorithms

There are several algorithms to compute the optimal policy in an MDP. The three most commonly used are value iteration, policy iteration, and linear programming. A key component in all three algorithms is the computation of value functions, as defined in Section 2.1. Recall that a value function defines a value for each state $\mathbf{x}$ in the state space. With an explicit representation of the value function as a vector of values for the different states, the solution algorithms all can be implemented as a series of simple algebraic steps. Thus, in this case, all three can be implemented very efficiently.

Unfortunately, in the case of factored MDPs, the state space is exponential in the number of variables in the domain. In the *SysAdmin* problem, for example, the state $\mathbf{x}$ of the system is an assignment describing which machines are working or have failed; that is, a state $\mathbf{x}$ is an assignment to each random variable $X_i$. Thus, the number of states is exponential in the number $m$ of machines in the network ($|\mathbf{X}| = N = 2^m$). Hence, even representing an explicit value function in problems with more than about ten machines is infeasible. One might be tempted to believe that factored transition dynamics and rewards would result in a factored value function, which can thereby be represented compactly. Unfortunately, even in trivial factored MDPs, there is no guarantee that structure in the model is preserved in the value function (Koller & Parr, 1999).

In this section, we discuss the use of an *approximate* value function, that admits a compact representation. We also describe approximate versions of these exact algorithms, that use approximate value functions. Our description in this section is somewhat abstract, and does not specify how the basic operations required by the algorithms can be performed explicitly. In later sections, we elaborate on these issues, and describe the algorithms in detail. For brevity, we choose to focus on policy iteration and linear programming; our techniques easily extend to value iteration.

### 3.1 Linear Value Functions

A very popular choice for approximating value functions is by using *linear regression*, as first proposed by Bellman *et al.* (1963). Here, we define our space of allowable value functions $\mathcal{V} \in \mathcal{H} \subseteq \mathbb{R}^N$ via a set of *basis functions*:

**Definition 3.1** *A* linear value function *over a set of basis functions* $H = \{h_1, \ldots, h_k\}$ *is a function* $\mathcal{V}$ *that can be written as* $\mathcal{V}(\mathbf{x}) = \sum_{j=1}^k w_j \; h_j(\mathbf{x})$ *for some coefficients* $\mathbf{w} = (w_1, \ldots, w_k)'$. $\square$

We can now define $\mathcal{H}$ to be the linear subspace of $\mathbb{R}^N$ spanned by the basis functions $H$. It is useful to define an $N \times k$ matrix H whose columns are the $k$ basis functions viewed as vectors. In a more compact notation, our approximate value function is then represented by $\mathbf{Hw}$.

The expressive power of this linear representation is equivalent, for example, to that of a single layer neural network with features corresponding to the basis functions defining $\mathcal{H}$. Once the features are defined, we must optimize the coefficients $\mathbf{w}$ in order to obtain a good approximation for the true value function. We can view this approach as separating the problem of defining a reasonable space of features and the induced space $\mathcal{H}$, from the problem of searching within the space. The former problem is typically the purview of domain experts, while the latter is the focus of analysis and algorithmic design. Clearly, feature selection is an important issue for essentially all areas of learning and approximation. We offer some simple methods for selecting good features for MDPs in Section 11, but it is not our goal to address this large and important topic in this paper.

Once we have a chosen a linear value function representation and a set of basis functions, the problem becomes one of finding values for the weights $\mathbf{w}$ such that $\mathbf{Hw}$ will yield a good approximation of the true value function. In this paper, we consider two such approaches: approximate dynamic programming using policy iteration and approximate linear programming. In this section, we present these two approaches. In Section 4, we show how we can exploit problem structure to transform these approaches into practical algorithms that can deal with exponentially large state spaces.

## 3.2 Policy Iteration

### 3.2.1 The Exact Algorithm

The exact policy iteration algorithm iterates over policies, producing an improved policy at each iteration. Starting with some initial policy $\pi^{(0)}$, each iteration consists of two phases. *Value determination* computes, for a policy $\pi^{(t)}$, the value function $\mathcal{V}_{\pi^{(t)}}$, by finding the fixed point of the equation $\mathcal{T}_{\pi^{(t)}}\mathcal{V}_{\pi^{(t)}} = \mathcal{V}_{\pi^{(t)}}$, that is, the unique solution to the set of linear equations:

$$\mathcal{V}_{\pi^{(t)}}(\mathbf{x}) = R(\mathbf{x}, \pi^{(t)}(\mathbf{x})) + \gamma \sum_{\mathbf{x}'} P(\mathbf{x}' \mid \mathbf{x}, \pi^{(t)}(\mathbf{x}))\mathcal{V}_{\pi^{(t)}}(\mathbf{x}'), \forall \mathbf{x}.$$

The *policy improvement* step defines the next policy as

$$\pi^{(t+1)} = \mathit{Greedy}(\mathcal{V}_{\pi^{(t)}}).$$

It can be shown that this process converges to the optimal policy (Bertsekas & Tsitsiklis, 1996). Furthermore, in practice, the convergence to the optimal policy is often very quick.

### 3.2.2 Approximate Policy Iteration

The steps in the policy iteration algorithm require a manipulation of both value functions and policies, both of which often cannot be represented explicitly in large MDPs. To define a version of the policy iteration algorithm that uses approximate value functions, we use the following basic idea: We restrict the algorithm to using only value functions within the provided $\mathcal{H}$; whenever the algorithm takes a step that results in a value function $\mathcal{V}$ that is outside this space, we *project* the result back into the space by finding the value function within the space which is closest to $\mathcal{V}$. More precisely:

**Definition 3.2** *A* projection operator $\Pi$ *is a mapping* $\Pi : \mathbb{R}^N \to \mathcal{H}$. $\Pi$ *is said to be a* projection w.r.t. a norm $\|\cdot\|$ *if* $\Pi\mathcal{V} = \mathrm{H}\mathbf{w}^*$ *such that* $\mathbf{w}^* \in \arg\min_{\mathbf{w}} \|\mathrm{H}\mathbf{w} - \mathcal{V}\|$. □

That is, $\Pi\mathcal{V}$ is the linear combination of the basis functions, that is closest to $\mathcal{V}$ with respect to the chosen norm.

Our approximate policy iteration algorithm performs the policy improvement step exactly. In the value determination step, the value function — the value of acting according to the current policy $\pi^{(t)}$ — is approximated through a linear combination of basis functions.

We now consider the problem of value determination for a policy $\pi^{(t)}$. At this point, it is useful to introduce some notation: Although the rewards are a function of the state and action choice, once the policy is fixed, the rewards become a function of the state only, which we denote as $R_{\pi^{(t)}}$, where $R_{\pi^{(t)}}(\mathbf{x}) = R(\mathbf{x}, \pi^{(t)}(\mathbf{x}))$. Similarly, for the transition model: $P_{\pi^{(t)}}(\mathbf{x}' \mid \mathbf{x}) = P(\mathbf{x}' \mid \mathbf{x}, \pi^{(t)}(\mathbf{x}))$. We can now rewrite the value determination step in terms of matrices and vectors. If we view $\mathcal{V}_{\pi^{(t)}}$ and $R_{\pi^{(t)}}$ as $N$-vectors, and $P_{\pi^{(t)}}$ as an $N \times N$ matrix, we have the equations:

$$\mathcal{V}_{\pi^{(t)}} = R_{\pi^{(t)}} + \gamma P_{\pi^{(t)}} \mathcal{V}_{\pi^{(t)}}.$$

This is a system of linear equations with one equation for each state, which can only be solved exactly for relatively small $N$. Our goal is to provide an approximate solution, within $\mathcal{H}$. More precisely, we want to find:

$$\begin{aligned} \mathbf{w}^{(t)} &= \arg\min_{\mathbf{w}} \left\|\mathrm{H}\mathbf{w} - (R_{\pi^{(t)}} + \gamma P_{\pi^{(t)}} \mathrm{H}\mathbf{w})\right\|; \\ &= \arg\min_{\mathbf{w}} \left\|\left(\mathrm{H} - \gamma P_{\pi^{(t)}} \mathrm{H}\right) \mathbf{w}^{(t)} - R_{\pi^{(t)}}\right\|. \end{aligned}$$

Thus, our approximate policy iteration alternates between two steps:

$$\mathbf{w}^{(t)} = \arg\min_{\mathbf{w}} \left\|\mathrm{H}\mathbf{w} - (R_{\pi^{(t)}} + \gamma P_{\pi^{(t)}} \mathrm{H}\mathbf{w})\right\|; \quad (2)$$

$$\pi^{(t+1)} = \mathit{Greedy}(\mathrm{H}\mathbf{w}^{(t)}). \quad (3)$$

#### 3.2.3 Max-norm Projection

An approach along the lines described above has been used in various papers, with several recent theoretical and algorithmic results (Schweitzer & Seidmann, 1985; Tsitsiklis & Van Roy, 1996b; Van Roy, 1998; Koller & Parr, 1999, 2000). However, these approaches suffer from a problem that we might call "norm incompatibility." When computing the projection, they utilize the standard Euclidean projection operator with respect to the $\mathcal{L}_2$ norm or a *weighted* $\mathcal{L}_2$ norm.[1] On the other hand, most of the convergence and error analyses for MDP algorithms utilize max-norm ($\mathcal{L}_\infty$). This incompatibility has made it difficult to provide error guarantees.

We can tie the projection operator more closely to the error bounds through the use of a projection operator in $\mathcal{L}_\infty$ norm. The problem of minimizing the $\mathcal{L}_\infty$ norm has been studied in the optimization literature as the problem of finding the Chebyshev solution[2] to

1. Weighted $\mathcal{L}_2$ norm projections are stable and have meaningful error bounds when the weights correspond to the stationary distribution of a fixed policy under evaluation (value determination) (Van Roy, 1998), but they are not stable when combined with $\mathcal{T}^*$. Averagers (Gordon, 1995) are stable and non-expansive in $\mathcal{L}_\infty$, but require that the mixture weights be determined *a priori*. Thus, they do not, in general, minimize $\mathcal{L}_\infty$ error.
2. The Chebyshev norm is also referred to as max, supremum and $\mathcal{L}_\infty$ norms and the minimax solution.

an overdetermined linear system of equations (Cheney, 1982). The problem is defined as finding $\mathbf{w}^*$ such that:

$$\mathbf{w}^* \in \arg\min_{\mathbf{w}} \|C\mathbf{w} - \mathbf{b}\|_\infty . \tag{4}$$

We use an algorithm due to Stiefel (1960), that solves this problem by linear programming:

$$\begin{array}{ll} \text{Variables:} & w_1, \ldots, w_k, \phi \ ; \\ \text{Minimize:} & \phi \ ; \\ \text{Subject to:} & \phi \geq \sum_{j=1}^{k} c_{ij} w_j - b_i \quad \text{and} \\ & \phi \geq b_i - \sum_{j=1}^{k} c_{ij} w_j, \quad i = 1 \ldots N. \end{array} \tag{5}$$

The constraints in this linear program imply that $\phi \geq \left|\sum_{j=1}^{k} c_{ij} w_j - b_i\right|$ for each $i$, or equivalently, that $\phi \geq \|C\mathbf{w} - \mathbf{b}\|_\infty$. The objective of the LP is to minimize $\phi$. Thus, at the solution $(\mathbf{w}^*, \phi^*)$ of this linear program, $\mathbf{w}^*$ is the solution of Equation (4) and $\phi$ is the $\mathcal{L}_\infty$ projection error.

We can use the $\mathcal{L}_\infty$ projection in the context of the approximate policy iteration in the obvious way. When implementing the projection operation of Equation (2), we can use the $\mathcal{L}_\infty$ projection (as in Equation (4)), where $C = (\mathrm{H} - \gamma P_{\pi^{(t)}} \mathrm{H})$ and $\mathbf{b} = R_{\pi^{(t)}}$. This minimization can be solved using the linear program of (5).

A key point is that this LP only has $k+1$ variables. However, there are $2N$ constraints, which makes it impractical for large state spaces. In the *SysAdmin* problem, for example, the number of constraints in this LP is exponential in the number of machines in the network (a total of $2 \cdot 2^m$ constraints for $m$ machines). In Section 4, we show that, in *factored* MDPs with linear value functions, all the $2N$ constraints can be represented efficiently, leading to a tractable algorithm.

#### 3.2.4 Error Analysis

We motivated our use of the max-norm projection within the approximate policy iteration algorithm via its compatibility with standard error analysis techniques for MDP algorithms. We now provide a careful analysis of the impact of the $\mathcal{L}_\infty$ error introduced by the projection step. The analysis provides motivation for the use of a projection step that directly minimizes this quantity. We acknowledge, however, that the main impact of this analysis is motivational. In practice, we cannot provide *a priori* guarantees that an $\mathcal{L}_\infty$ projection will outperform other methods.

Our goal is to analyze approximate policy iteration in terms of the amount of error introduced at each step by the projection operation. If the error is zero, then we are performing exact value determination, and no error should accrue. If the error is small, we should get an approximation that is accurate. This result follows from the analysis below. More precisely, we define the projection error as the error resulting from the approximate value determination step:

$$\beta^{(t)} = \left\| \mathrm{H}\mathbf{w}^{(t)} - \left( R_{\pi^{(t)}} + \gamma P_{\pi^{(t)}} \mathrm{H}\mathbf{w}^{(t)} \right) \right\|_\infty .$$

Note that, by using our max-norm projection, we are finding the set of weights $\mathbf{w}^{(t)}$ that exactly minimizes the one-step projection error $\beta^{(t)}$. That is, we are choosing the best possible weights with respect to this error measure. Furthermore, this is exactly the error measure that is going to appear in the bounds of our theorem. Thus, we can now make the bounds for each step as tight as possible.

We first show that the projection error accrued in each step is bounded:

**Lemma 3.3** *The value determination error is bounded: There exists a constant $\beta_P \leq R_{max}$ such that $\beta_P \geq \beta^{(t)}$ for all iterations $t$ of the algorithm.*

**Proof:** See Appendix A.1. □

Due to the contraction property of the Bellman operator, the overall accumulated error is a decaying average of the projection error incurred throughout all iterations:

**Definition 3.4** *The* discounted value determination error *at iteration $t$ is defined as: $\overline{\beta}^{(t)} = \beta^{(t)} + \gamma\overline{\beta}^{(t-1)}$; $\overline{\beta}^{(0)} = 0$.* □

Lemma 3.3 implies that the accumulated error remains bounded in approximate policy iteration: $\overline{\beta}^{(t)} \leq \frac{\beta_P(1-\gamma^t)}{1-\gamma}$. We can now bound the loss incurred when acting according to the policy generated by our approximate policy iteration algorithm, as opposed to the optimal policy:

**Theorem 3.5** *In the approximate policy iteration algorithm, let $\pi^{(t)}$ be the policy generated at iteration $t$. Furthermore, let $\mathcal{V}_{\pi^{(t)}}$ be the* actual *value of acting according to this policy. The loss incurred by using policy $\pi^{(t)}$ as opposed to the optimal policy $\pi^*$ with value $\mathcal{V}^*$ is bounded by:*

$$\|\mathcal{V}^* - \mathcal{V}_{\pi^{(t)}}\|_\infty \leq \gamma^t \|\mathcal{V}^* - \mathcal{V}_{\pi^{(0)}}\|_\infty + \frac{2\gamma\overline{\beta}^{(t)}}{(1-\gamma)^2}. \quad (6)$$

**Proof:** See Appendix A.2. □

In words, Equation (6) shows that the difference between our approximation at iteration $t$ and the optimal value function is bounded by the sum of two terms. The first term is present in standard policy iteration and goes to zero exponentially fast. The second is the discounted accumulated projection error and, as Lemma 3.3 shows, is bounded. This second term can be minimized by choosing $\mathbf{w}^{(t)}$ as the one that minimizes:

$$\left\|\mathrm{H}\mathbf{w}^{(t)} - \left(R_{\pi^{(t)}} + \gamma P_{\pi^{(t)}}\mathrm{H}\mathbf{w}^{(t)}\right)\right\|_\infty,$$

which is exactly the computation performed by the max-norm projection. Therefore, this theorem motivates the use of max-norm projections to minimize the error term that appears in our bound.

The bounds we have provided so far may seem fairly trivial, as we have not provided a strong *a priori* bound on $\beta^{(t)}$. Fortunately, several factors make these bounds interesting despite the lack of *a priori* guarantees. If approximate policy iteration converges, as occurred in all of our experiments, we can obtain a much tighter bound: If $\widehat{\pi}$ is the policy after convergence, then:

$$\|\mathcal{V}^* - \mathcal{V}_{\widehat{\pi}}\|_\infty \leq \frac{2\gamma\beta_{\widehat{\pi}}}{(1-\gamma)},$$

where $\beta_{\widehat{\pi}}$ is the one-step max-norm projection error associated with estimating the value of $\widehat{\pi}$. Since the max-norm projection operation provides $\beta_{\widehat{\pi}}$, we can easily obtain an *a* *posteriori* bound as part of the policy iteration procedure. More details are provided in Section 7.

One could rewrite the bound in Theorem 3.5 in terms of the worst case projection error $\beta_P$, or the worst projection error in a cycle of policies, if approximate policy iteration gets stuck in a cycle. These formulations would be closer to the analysis of Bertsekas and Tsitsiklis (1996, Proposition 6.2, p.276). However, consider the case where most policies (or most policies in the final cycle) have a low projection error, but there are a few policies that cannot be approximated well using the projection operation, so that they have a large one-step projection error. A worst-case bound would be very loose, because it would be dictated by the error of the most difficult policy to approximate. On the other hand, using our discounted accumulated error formulation, errors introduced by policies that are hard to approximate decay very rapidly. Thus, the error bound represents an "average" case analysis: a decaying average of the projection errors for policies encountered at the successive iterations of the algorithm. As in the convergent case, this bound can be computed easily as part of the policy iteration procedure when max-norm projection is used.

The practical benefit of *a posteriori* bounds is that they can give meaningful feedback on the impact of the choice of the value function approximation architecture. While we are not explicitly addressing the difficult and general problem of feature selection in this paper, our error bounds motivate algorithms that aim to minimize the error *given* an approximation architecture and provide feedback that could be useful in future efforts to automatically discover or improve approximation architectures.

## 3.3 Approximate Linear Programming

### 3.3.1 The Exact Algorithm

Linear programming provides an alternative method for solving MDPs. It formulates the problem of finding a value function as a linear program (LP). Here the LP variables are $V_1, \ldots, V_N$, where $V_i$ represents $\mathcal{V}(\mathbf{x}_i)$: the value of starting at the $i$th state of the system. The LP is given by:

$$\begin{array}{ll} \text{Variables:} & V_1, \ldots, V_N \ ; \\ \text{Minimize:} & \sum_{\mathbf{x}_i} \alpha(\mathbf{x}_i)\ V_i \ ; \\ \text{Subject to:} & V_i \geq [R(\mathbf{x}_i, a) + \gamma \sum_j P(\mathbf{x}_j \mid \mathbf{x}_i, a) V_j] \quad \forall \mathbf{x}_i \in \mathbf{X}, a \in A, \end{array} \tag{7}$$

where the *state relevance weights* $\alpha$ are positive. Note that, in this exact case, the solution obtained is the same for any positive weight vector. It is interesting to note that steps of the simplex algorithm correspond to policy changes at single states, while steps of policy iteration can involve policy changes at multiple states. In practice, policy iteration tends to be faster than the linear programming approach (Puterman, 1994).

### 3.3.2 Approximate Linear Program

The approximate formulation for the LP approach, first proposed by Schweitzer and Seidmann (1985), restricts the space of allowable value functions to the linear space spanned by our basis functions. In this approximate formulation, the variables are $w_1, \ldots, w_k$: the weights for our basis functions. The LP is given by:

$$
\begin{array}{ll}
\text{Variables:} & w_1, \ldots, w_k \ ; \\
\text{Minimize:} & \sum_{\mathbf{x}} \alpha(\mathbf{x}) \sum_i w_i \ h_i(\mathbf{x}) \ ; \\
\text{Subject to:} & \sum_i w_i \ h_i(\mathbf{x}) \geq [R(\mathbf{x}, a) + \gamma \sum_{\mathbf{x}'} P(\mathbf{x}' \mid \mathbf{x}, a) \sum_i w_i \ h_i(\mathbf{x}')] \quad \forall \mathbf{x} \in \mathbf{X}, \forall a \in A.
\end{array}
\tag{8}
$$

In other words, this formulation takes the LP in (7) and substitutes the explicit state value function by a linear value function representation $\sum_i w_i \ h_i(\mathbf{x})$, or, in our more compact notation, $\mathcal{V}$ is replaced by $\mathrm{H}\mathbf{w}$. This linear program is guaranteed to be feasible if a constant function — a function with the same constant value for all states — is included in the set of basis functions.

In this approximate linear programming formulation, the choice of state relevance weights, $\alpha$, becomes important. Intuitively, not all constraints in this LP are binding; that is, the constraints are tighter for some states than for others. For each state $\mathbf{x}$, the relevance weight $\alpha(\mathbf{x})$ indicates the relative importance of a tight constraint. Therefore, unlike the exact case, the solution obtained may differ for different choices of the positive weight vector $\alpha$. Furthermore, there is, in general, no guarantee as to the quality of the greedy policy generated from the approximation $\mathrm{H}\mathbf{w}$. However, the recent work of de Farias and Van Roy (2001a) provides some analysis of the error relative to that of the best possible approximation in the subspace, and some guidance as to selecting $\alpha$ so as to improve the quality of the approximation. In particular, their analysis shows that this LP provides the best approximation $\mathrm{H}\mathbf{w}^*$ of the optimal value function $\mathcal{V}^*$ in a weighted $\mathcal{L}_1$ sense subject to the constraint that $\mathrm{H}\mathbf{w}^* \geq \mathcal{T}^*\mathrm{H}\mathbf{w}^*$, where the weights in the $\mathcal{L}_1$ norm are the state relevance weights $\alpha$.

The transformation from an exact to an approximate problem formulation has the effect of reducing the number of free variables in the LP to $k$ (one for each basis function coefficient), but the number of constraints remains $N \times |A|$. In our *SysAdmin* problem, for example, the number of constraints in the LP in (8) is $(m+1) \cdot 2^m$, where $m$ is the number of machines in the network. Thus, the process of generating the constraints and solving the LP still seems unmanageable for more than a few machines. In the next section, we discuss how we can use the structure of a factored MDP to provide for a compact representation and an efficient solution to this LP.

## 4. Factored Value Functions

The linear value function approach, and the algorithms described in Section 3, apply to any choice of basis functions. In the context of factored MDPs, Koller and Parr (1999) suggest a particular type of basis function, that is particularly compatible with the structure of a factored MDP. They suggest that, although the value function is typically not structured, there are many cases where it might be "close" to structured. That is, it might be well-approximated using a linear combination of functions each of which refers only to a small number of variables. More precisely, we define:

**Definition 4.1** *A* factored (linear) value function *is a linear function over the basis set $h_1, \ldots, h_k$, where the scope of each $h_i$ is restricted to some subset of variables $\mathbf{C}_i$.* □

Value functions of this type have a long history in the area of multi-attribute utility theory (Keeney & Raiffa, 1976). In our example, we might have a basis function $h_i$ for each machine, indicating whether it is working or not. Each basis function has scope restricted to $X_i$. These are represented as diamonds in the next time step in Figure 2(b).

Factored value functions provide the key to performing efficient computations over the exponential-sized state spaces we have in factored MDPs. The main insight is that restricted scope functions (including our basis functions) allow for certain basic operations to be implemented very efficiently. In the remainder of this section, we show how structure in factored MDPs can be exploited to perform two crucial operations very efficiently: one-step lookahead (backprojection), and the representation of exponentially many constraints in the LPs. Then, we use these basic building blocks to formulate very efficient approximation algorithms for factored MDPs, each presented in its own self-contained section: the approximate linear programming for factored MDPs in Section 5, and approximate policy iteration with max-norm projection in Section 6.

### 4.1 One-step Lookahead

A key step in all of our algorithms is the computation of the one-step lookahead value of some action $a$. This is necessary, for example, when computing the greedy policy as in Equation (1). Let's consider the computation of a $Q$ function, $Q_a(\mathbf{x})$, which represents the expected value the agent obtains after taking action $a$ at the current time step and receiving a long-term value $\mathcal{V}$ thereafter. This $Q$ function can be computed by:

$$Q_a(\mathbf{x}) = R(\mathbf{x}, a) + \gamma \sum_{\mathbf{x}'} P(\mathbf{x}' \mid \mathbf{x}, a)\mathcal{V}(\mathbf{x}). \tag{9}$$

That is, $Q_a(\mathbf{x})$ is given by the current reward plus the discounted expected future value. Using this notation, we can express the greedy policy as: $\mathit{Greedy}(\mathcal{V})(\mathbf{x}) = \max_a Q_a(\mathbf{x})$.

Recall that we are estimating the long-term value of our policy using a set of basis functions: $\mathcal{V}(\mathbf{x}) = \sum_i w_i\ h_i(\mathbf{x})$. Thus, we can rewrite Equation (9) as:

$$Q_a(\mathbf{x}) = R(\mathbf{x}, a) + \gamma \sum_{\mathbf{x}'} P(\mathbf{x}' \mid \mathbf{x}, a) \sum_i w_i\ h_i(\mathbf{x}). \tag{10}$$

The size of the state space is exponential, so that computing the expectation $\sum_{\mathbf{x}'} P(\mathbf{x}' \mid \mathbf{x}, a) \sum_i w_i\ h_i(\mathbf{x})$ seems infeasible. Fortunately, as discussed by Koller and Parr (1999), this expectation operation, or backprojection, can be performed efficiently if the transition model and the value function are both factored appropriately. The linearity of the value function permits a linear decomposition, where each summand in the expectation can be viewed as an independent value function and updated in a manner similar to the value iteration procedure used by Boutilier *et al.* (2000). We now recap the construction briefly, by first defining:

$$G^a(\mathbf{x}) = \sum_{\mathbf{x}'} P(\mathbf{x}' \mid \mathbf{x}, a) \sum_i w_i\ h_i(\mathbf{x}') = \sum_i w_i \sum_{\mathbf{x}'} P(\mathbf{x}' \mid \mathbf{x}, a) h_i(\mathbf{x}').$$

Thus, we can compute the expectation of each basis function separately:

$$g_i^a(\mathbf{x}) = \sum_{\mathbf{x}'} P(\mathbf{x}' \mid \mathbf{x}, a) h_i(\mathbf{x}'),$$

and then weight them by $w_i$ to obtain the total expectation $G^a(\mathbf{x}) = \sum_i w_i \; g_i^a(\mathbf{x})$. The intermediate function $g_i^a$ is called the *backprojection* of the basis function $h_i$ through the transition model $P_a$, which we denote by $g_i^a = P_a h_i$. Note that, in factored MDPs, the transition model $P_a$ is factored (represented as a DBN) and the basis functions $h_i$ have scope restricted to a small set of variables. These two important properties allow us to compute the backprojections very efficiently.

We now show how some restricted scope function $h$ (such as our basis functions) can be backprojected through some transition model $P_\tau$ represented as a DBN $\tau$. Here $h$ has scope restricted to $\mathbf{Y}$; our goal is to compute $g = P_\tau h$. We define the *back-projected scope of* $\mathbf{Y}$ *through* $\tau$ as the set of parents of $\mathbf{Y}'$ in the transition graph $G_\tau$; $\Gamma_\tau(\mathbf{Y}') = \cup_{Y_i' \in \mathbf{Y}'} \mathit{Parents}_\tau(Y_i')$. If intra-time slice arcs are included, so that $\mathit{Parents}_\tau(X_i') \in \{X_1, \ldots, X_n, X_1', \ldots, X_n'\}$, then the only change in our algorithm is in the definition of back-projected scope of $\mathbf{Y}$ through $\tau$. The definition now includes not only direct parents of $Y'$, but also all variables in $\{X_1, \ldots, X_n\}$ that are ancestors of $Y'$:

$$\Gamma_\tau(\mathbf{Y}') = \{X_j \mid \text{there exist a directed path from } X_j \text{ to any } X_i' \in \mathbf{Y}'\}.$$

Thus, the backprojected scope may become larger, but the functions are still factored.

We can now show that, if $h$ has scope restricted to $\mathbf{Y}$, then its backprojection $g$ has scope restricted to the parents of $\mathbf{Y}'$, *i.e.*, $\Gamma_\tau(\mathbf{Y}')$. Furthermore, each backprojection can be computed by only enumerating settings of variables in $\Gamma_\tau(\mathbf{Y}')$, rather than settings of all variables $\mathbf{X}$:

$$\begin{aligned}
g(\mathbf{x}) &= (P_\tau h)(\mathbf{x}); \\
&= \sum_{\mathbf{x}'} P_\tau(\mathbf{x}' \mid \mathbf{x}) h(\mathbf{x}'); \\
&= \sum_{\mathbf{x}'} P_\tau(\mathbf{x}' \mid \mathbf{x}) h(\mathbf{y}'); \\
&= \sum_{\mathbf{y}'} P_\tau(\mathbf{y}' \mid \mathbf{x}) h(\mathbf{y}') \sum_{\mathbf{u}' \in (\mathbf{x}' - \mathbf{y}')} P_\tau(\mathbf{u}' \mid \mathbf{x}); \\
&= \sum_{\mathbf{y}'} P_\tau(\mathbf{y}' \mid \mathbf{z}) h(\mathbf{y}'); \\
&= g(\mathbf{z});
\end{aligned}$$

where $\mathbf{z}$ is the value of $\Gamma_\tau(\mathbf{Y}')$ in $\mathbf{x}$ and the term $\sum_{\mathbf{u}' \in (\mathbf{x}' - \mathbf{y}')} P_\tau(\mathbf{u}' \mid \mathbf{x}) = 1$ as it is the sum of a probability distribution over a complete domain. Therefore, we see that $(P_\tau h)$ is a function whose scope is restricted to $\Gamma_\tau(\mathbf{Y}')$. Note that the cost of the computation depends linearly on $|\mathrm{Dom}(\Gamma_\tau(\mathbf{Y}'))|$, which depends on $\mathbf{Y}$ (the scope of $h$) and on the complexity of the process dynamics. This backprojection procedure is summarized in Figure 3.

Returning to our example, consider a basis function $h_i$ that is an indicator of variable $X_i$: it takes value 1 if the $i$th machine is working and 0 otherwise. Each $h_i$ has scope restricted to $X_i'$, thus, its backprojection $g_i$ has scope restricted to $\mathit{Parents}_\tau(X_i')$: $\Gamma_\tau(X_i') = \{X_{i-1}, X_i\}$.

### 4.2 Representing Exponentially Many Constraints

As seen in Section 3, both our approximation algorithms require the solution of linear programs: the LP in (5) for approximate policy iteration, and the LP in (8) for the approximate

*Backproj*$_a(h)$ — WHERE BASIS FUNCTION $h$ HAS SCOPE $\mathbf{C}$.
**Define** THE SCOPE OF THE BACKPROJECTION: $\Gamma_a(\mathbf{C}') = \cup_{X'_i \in \mathbf{C}'} Parents_a(X'_i)$.
**For** EACH ASSIGNMENT $\mathbf{y} \in \Gamma_a(\mathbf{C}')$:
$g^a(\mathbf{y}) = \sum_{\mathbf{c}' \in \mathbf{C}'} \prod_{i | X'_i \in \mathbf{C}'} P_a(\mathbf{c}'[X'_i] \mid \mathbf{y}) h(\mathbf{c}')$.

**Return** $g^a$.

Figure 3: Backprojection of basis function $h$.

linear programming algorithm. These LPs have some common characteristics: they have a small number of free variables (for $k$ basis functions there are $k+1$ free variables in approximate policy iteration and $k$ in approximate linear programming), but the number of constraints is still exponential in the number of state variables. However, in factored MDPs, these LP constraints have another very useful property: the functionals in the constraints have restricted scope. This key observation allows us to represent these constraints very compactly.

First, observe that the constraints in the linear programs are all of the form:

$$\phi \geq \sum_i w_i \; c_i(\mathbf{x}) - b(\mathbf{x}), \forall \mathbf{x}, \tag{11}$$

where only $\phi$ and $w_1, \ldots, w_k$ are free variables in the LP and $\mathbf{x}$ ranges over all states. This general form represents both the type of constraint in the max-norm projection LP in (5) and the approximate linear programming formulation in (8).[3]

The first insight in our construction is that we can replace the entire set of constraints in Equation (11) by one equivalent non-linear constraint:

$$\phi \geq \max_{\mathbf{x}} \sum_i w_i \; c_i(\mathbf{x}) - b(\mathbf{x}). \tag{12}$$

The second insight is that this new non-linear constraint can be implemented by a set of linear constraints using a construction that follows the structure of variable elimination in cost networks. This insight allows us to exploit structure in factored MDPs to represent this constraint compactly.

We tackle the problem of representing the constraint in Equation (12) in two steps: first, computing the maximum assignment for a fixed set of weights; then, representing the non-linear constraint by small set of linear constraints, using a construction we call the *factored LP*.

#### 4.2.1 Maximizing Over the State Space

The key computation in our algorithms is to represent a non-linear constraint of the form in Equation (12) efficiently by a small set of linear constraints. Before presenting this construction, let's first consider a simpler problem: Given some *fixed* weights $w_i$, we would like to compute the maximization: $\phi^* = \max_{\mathbf{x}} \sum_i w_i \; c_i(\mathbf{x}) - b(\mathbf{x})$, that is, the state $\mathbf{x}$, such

3. The complementary constraints in (5), $\phi \geq b(\mathbf{x}) - \sum_i w_i \; c_i(\mathbf{x})$, can be formulated using an analogous construction to the one we present in this section by changing the sign of $c_i(\mathbf{x})$ and $b(\mathbf{x})$. The approximate linear programming constraints of (8) can also be formulated in this form, as we show in Section 5.

that the difference between $\sum_i w_i \; c_i(\mathbf{x})$ and $b(\mathbf{x})$ is maximal. However, we cannot explicitly enumerate the exponential number of states and compute the difference. Fortunately, structure in factored MDPs allows us to compute this maximum efficiently.

In the case of factored MDPs, our state space is a set of vectors $\mathbf{x}$ which are assignments to the state variables $\mathbf{X} = \{X_1, \ldots, X_n\}$. We can view both $C\mathbf{w}$ and $\mathbf{b}$ as functions of these state variables, and hence also their difference. Thus, we can define a function $F^{\mathbf{w}}(X_1, \ldots, X_n)$ such that $F^{\mathbf{w}}(\mathbf{x}) = \sum_i w_i \; c_i(\mathbf{x}) - b(\mathbf{x})$. Note that we have executed a representation shift; we are viewing $F^{\mathbf{w}}$ as a function of the variables $\mathbf{X}$, which is parameterized by $\mathbf{w}$. Recall that the size of the state space is exponential in the number of variables. Hence, our goal in this section is to compute $\max_{\mathbf{x}} F^{\mathbf{w}}(\mathbf{x})$ without explicitly considering each of the exponentially many states. The solution is to use the fact that $F^{\mathbf{w}}$ has a factored representation. More precisely, $C\mathbf{w}$ has the form $\sum_i w_i \; c_i(\mathbf{Z}_i)$, where $\mathbf{Z}_i$ is a subset of $\mathbf{X}$. For example, we might have $c_1(X_1, X_2)$ which takes value 1 in states where $X_1 = \mathit{true}$ and $X_2 = \mathit{false}$ and 0 otherwise. Similarly, the vector $\mathbf{b}$ in our case is also a sum of restricted scope functions. Thus, we can express $F^{\mathbf{w}}$ as a sum $\sum_j f_j^{\mathbf{w}}(\mathbf{Z}_j)$, where $f_j^{\mathbf{w}}$ may or may not depend on $\mathbf{w}$. In the future, we sometimes drop the superscript $\mathbf{w}$ when it is clear from context.

Using our more compact notation, our goal here is simply to compute $\max_{\mathbf{x}} \sum_i w_i \; c_i(\mathbf{x}) - b(\mathbf{x}) = \max_{\mathbf{x}} F^{\mathbf{w}}(\mathbf{x})$, that is, to find the state $\mathbf{x}$ over which $F^{\mathbf{w}}$ is maximized. Recall that $F^{\mathbf{w}} = \sum_{j=1}^{m} f_j(\mathbf{Z}_j)$. We can maximize such a function, $F^{\mathbf{w}}$, without enumerating every state using *non-serial dynamic programming* (Bertele & Brioschi, 1972). The idea is virtually identical to variable elimination in a Bayesian network. We review this construction here, as it is a central component in our solution LP.

Our goal is to compute

$$\max_{x_1, \ldots, x_n} \sum_j f_j(\mathbf{x}[\mathbf{Z}_j]).$$

The main idea is that, rather than summing all functions and then doing the maximization, we maximize over variables one at a time. When maximizing over $x_l$, only summands involving $x_l$ participate in the maximization.

**Example 4.2** *Assume*

$$F = f_1(x_1, x_2) + f_2(x_1, x_3) + f_3(x_2, x_4) + f_4(x_3, x_4).$$

*We therefore wish to compute:*

$$\max_{x_1, x_2, x_3, x_4} f_1(x_1, x_2) + f_2(x_1, x_3) + f_3(x_2, x_4) + f_4(x_3, x_4).$$

*We can first compute the maximum over $x_4$; the functions $f_1$ and $f_2$ are irrelevant, so we can push them out. We get*

$$\max_{x_1, x_2, x_3} f_1(x_1, x_2) + f_2(x_1, x_3) + \max_{x_4} [f_3(x_2, x_4) + f_4(x_3, x_4)].$$

*The result of the internal maximization depends on the values of $x_2, x_3$; thus, we can introduce a new function $e_1(X_2, X_3)$ whose value at the point $x_2, x_3$ is the value of the internal* max *expression. Our problem now reduces to computing*

$$\max_{x_1, x_2, x_3} f_1(x_1, x_2) + f_2(x_1, x_3) + e_1(x_2, x_3),$$

VariableElimination ($\mathcal{F}$, $\mathcal{O}$)
  *//$\mathcal{F} = \{f_1, \ldots, f_m\}$ is the set of functions to be maximized;*
  *//$\mathcal{O}$ stores the elimination order.*
 **For** $i = 1$ to number of variables:
  *//Select the next variable to be eliminated.*
  **Let** $l = \mathcal{O}(i)$ ;
  *//Select the relevant functions.*
  **Let** $e_1, \ldots, e_L$ be the functions in $\mathcal{F}$ whose scope contains $X_l$.
  *//Maximize over current variable $X_l$.*
  **Define** a new function $e = \max_{x_l} \sum_{j=1}^{L} e_j$ ; note that $\text{Scope}[e] = \cup_{j=1}^{L}\text{Scope}[e_j] - \{X_l\}$.
  *//Update set of functions.*
  **Update** the set of functions $\mathcal{F} = \mathcal{F} \cup \{e\} \setminus \{e_1, \ldots, e_L\}$.
 *//Now, all functions have empty scope and their sum is the maximum value of $f_1 + \cdots + f_m$.*
 **Return** the maximum value $\sum_{e_i \in \mathcal{F}} e_i$.

Figure 4: Variable elimination procedure for computing the maximum value $f_1 + \cdots + f_m$, where each $f_i$ is a restricted scope function.

*having one fewer variable. Next, we eliminate another variable, say $X_3$, with the resulting expression reducing to:*

$$\max_{x_1,x_2} f_1(x_1, x_2) + e_2(x_1, x_2),$$

$$\textit{where} \quad e_2(x_1, x_2) = \max_{x_3}[f_2(x_1, x_3) + e_1(x_2, x_3)].$$

*Finally, we define*

$$e_3 = \max_{x_1,x_2} f_1(x_1, x_2) + e_2(x_1, x_2).$$

*The result at this point is a number, which is the desired maximum over $x_1, \ldots, x_4$. While the naive approach of enumerating all states requires 63 arithmetic operations if all variables are binary, using variable elimination we only need to perform 23 operations.* □

The general variable elimination algorithm is described in Figure 4. The inputs to the algorithm are the functions to be maximized $\mathcal{F} = \{f_1, \ldots, f_m\}$ and an elimination ordering $\mathcal{O}$ on the variables, where $\mathcal{O}(i)$ returns the $i$th variable to be eliminated. As in the example above, for each variable $X_l$ to be eliminated, we select the relevant functions $e_1, \ldots, e_L$, those whose scope contains $X_l$. These functions are removed from the set $\mathcal{F}$ and we introduce a new function $e = \max_{x_l} \sum_{j=1}^{L} e_j$. At this point, the scope of the functions in $\mathcal{F}$ no longer depends on $X_l$, that is, $X_l$ has been 'eliminated'. This procedure is repeated until all variables have been eliminated. The remaining functions in $\mathcal{F}$ thus have empty scope. The desired maximum is therefore given by the sum of these remaining functions.

The computational cost of this algorithm is linear in the number of new "function values" introduced in the elimination process. More precisely, consider the computation of a new function $e$ whose scope is $\mathbf{Z}$. To compute this function, we need to compute $|\text{Dom}[\mathbf{Z}]|$ different values. The cost of the algorithm is linear in the overall number of these values, introduced throughout the execution. As shown by Dechter (1999), this cost is exponential in the induced width of the *cost network*, the undirected graph defined over the variables $X_1, \ldots, X_n$, with an edge between $X_l$ and $X_m$ if they appear together in one of the original functions $f_j$. The complexity of this algorithm is, of course, dependent on the variable elimination order and the problem structure. Computing the optimal elimination order is an NP-hard problem (Arnborg, Corneil, & Proskurowski, 1987) and elimination orders yielding low induced tree width do not exist for some problems. These issues have been confronted successfully for a large variety of practical problems in the Bayesian network community, which has benefited from a large variety of good heuristics which have been developed for the variable elimination ordering problem (Bertele & Brioschi, 1972; Kjaerulff, 1990; Reed, 1992; Becker & Geiger, 2001).

#### 4.2.2 Factored LP

In this section, we present the centerpiece of our planning algorithms: a new, general approach for compactly representing exponentially large sets of LP constraints in problems with factored structure — those where the functions in the constraints can be decomposed as the sum of restricted scope functions. Consider our original problem of representing the non-linear constraint in Equation (12) compactly. Recall that we wish to represent the non-linear constraint $\phi \geq \max_{\mathbf{x}} \sum_i w_i \; c_i(\mathbf{x}) - b(\mathbf{x})$, or equivalently, $\phi \geq \max_{\mathbf{x}} F^{\mathbf{w}}(\mathbf{x})$, without generating one constraint for each state as in Equation (11). The new, key insight is that this non-linear constraint can be implemented using a construction that follows the structure of variable elimination in cost networks.

Consider any function $e$ used within $\mathcal{F}$ (including the original $f_i$'s), and let $\mathbf{Z}$ be its scope. For any assignment $\mathbf{z}$ to $\mathbf{Z}$, we introduce variable $u^e_{\mathbf{z}}$, whose value represents $e_{\mathbf{z}}$, into the linear program. For the initial functions $f_i^{\mathbf{w}}$, we include the constraint that $u^{f_i}_{\mathbf{z}} = f_i^{\mathbf{w}}(\mathbf{z})$. As $f_i^{\mathbf{w}}$ is linear in $\mathbf{w}$, this constraint is linear in the LP variables. Now, consider a new function $e$ introduced into $\mathcal{F}$ by eliminating a variable $X_l$. Let $e_1, \ldots, e_L$ be the functions extracted from $\mathcal{F}$, and let $\mathbf{Z}$ be the scope of the resulting $e$. We introduce a set of constraints:

$$u^e_{\mathbf{z}} \geq \sum_{j=1}^{L} u^{e_j}_{(\mathbf{z},x_l)[\mathbf{Z}_j]} \qquad \forall x_l. \tag{13}$$

Let $e_n$ be the last function generated in the elimination, and recall that its scope is empty. Hence, we have only a single variable $u^{e_n}$. We introduce the additional constraint $\phi \geq u^{e_n}$.

The complete algorithm, presented in Figure 5, is divided into three parts: First, we generate equality constraints for functions that depend on the weights $w_i$ (basis functions). In the second part, we add the equality constraints for functions that do not depend on the weights (target functions). These equality constraints let us abstract away the differences between these two types of functions and manage them in a unified fashion in the third part of the algorithm. This third part follows a procedure similar to variable elimination described in Figure 4. However, unlike standard variable elimination where we would introduce a new function $e$, such that $e = \max_{x_l} \sum_{j=1}^{L} e_j$, in our factored LP procedure we introduce new LP variables $u^e_{\mathbf{z}}$. To enforce the definition of $e$ as the maximum over $X_l$ of $\sum_{j=1}^{L} e_j$, we introduce the new LP constraints in Equation (13).

**Example 4.3** *To understand this construction, consider our simple example above, and assume we want to express the fact that* $\phi \geq \max_{\mathbf{x}} F^{\mathbf{w}}(\mathbf{x})$. *We first introduce a set of*

FactoredLP ($C$, $\mathbf{b}$,$\mathcal{O}$)

*// $C = \{c_1, \ldots, c_k\}$ is the set of basis functions.*
*// $\mathbf{b} = \{b_1, \ldots, b_m\}$ is the set of target functions.*
*//$\mathcal{O}$ stores the elimination order.*
*//Return a (polynomial) set of constraints $\Omega$ equivalent to $\phi \geq \sum_i w_i c_i(\mathbf{x}) + \sum_j b_j(\mathbf{x}), \forall \mathbf{x}$ .*

*//Data structure for the constraints in factored LP.*
**Let** $\Omega = \{\}$ .
*//Data structure for the intermediate functions generated in variable elimination.*
**Let** $\mathcal{F} = \{\}$ .
*//Generate equality constraint to abstract away basis functions.*
**For** each $c_i \in C$:
- **Let** $\mathbf{Z} = \text{Scope}[c_i]$.
- **For** each assignment $\mathbf{z} \in \mathbf{Z}$, create a new LP variable $u^{f_i}_{\mathbf{z}}$ and add a constraint to $\Omega$:

$$u^{f_i}_{\mathbf{z}} = w_i c_i(\mathbf{z}).$$

- **Store** new function $f_i$ to use in variable elimination step: $\mathcal{F} = \mathcal{F} \cup \{f_i\}$.

*//Generate equality constraint to abstract away target functions.*
**For** each $b_j \in \mathbf{b}$:
- **Let** $\mathbf{Z} = \text{Scope}[b_j]$.
- **For** each assignment $\mathbf{z} \in \mathbf{Z}$, create a new LP variable $u^{f_j}_{\mathbf{z}}$ and add a constraint to $\Omega$:

$$u^{f_j}_{\mathbf{z}} = b_j(\mathbf{z}).$$

- **Store** new function $f_j$ to use in variable elimination step: $\mathcal{F} = \mathcal{F} \cup \{f_j\}$.

*//Now, $\mathcal{F}$ contains all of the functions involved in the LP, our constraints become: $\phi \geq \sum_{e_i \in \mathcal{F}} e_i(\mathbf{x}), \forall \mathbf{x}$ , which we represent compactly using a variable elimination procedure.*
**For** $i = 1$ to number of variables:
- *//Select the next variable to be eliminated.*
- **Let** $l = \mathcal{O}(i)$ ;
- *//Select the relevant functions.*
- **Let** $e_1, \ldots, e_L$ be the functions in $\mathcal{F}$ whose scope contains $X_l$, and let $\mathbf{Z}_j = \text{Scope}[e_j]$.
- *//Introduce linear constraints for the maximum over current variable $X_l$.*
- **Define** a new function $e$ with scope $\mathbf{Z} = \cup_{j=1}^{L} \mathbf{Z}_j - \{X_l\}$ to represent $\max_{x_l} \sum_{j=1}^{L} e_j$.
- **Add** constraints to $\Omega$ to enforce maximum: for each assignment $\mathbf{z} \in \mathbf{Z}$:

$$u^{e}_{\mathbf{z}} \geq \sum_{j=1}^{L} u^{e_j}_{(\mathbf{z},x_l)[\mathbf{Z}_j]} \qquad \forall x_l.$$

- *//Update set of functions.*
- **Update** the set of functions $\mathcal{F} = \mathcal{F} \cup \{e\} \setminus \{e_1, \ldots, e_L\}$.

*//Now, all variables have been eliminated and all functions have empty scope.*
**Add** last constraint to $\Omega$:

$$\phi \geq \sum_{e_i \in \mathcal{F}} e_i.$$

**Return** $\Omega$.

Figure 5: Factored LP algorithm for the compact representation of the exponential set of constraints $\phi \geq \sum_i w_i c_i(\mathbf{x}) + \sum_j b_j(\mathbf{x}), \forall \mathbf{x}$.

*variables $u^{f_1}_{x_1,x_2}$ for every instantiation of values $x_1, x_2$ to the variables $X_1, X_2$. Thus, if $X_1$ and $X_2$ are both binary, we have four such variables. We then introduce a constraint defining the value of $u^{f_1}_{x_1,x_2}$ appropriately. For example, for our $f_1$ above, we have $u^{f_1}_{t,t} = 0$ and $u^{f_1}_{t,f} = w_1$. We have similar variables and constraints for each $f_j$ and each value $\mathbf{z}$ in $\mathbf{Z}_j$. Note that each of the constraints is a simple equality constraint involving numerical constants and perhaps the weight variables $\mathbf{w}$.*

*Next, we introduce variables for each of the intermediate expressions generated by variable elimination. For example, when eliminating $X_4$, we introduce a set of LP variables $u^{e_1}_{x_2,x_3}$; for each of them, we have a set of constraints*

$$u^{e_1}_{x_2,x_3} \geq u^{f_3}_{x_2,x_4} + u^{f_4}_{x_3,x_4}$$

*one for each value $x_4$ of $X_4$. We have a similar set of constraint for $u^{e_2}_{x_1,x_2}$ in terms of $u^{f_2}_{x_1,x_3}$ and $u^{e_1}_{x_2,x_3}$. Note that each constraint is a simple linear inequality.* □

We can now prove that our factored LP construction represents the same constraint as non-linear constraint in Equation (12):

**Theorem 4.4** *The constraints generated by the factored LP construction are equivalent to the non-linear constraint in Equation (12). That is, an assignment to $(\phi, \mathbf{w})$ satisfies the factored LP constraints if and only if it satisfies the constraint in Equation (12).*

**Proof:** See Appendix A.3. □

Returning to our original formulation, we have that $\sum_j f^{\mathbf{w}}_j$ is $C\mathbf{w} - \mathbf{b}$ in the original set of constraints. Hence our new set of constraints is equivalent to the original set: $\phi \geq \max_{\mathbf{x}} \sum_i w_i \; c_i(\mathbf{x}) - b(\mathbf{x})$ in Equation (12), which in turn is equivalent to the exponential set of constraints $\phi \geq \sum_i w_i \; c_i(\mathbf{x}) - b(\mathbf{x}), \forall \mathbf{x}$ in Equation (11). Thus, we can represent this exponential set of constraints by a new set of constraints and LP variables. The size of this new set, as in variable elimination, is exponential only in the induced width of the cost network, rather than in the total number of variables.

In this section, we presented a new, general approach for compactly representing exponentially-large sets of LP constraints in problems with factored structure. In the remainder of this paper, we exploit this construction to design efficient planning algorithms for factored MDPs.

#### 4.2.3 Factored Max-norm Projection

We can now use our procedure for representing the exponential number of constraints in Equation (11) compactly to compute efficient max-norm projections, as in Equation (4):

$$\mathbf{w}^* \in \arg\min_{\mathbf{w}} \|C\mathbf{w} - \mathbf{b}\|_\infty .$$

The max-norm projection is computed by the linear program in (5). There are two sets of constraints in this LP: $\phi \geq \sum_{j=1}^k c_{ij} w_j - b_i, \forall i$ and $\phi \geq b_i - \sum_{j=1}^k c_{ij} w_j, \forall i$. Each of these sets is an instance of the constraints in Equation (11), which we have just addressed in the previous section. Thus, if each of the $k$ basis functions in $C$ is a restricted scope function and the target function $\mathbf{b}$ is the sum of restricted scope functions, then we can use our factored LP technique to represent the constraints in the max-norm projection LP compactly. The correctness of our algorithm is a corollary of Theorem 4.4:

**Corollary 4.5** *The solution* $(\phi^*, \mathbf{w}^*)$ *of a linear program that minimizes* $\phi$ *subject to the constraints in* FactoredLP*(*$C$*,* $-\mathbf{b}$*,*$\mathcal{O}$*) and* FactoredLP*(*$-C$*,* $\mathbf{b}$*,*$\mathcal{O}$*), for any elimination order* $\mathcal{O}$ *satisfies:*

$$\mathbf{w}^* \in \arg\min_{\mathbf{w}} \|C\mathbf{w} - \mathbf{b}\|_\infty, \qquad \textit{and} \qquad \phi^* = \min_{\mathbf{w}} \|C\mathbf{w} - \mathbf{b}\|_\infty. \ \square$$

The original max-norm projection LP had $k+1$ variables and two constraints for each state $\mathbf{x}$; thus, the number of constraints is exponential in the number of state variables. On the other hand, our new factored max-norm projection LP has more variables, but exponentially fewer constraints. The number of variables and constraints in the new factored LP is exponential only in the number of state variables in the largest factor in the cost network, rather than exponential in the total number of state variables. As we show in Section 9, this exponential gain allows us to compute max-norm projections efficiently when solving very large factored MDPs.

## 5. Approximate Linear Programming

We begin with the simplest of our approximate MDP solution algorithms, based on the approximate linear programming formulation in Section 3.3. Using the basic operations described in Section 4, we can formulate an algorithm that is both simple and efficient.

### 5.1 The Algorithm

As discussed in Section 3.3, approximate linear program formulation is based on the linear programming approach to solving MDPs presented in Section 3.3. However, in this approximate version, we restrict the space of value functions to the linear space defined by our basis functions. More precisely, in this approximate LP formulation, the variables are $w_1, \ldots, w_k$ — the weights for our basis functions. The LP is given by:

$$\begin{array}{ll} \text{Variables:} & w_1, \ldots, w_k \,; \\ \text{Minimize:} & \sum_{\mathbf{x}} \alpha(\mathbf{x}) \sum_i w_i \ h_i(\mathbf{x}) \,; \\ \text{Subject to:} & \sum_i w_i \ h_i(\mathbf{x}) \geq [R(\mathbf{x}, a) + \gamma \sum_{\mathbf{x}'} P(\mathbf{x}' \mid \mathbf{x}, a) \sum_i w_i \ h_i(\mathbf{x}')] \quad \forall \mathbf{x} \in \mathbf{X}, \forall a \in A. \end{array} \tag{14}$$

In other words, this formulation takes the LP in (7) and substitutes the explicit state value function with a linear value function representation $\sum_i w_i \ h_i(\mathbf{x})$. This transformation from an exact to an approximate problem formulation has the effect of reducing the number of free variables in the LP to $k$ (one for each basis function coefficient), but the number of constraints remains $|\mathbf{X}| \times |A|$. In our *SysAdmin* problem, for example, the number of constraints in the LP in (14) is $(m+1) \cdot 2^m$, where $m$ is the number of machines in the network. However, using our algorithm for representing exponentially large constraint sets compactly we are able to compute the solution to this approximate linear programming algorithm in *closed form* with an exponentially smaller LP, as in Section 4.2.

First, consider the objective function $\sum_{\mathbf{x}} \alpha(\mathbf{x}) \sum_i w_i \ h_i(\mathbf{x})$ of the LP (14). Naively representing this objective function requires a summation over a exponentially large state space. However, we can rewrite the objective and obtain a compact representation. We first reorder the terms:

FactoredALP ($P$, $R$, $\gamma$, $H$, $\mathcal{O}$, $\alpha$)

*//$P$ is the factored transition model.*
*//$R$ is the set of factored reward functions.*
*//$\gamma$ is the discount factor.*
*//$H$ is the set of basis functions $H = \{h_1, \ldots, h_k\}$.*
*//$\mathcal{O}$ stores the elimination order.*
*//$\alpha$ are the state relevance weights.*
*//Return the basis function weights $\mathbf{w}$ computed by approximate linear programming.*
*//Cache the backprojections of the basis functions.*
**For** each basis function $h_i \in H$; for each action $a$:
  **Let** $g_i^a = Backproj_a(h_i)$.
*//Compute factored state relevance weights.*
**For** each basis function $h_i$, compute the factored state relevance weights $\alpha_i$ as in Equation (15) .
*//Generate approximate linear programming constraints*
**Let** $\Omega = \{\}$.
**For** each action $a$:
  **Let** $\Omega = \Omega \cup \text{FactoredLP}(\{\gamma g_1^a - h_1, \ldots, \gamma g_k^a - h_k\}, R^a, \mathcal{O})$.
*//So far, our constraints guarantee that $\phi \geq R(\mathbf{x}, a) + \gamma \sum_{\mathbf{x}'} P(\mathbf{x}' \mid \mathbf{x}, a) \sum_i w_i \; h_i(\mathbf{x}') - \sum_i w_i \; h_i(\mathbf{x})$; to satisfy the approximate linear programming solution in (14) we must add a final constraint.*
**Let** $\Omega = \Omega \cup \{\phi = 0\}$.
*//We can now obtain the solution weights by solving an LP.*
**Let** $\mathbf{w}$ be the solution of the linear program: minimize $\sum_i \alpha_i w_i$, subject to the constraints $\Omega$.
**Return** $\mathbf{w}$.

Figure 6: Factored approximate linear programming algorithm.

$$\sum_{\mathbf{x}} \alpha(\mathbf{x}) \sum_i w_i \; h_i(\mathbf{x}) = \sum_i w_i \sum_{\mathbf{x}} \alpha(\mathbf{x}) \; h_i(\mathbf{x}).$$

Now, consider the state relevance weights $\alpha(\mathbf{x})$ as a distribution over states, so that $\alpha(\mathbf{x}) > 0$ and $\sum_{\mathbf{x}} \alpha(\mathbf{x}) = 1$. As in backprojections, we can now write:

$$\alpha_i = \sum_{\mathbf{x}} \alpha(\mathbf{x}) \; h_i(\mathbf{x}) = \sum_{\mathbf{c}_i \in \mathbf{C}_i} \alpha(\mathbf{c}_i) \; h_i(\mathbf{c}_i); \tag{15}$$

where $\alpha(\mathbf{c}_i)$ represents the marginal of the state relevance weights $\alpha$ over the domain $\text{Dom}[\mathbf{C}_i]$ of the basis function $h_i$. For example, if we use uniform state relevance weights as in our experiments — $\alpha(\mathbf{x}) = \frac{1}{|\mathbf{X}|}$ — then the marginals become $\alpha(\mathbf{c}_i) = \frac{1}{|\mathbf{C}_i|}$. Thus, we can rewrite the objective function as $\sum_i w_i \; \alpha_i$, where each basis weight $\alpha_i$ is computed as shown in Equation (15). If the state relevance weights are represented by marginals, then the cost of computing each $\alpha_i$ depends exponentially on the size of the scope of $\mathbf{C}_i$ only, rather than exponentially on the number of state variables. On the other hand, if the state relevance weights are represented by arbitrary distributions, we need to obtain the marginals over the $\mathbf{C}_i$'s, which may not be an efficient computation. Thus, greatest efficiency is achieved by using a compact representation, such as a Bayesian network, for the state relevance weights.

Second, note that the right side of the constraints in the LP (14) correspond to the $Q_a$ functions:

$$Q_a(\mathbf{x}) = R^a(\mathbf{x}) + \gamma \sum_{\mathbf{x}'} P(\mathbf{x}' \mid \mathbf{x}, a) \sum_i w_i \; h_i(\mathbf{x}').$$

Using the efficient backprojection operation in factored MDPs described in Section 4.1 we can rewrite the $Q_a$ functions as:

$$Q_a(\mathbf{x}) = R^a(\mathbf{x}) + \gamma \sum_i w_i \; g_i^a(\mathbf{x});$$

where $g_i^a$ is the backprojection of basis function $h_i$ through the transition model $P_a$. As we discussed, if $h_i$ has scope restricted to $\mathbf{C}_i$, then $g_i^a$ is a restricted scope function of $\Gamma_a(\mathbf{C}_i')$.

We can precompute the backprojections $g_i^a$ and the basis relevance weights $\alpha_i$. The approximate linear programming LP of (14) can be written as:

$$\begin{array}{ll} \text{Variables:} & w_1, \ldots, w_k \; ; \\ \text{Minimize:} & \sum_i \alpha_i \; w_i \; ; \\ \text{Subject to:} & \sum_i w_i \; h_i(\mathbf{x}) \geq [R^a(\mathbf{x}) + \gamma \sum_i w_i \; g_i^a(\mathbf{x})] \quad \forall \mathbf{x} \in \mathbf{X}, \forall a \in A. \end{array} \tag{16}$$

Finally, we can rewrite this LP to use constraints of the same form as the one in Equation (12):

$$\begin{array}{ll} \text{Variables:} & w_1, \ldots, w_k \; ; \\ \text{Minimize:} & \sum_i \alpha_i \; w_i \; ; \\ \text{Subject to:} & 0 \geq \max_{\mathbf{x}} \left\{ R^a(\mathbf{x}) + \sum_i w_i \; [\gamma g_i^a(\mathbf{x}) - h_i(\mathbf{x})] \right\} \quad \forall a \in A. \end{array} \tag{17}$$

We can now use our factored LP construction in Section 4.2 to represent these non-linear constraints compactly. Basically, there is one set of factored LP constraints for each action $a$. Specifically, we can write the non-linear constraint in the same form as those in Equation (12) by expressing the functions $C$ as: $c_i(\mathbf{x}) = h_i(\mathbf{x}) - \gamma g_i^a(\mathbf{x})$. Each $c_i(\mathbf{x})$ is a restricted scope function; that is, if $h_i(\mathbf{x})$ has scope restricted to $\mathbf{C}_i$, then $g_i^a(\mathbf{x})$ has scope restricted to $\Gamma_a(\mathbf{C}'_i)$, which means that $c_i(\mathbf{x})$ has scope restricted to $\mathbf{C}_i \cup \Gamma_a(\mathbf{C}'_i)$. Next, the target function $\mathbf{b}$ becomes the reward function $R^a(\mathbf{x})$ which, by assumption, is factored. Finally, in the constraint in Equation (12), $\phi$ is a free variable. On the other hand, in the LP in (17) the maximum in the right hand side must be less than zero. This final condition can be achieved by adding a constraint $\phi = 0$. Thus, our algorithm generates a set of factored LP constraints, one for each action. The total number of constraints and variables in this new LP is linear in the number of actions $|A|$ and only exponential in the induced width of each cost network, rather than in the total number of variables. The complete factored approximate linear programming algorithm is outlined in Figure 6.

### 5.2 An Example

We now present a complete example of the operations required by the approximate LP algorithm to solve the factored MDP shown in Figure 2(a). Our presentation follows four steps: problem representation, basis function selection, backprojections and LP construction.

**Problem Representation:** First, we must fully specify the factored MDP model for the problem. The structure of the DBN is shown in Figure 2(b). This structure is maintained for all action choices. Next, we must define the transition probabilities for each action. There are 5 actions in this problem: do nothing, or reboot one of the 4 machines in the network. The CPDs for these actions are shown in Figure 2(c). Finally, we must define the reward function. We decompose the global reward as the sum of 4 local reward functions, one for each machine, such that there is a reward if the machine is working. Specifically, $R_i(X_i = \mathit{true}) = 1$ and $R_i(X_i = \mathit{false}) = 0$, breaking symmetry by setting $R_4(X_4 = \mathit{true}) = 2$. We use a discount factor of $\gamma = 0.9$.

**Basis Function Selection:** In this simple example, we use five simple basis functions. First, we include the constant function $h_0 = 1$. Next, we add indicators for each machine which take value 1 if the machine is working: $h_i(X_i = \mathit{true}) = 1$ and $h_i(X_i = \mathit{false}) = 0$.

**Backprojections:** The first algorithmic step is computing the backprojection of the basis functions, as defined in Section 4.1. The backprojection of the constant basis is simple:

$$
\begin{aligned}
g_0^a &= \sum_{\mathbf{x}'} P_a(\mathbf{x}' \mid \mathbf{x}) h_0 \; ; \\
&= \sum_{\mathbf{x}'} P_a(\mathbf{x}' \mid \mathbf{x}) \; 1 \; ; \\
&= 1 \; .
\end{aligned}
$$

Next, we must backproject our indicator basis functions $h_i$:

$$
\begin{aligned}
g_i^a &= \sum_{\mathbf{x}'} P_a(\mathbf{x}' \mid \mathbf{x}) h_i(x'_i) \; ; \\
&= \sum_{x'_1, x'_2, x'_3, x'_4} \prod_j P_a(x'_j \mid x_{j-1}, x_j) h_i(x'_i) \; ;
\end{aligned}
$$

$$
\begin{aligned}
&= \sum_{x'_i} P_a(x'_i \mid x_{i-1}, x_i) h_i(x'_i) \sum_{\mathbf{x}'[\mathbf{X}'-\{X'_i\}]} \prod_{j \neq i} P_a(x'_j \mid x_{j-1}, x_j) \ ; \\
&= \sum_{x'_i} P_a(x'_i \mid x_{i-1}, x_i) h_i(x'_i) \ ; \\
&= P_a(X'_i = \mathit{true} \mid x_{i-1}, x_i) \ 1 + P_a(X'_i = \mathit{false} \mid x_{i-1}, x_i) \ 0 \ ; \\
&= P_a(X'_i = \mathit{true} \mid x_{i-1}, x_i) \ .
\end{aligned}
$$

Thus, $g_i^a$ is a restricted scope function of $\{X_{i-1}, X_i\}$. We can now use the CPDs in Figure 2(c) to specify $g_i^a$:

$g_i^{\mathit{reboot} = i}(X_{i-1}, X_i) =$

| | $X_i = \mathit{true}$ | $X_i = \mathit{false}$ |
|---|---|---|
| $X_{i-1} = \mathit{true}$ | 1 | 1 |
| $X_{i-1} = \mathit{false}$ | 1 | 1 |

;

$g_i^{\mathit{reboot} \neq i}(X_{i-1}, X_i) =$

| | $X_i = \mathit{true}$ | $X_i = \mathit{false}$ |
|---|---|---|
| $X_{i-1} = \mathit{true}$ | 0.9 | 0.09 |
| $X_{i-1} = \mathit{false}$ | 0.5 | 0.05 |

.

**LP Construction:** To illustrate the factored LPs constructed by our algorithms, we define the constraints for the approximate linear programming approach presented above. First, we define the functions $c_i^a = \gamma g_i^a - h_i$, as shown in Equation (17). In our example, these functions are $c_0^a = \gamma - 1 = -0.1$ for the constant basis, and for the indicator bases:

$c_i^{\mathit{reboot} = i}(X_{i-1}, X_i) =$

| | $X_i = \mathit{true}$ | $X_i = \mathit{false}$ |
|---|---|---|
| $X_{i-1} = \mathit{true}$ | −0.1 | 0.9 |
| $X_{i-1} = \mathit{false}$ | −0.1 | 0.9 |

;

$c_i^{\mathit{reboot} \neq i}(X_{i-1}, X_i) =$

| | $X_i = \mathit{true}$ | $X_i = \mathit{false}$ |
|---|---|---|
| $X_{i-1} = \mathit{true}$ | −0.19 | 0.081 |
| $X_{i-1} = \mathit{false}$ | −0.55 | 0.045 |

.

Using this definition of $c_i^a$, the approximate linear programming constraints are given by:

$$0 \geq \max_{\mathbf{x}} \sum_i R_i + \sum_j w_j c_j^a \ , \ \forall a \ . \tag{18}$$

We present the LP construction for one of the 5 actions: $\mathit{reboot} = 1$. Analogous constructions can be made for the other actions.

In the first set of constraints, we abstract away the difference between rewards and basis functions by introducing LP variables $u$ and equality constraints. We begin with the reward functions:

$$
\begin{aligned}
u_{x_1}^{R_1} = 1 \ , \ u_{\bar{x}_1}^{R_1} = 0 \ ; \qquad & u_{x_2}^{R_2} = 1 \ , \ u_{\bar{x}_2}^{R_2} = 0 \ ; \\
u_{x_3}^{R_3} = 1 \ , \ u_{\bar{x}_3}^{R_3} = 0 \ ; \qquad & u_{x_4}^{R_4} = 2 \ , \ u_{\bar{x}_4}^{R_4} = 0 \ .
\end{aligned}
$$

We now represent the equality constraints for the $c_j^a$ functions for the $\mathit{reboot} = 1$ action. Note that the appropriate basis function weight from Equation (18) appears in these constraints:

$$
\begin{array}{llll}
u^{c_0} = -0.1\ w_0 \ ; & & & \\
u^{c_1}_{x_1,x_4} = -0.1\ w_1 \ , & u^{c_1}_{\bar{x}_1,x_4} = 0.9\ w_1 \ , & u^{c_1}_{x_1,\bar{x}_4} = -0.1\ w_1 \ , & u^{c_1}_{\bar{x}_1,\bar{x}_4} = 0.9\ w_1 \ ; \\
u^{c_2}_{x_1,x_2} = -0.19\ w_2 \ , & u^{c_2}_{\bar{x}_1,x_2} = -0.55\ w_2 \ , & u^{c_2}_{x_1,\bar{x}_2} = 0.081\ w_2 \ , & u^{c_2}_{\bar{x}_1,\bar{x}_2} = 0.045\ w_2 \ ; \\
u^{c_3}_{x_2,x_3} = -0.19\ w_3 \ , & u^{c_3}_{\bar{x}_2,x_3} = -0.55\ w_3 \ , & u^{c_3}_{x_2,\bar{x}_3} = 0.081\ w_3 \ , & u^{c_3}_{\bar{x}_2,\bar{x}_3} = 0.045\ w_3 \ ; \\
u^{c_4}_{x_3,x_4} = -0.19\ w_4 \ , & u^{c_4}_{\bar{x}_3,x_4} = -0.55\ w_4 \ , & u^{c_4}_{x_3,\bar{x}_4} = 0.081\ w_4 \ , & u^{c_4}_{\bar{x}_3,\bar{x}_4} = 0.045\ w_4 \ .
\end{array}
$$

Using these new LP variables, our LP constraint from Equation (18) for the *reboot* $= 1$ action becomes:

$$
0 \geq \max_{X_1,X_2,X_3,X_4} \sum_{i=1}^{4} u^{R_i}_{X_i} + u^{c_0} + \sum_{j=1}^{4} u^{c_j}_{X_{j-1},X_j} \ .
$$

We are now ready for the variable elimination process. We illustrate the elimination of variable $X_4$:

$$
0 \geq \max_{X_1,X_2,X_3} \sum_{i=1}^{3} u^{R_i}_{X_i} + u^{c_0} + \sum_{j=2}^{3} u^{c_j}_{X_{j-1},X_j} + \max_{X_4} \left[ u^{R_4}_{X_4} + u^{c_1}_{X_1,X_4} + u^{c_4}_{X_3,X_4} \right] \ .
$$

We can represent the term $\max_{X_4} \left[ u^{R_4}_{X_4} + u^{c_1}_{X_1,X_4} + u^{c_4}_{X_3,X_4} \right]$ by a set of linear constraints, one for each assignment of $X_1$ and $X_3$, using the new LP variables $u^{e_1}_{X_1,X_3}$ to represent this maximum:

$$
\begin{array}{rcl}
u^{e_1}_{x_1,x_3} & \geq & u^{R_4}_{x_4} + u^{c_1}_{x_1,x_4} + u^{c_4}_{x_3,x_4} \ ; \\
u^{e_1}_{x_1,x_3} & \geq & u^{R_4}_{\bar{x}_4} + u^{c_1}_{x_1,\bar{x}_4} + u^{c_4}_{x_3,\bar{x}_4} \ ; \\
u^{e_1}_{\bar{x}_1,x_3} & \geq & u^{R_4}_{x_4} + u^{c_1}_{\bar{x}_1,x_4} + u^{c_4}_{x_3,x_4} \ ; \\
u^{e_1}_{\bar{x}_1,x_3} & \geq & u^{R_4}_{\bar{x}_4} + u^{c_1}_{\bar{x}_1,\bar{x}_4} + u^{c_4}_{x_3,\bar{x}_4} \ ; \\
u^{e_1}_{x_1,\bar{x}_3} & \geq & u^{R_4}_{x_4} + u^{c_1}_{x_1,x_4} + u^{c_4}_{\bar{x}_3,x_4} \ ; \\
u^{e_1}_{x_1,\bar{x}_3} & \geq & u^{R_4}_{\bar{x}_4} + u^{c_1}_{x_1,\bar{x}_4} + u^{c_4}_{\bar{x}_3,\bar{x}_4} \ ; \\
u^{e_1}_{\bar{x}_1,\bar{x}_3} & \geq & u^{R_4}_{x_4} + u^{c_1}_{\bar{x}_1,x_4} + u^{c_4}_{\bar{x}_3,x_4} \ ; \\
u^{e_1}_{\bar{x}_1,\bar{x}_3} & \geq & u^{R_4}_{\bar{x}_4} + u^{c_1}_{\bar{x}_1,\bar{x}_4} + u^{c_4}_{\bar{x}_3,\bar{x}_4} \ .
\end{array}
$$

We have now eliminated variable $X_4$ and our global non-linear constraint becomes:

$$
0 \geq \max_{X_1,X_2,X_3} \sum_{i=1}^{3} u^{R_i}_{X_i} + u^{c_0} + \sum_{j=2}^{3} u^{c_j}_{X_{j-1},X_j} + u^{e_1}_{X_1,X_3} \ .
$$

Next, we eliminate variable $X_3$. The new LP constraints and variables have the form:

$$
u^{e_2}_{X_1,X_2} \geq u^{R_3}_{X_3} + u^{c_3}_{X_2,X_3} + u^{e_1}_{X_1,X_3} \ , \ \forall\ X_1, X_2, X_3 \ ;
$$

thus, removing $X_3$ from the global non-linear constraint:

$$
0 \geq \max_{X_1,X_2} \sum_{i=1}^{2} u^{R_i}_{X_i} + u^{c_0} + u^{c_2}_{X_1,X_2} + u^{e_2}_{X_1,X_2} \ .
$$

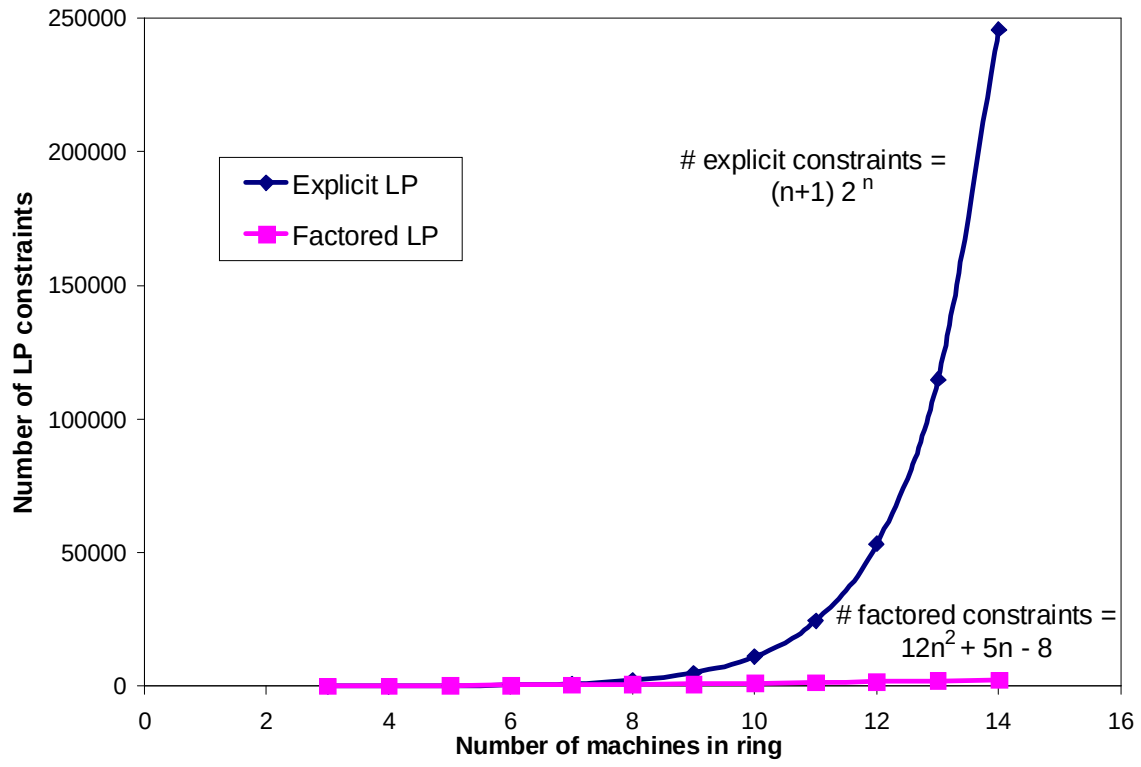

Figure 7: Number of constraints in the LP generated by the explicit state representation versus the factored LP construction for the solution of the ring problem with basis functions over single variables and approximate linear programming as the solution algorithm.

We can now eliminate $X_2$, generating the linear constraints:

$$u^{e_3}_{X_1} \geq u^{R_2}_{X_2} + u^{c_2}_{X_1,X_2} + u^{e_2}_{X_1,X_2} \ , \ \forall \ X_1, X_2 \ .$$

Now, our global non-linear constraint involves only $X_1$:

$$0 \geq \max_{X_1} u^{R_1}_{X_1} + u^{c_0} + u^{e_3}_{X_1} \ .$$

As $X_1$ is the last variable to be eliminated, the scope of the new LP variable is empty and the linear constraints are given by:

$$u^{e_4} \geq u^{R_1}_{X_1} + u^{e_3}_{X_1} \ , \ \forall \ X_1 \ .$$

All of the state variables have now been eliminated, turning our global non-linear constraint into a simple linear constraint:

$$0 \geq u^{c_0} + u^{e_4} \ ,$$

which completes the LP description for the approximate linear programming solution to the problem in Figure 2.

In this small example with only four state variables, our factored LP technique generates a total of 89 equality constraints, 115 inequality constraints and 149 LP variables, while the explicit state representation in Equation (8) generates only 80 inequality constraints and 5 LP variables. However, as the problem size increases, the number of constraints and LP variables in our factored LP approach grow as $O(n^2)$, while the explicit state approach grows exponentially, at $O(n2^n)$. This scaling effect is illustrated in Figure 7.

## 6. Approximate Policy Iteration with Max-norm Projection

The factored approximate linear programming approach described in the previous section is both elegant and easy to implement. However, we cannot, in general, provide strong guarantees about the error it achieves. An alternative is to use the approximate policy iteration described in Section 3.2, which does offer certain bounds on the error. However, as we shall see, this algorithm is significantly more complicated, and requires that we place additional restrictions on the factored MDP.

In particular, approximate policy iteration requires a representation of the policy at each iteration. In order to obtain a compact policy representation, we must make an additional assumption: each action only affects a small number of state variables. We first state this assumption formally. Then, we show how to obtain a compact representation of the greedy policy with respect to a factored value function, under this assumption. Finally, we describe our factored approximate policy iteration algorithm using max-norm projections.

### 6.1 Default Action Model

In Section 2.2, we presented the factored MDP model, where each action is associated with its own factored transition model represented as a DBN and with its own factored reward function. However, different actions often have very similar transition dynamics, only differing in their effect on some small set of variables. In particular, in many cases a variable has a default evolution model, which only changes if an action affects it directly (Boutilier *et al.*, 2000).

This type of structure turns out to be useful for compactly representing policies, a property which is important in our approximate policy iteration algorithm. Thus, in this section of the paper, we restrict attention to factored MDPs that are defined using a *default transition model* $\tau_d = \langle G_d, P_d \rangle$ (Koller & Parr, 2000). For each action $a$, we define $\mathit{Effects}[a] \subseteq \mathbf{X}'$ to be the variables in the next state whose local probability model is different from $\tau_d$, *i.e.*, those variables $X_i'$ such that $P_a(X_i' \mid \mathit{Parents}_a(X_i')) \neq P_d(X_i' \mid \mathit{Parents}_d(X_i'))$.

**Example 6.1** *In our system administrator example, we have an action $a_i$ for rebooting each one of the machines, and a default action $d$ for doing nothing. The transition model described above corresponds to the "do nothing" action, which is also the default transition model. The transition model for $a_i$ is different from $d$ only in the transition model for the variable $X_i'$, which is now $X_i' = \mathit{true}$ with probability one, regardless of the status of the neighboring machines. Thus, in this example, $\mathit{Effects}[a_i] = X_i'$.* $\square$

As in the transition dynamics, we can also define the notion of *default reward model.* In this case, there is a set of reward functions $\sum_{i=1}^{r} R_i(\mathbf{U}_i)$ associated with the default action $d$. In addition, each action $a$ can have a reward function $R^a(\mathbf{U}^a)$. Here, the extra reward of action $a$ has scope restricted to $\mathit{Rewards}[a] = \mathbf{U}_i^a \subset \{X_1, \ldots, X_n\}$. Thus, the total reward associated with action $a$ is given by $R^a + \sum_{i=1}^{r} R_i$. Note that $R^a$ can also be factored as a linear combination of smaller terms for an even more compact representation.

We can now build on this additional assumption to define the complete algorithm. Recall that the approximate policy iteration algorithm iterates through two steps: policy improvement and approximate value determination. We now discuss each of these steps.

### 6.2 Computing Greedy Policies

The policy improvement step computes the greedy policy relative to a value function $\mathcal{V}^{(t-1)}$:

$$\pi^{(t)} = \mathit{Greedy}(\mathcal{V}^{(t-1)}).$$

Recall that our value function estimates have the linear form $\mathrm{H}\mathbf{w}$. As we described in Section 4.1, the greedy policy for this type of value function is given by:

$$\mathit{Greedy}(\mathrm{H}\mathbf{w})(\mathbf{x}) = \arg\max_a Q_a(\mathbf{x}),$$

where each $Q_a$ can be represented by: $Q_a(\mathbf{x}) = R(\mathbf{x}, a) + \sum_i w_i \; g_i^a(\mathbf{x})$.

If we attempt to represent this policy naively, we are again faced with the problem of exponentially large state spaces. Fortunately, as shown by Koller and Parr (2000), the greedy policy relative to a factored value function has the form of a *decision list*. More precisely, the policy can be written in the form $\langle \mathbf{t}_1, a_1 \rangle, \langle \mathbf{t}_2, a_2 \rangle, \ldots, \langle \mathbf{t}_L, a_L \rangle$, where each $\mathbf{t}_i$ is an assignment of values to some small subset $\mathbf{T}_i$ of variables, and each $a_i$ is an action. The greedy action to take in state $\mathbf{x}$ is the action $a_j$ corresponding to the first event $\mathbf{t}_j$ in the list with which $\mathbf{x}$ is consistent. For completeness, we now review the construction of this decision-list policy.

The critical assumption that allows us to represent the policy as a compact decision list is the default action assumption described in Section 6.1. Under this assumption, the $Q_a$ functions can be written as:

$$Q_a(\mathbf{x}) = R^a(\mathbf{x}) + \sum_{i=1}^{r} R_i(\mathbf{x}) + \sum_i w_i \; g_i^a(\mathbf{x}),$$

where $R^a$ has scope restricted to $\mathbf{U}^a$. The $Q$ function for the default action $d$ is just: $Q_d(\mathbf{x}) = \sum_{i=1}^{r} R_i(\mathbf{x}) + \sum_i w_i \; g_i^d(\mathbf{x})$.

We now have a set of linear $Q$-functions which implicitly describes a policy $\pi$. It is not immediately obvious that these $Q$ functions result in a compactly expressible policy. An important insight is that most of the components in the weighted combination are identical, so that $g_i^a$ is equal to $g_i^d$ for most $i$. Intuitively, a component $g_i^a$ corresponding to the backprojection of basis function $h_i(\mathbf{C}_i)$ is only different if the action $a$ influences one of the variables in $\mathbf{C}_i$. More formally, assume that $\mathit{Effects}[a] \cap \mathbf{C}_i = \emptyset$. In this case, all of the variables in $\mathbf{C}_i$ have the same transition model in $\tau_a$ and $\tau_d$. Thus, we have that $g_i^a(\mathbf{x}) = g_i^d(\mathbf{x})$; in other words, the $i$th component of the $Q_a$ function is irrelevant when deciding whether action $a$ is better than the default action $d$. We can define which components are actually relevant: let $I_a$ be the set of indices $i$ such that $\mathit{Effects}[a] \cap \mathbf{C}_i \neq \emptyset$. These are the indices of those basis functions whose backprojection differs in $P_a$ and $P_d$. In our example DBN of Figure 2, actions and basis functions involve single variables, so $I_{a_i} = i$.

Let us now consider the impact of taking action $a$ over the default action $d$. We can define the impact — the difference in value — as:

$$\begin{aligned} \delta_a(\mathbf{x}) &= Q_a(\mathbf{x}) - Q_d(\mathbf{x}); \\ &= R^a(\mathbf{x}) + \sum_{i \in I_a} w_i \; \left[ g_i^a(\mathbf{x}) - g_i^d(\mathbf{x}) \right]. \end{aligned} \tag{19}$$

This analysis shows that $\delta_a(\mathbf{x})$ is a function whose scope is restricted to

$$\mathbf{T}_a = \mathbf{U}^a \cup \left[ \cup_{i \in I_a} \Gamma_a(\mathbf{C}_i') \right]. \tag{20}$$

DecisionListPolicy ($Q_a$)
*//$Q_a$ is the set of Q-functions, one for each action;*
*//Return the decision list policy $\Delta$.*
*//Initialize decision list.*
**Let** $\Delta = \{\}$.
*//Compute the bonus functions.*
**For** each action $a$, other than the default action $d$:
**Compute** the bonus for taking action $a$,

$$\delta_a(\mathbf{x}) = Q_a(\mathbf{x}) - Q_d(\mathbf{x});$$

as in Equation (19). Note that $\delta_a$ has scope restricted to $\mathbf{T}_a$, as in Equation (20).
*//Add states with positive bonuses to the (unsorted) decision list.*
**For** each assignment $\mathbf{t} \in \mathbf{T}_a$:
**If** $\delta_a(\mathbf{t}) > 0$, add branch to decision list:

$$\Delta = \Delta \cup \{\langle \mathbf{t}, a, \delta_a(\mathbf{t})\rangle\}.$$

*//Add the default action to the (unsorted) decision list.*
**Let** $\Delta = \Delta \cup \{\langle \emptyset, d, 0\rangle\}$.
*//Sort decision list to obtain final policy.*
**Sort** the decision list $\Delta$ in decreasing order on the $\delta$ element of $\langle \mathbf{t}, a, \delta\rangle$.
**Return** $\Delta$.

Figure 8: Method for computing the decision list policy $\Delta$ from the factored representation of the $Q_a$ functions.

In our example DBN, $\mathbf{T}_{a_2} = \{X_1, X_2\}$.

Intuitively, we now have a situation where we have a "baseline" value function $Q_d(\mathbf{x})$ which defines a value for each state $\mathbf{x}$. Each action $a$ changes that baseline by adding or subtracting an amount from each state. The point is that this amount depends only on $\mathbf{T}_a$, so that it is the same for all states in which the variables in $\mathbf{T}_a$ take the same values.

We can now define the greedy policy relative to our $Q$ functions. For each action $a$, define a set of *conditionals* $\langle \mathbf{t}, a, \delta\rangle$, where each $\mathbf{t}$ is some assignment of values to the variables $\mathbf{T}_a$, and $\delta$ is $\delta_a(\mathbf{t})$. Now, sort the conditionals for all of the actions by order of decreasing $\delta$:

$$\langle \mathbf{t}_1, a_1, \delta_1\rangle, \langle \mathbf{t}_2, a_2, \delta_2\rangle, \ldots, \langle \mathbf{t}_L, a_L, \delta_L\rangle.$$

Consider our optimal action in a state $\mathbf{x}$. We would like to get the largest possible "bonus" over the default value. If $\mathbf{x}$ is consistent with $\mathbf{t}_1$, we should clearly take action $a_1$, as it gives us bonus $\delta_1$. If not, then we should try to get $\delta_2$; thus, we should check if $\mathbf{x}$ is consistent with $\mathbf{t}_2$, and if so, take $a_2$. Using this procedure, we can compute the decision-list policy associated with our linear estimate of the value function. The complete algorithm for computing the decision list policy is summarized in Figure 8.

Note that the number of conditionals in the list is $\sum_a |\mathrm{Dom}(\mathbf{T}_a)|$; $\mathbf{T}_a$, in turn, depends on the set of basis function clusters that intersect with the effects of $a$. Thus, the size of the policy depends in a natural way on the interaction between the structure of our process description and the structure of our basis functions. In problems where the actions modify a large number of variables, the policy representation could become unwieldy. The approximate linear programming approach in Section 5 is more appropriate in such cases, as it does not require an explicit representation of the policy.

### 6.3 Value Determination

In the approximate value determination step our algorithm computes:

$$\mathbf{w}^{(t)} = \arg\min_{\mathbf{w}} \left\| \mathrm{H}\mathbf{w} - (R_{\pi^{(t)}} + \gamma P_{\pi^{(t)}} \mathrm{H}\mathbf{w}) \right\|_{\infty}.$$

By rearranging the expression, we get:

$$\mathbf{w}^{(t)} = \arg\min_{\mathbf{w}} \left\| (\mathrm{H} - \gamma P_{\pi^{(t)}} \mathrm{H})\,\mathbf{w} - R_{\pi^{(t)}} \right\|_{\infty}.$$

This equation is an instance of the optimization in Equation (4). If $P_{\pi^{(t)}}$ is factored, we can conclude that $C = (\mathrm{H} - \gamma P_{\pi^{(t)}} \mathrm{H})$ is also a matrix whose columns correspond to restricted-scope functions. More specifically:

$$c_i(\mathbf{x}) = h_i(\mathbf{x}) - \gamma g_i^{\pi^{(t)}}(\mathbf{x}),$$

where $g_i^{\pi^{(t)}}$ is the backprojection of the basis function $h_i$ through the transition model $P_{\pi^{(t)}}$, as described in Section 4.1. The target $\mathbf{b} = R_{\pi^{(t)}}$ corresponds to the reward function, which for the moment is assumed to be factored. Thus, we can again apply our factored LP in Section 4.2.3 to estimate the value of the policy $\pi^{(t)}$.

Unfortunately, the transition model $P_{\pi^{(t)}}$ is not factored, as a decision list representation for the policy $\pi^{(t)}$ will, in general, induce a transition model $P_{\pi^{(t)}}$ which cannot be represented by a compact DBN. Nonetheless, we can still generate a compact LP by exploiting the decision list structure of the policy. The basic idea is to introduce cost networks corresponding to each branch in the decision list, ensuring, additionally, that only states consistent with this branch are considered in the cost network maximization. Specifically, we have a factored LP construction for each branch $\langle \mathbf{t}_i, a_i \rangle$. The $i$th cost network only considers a subset of the states that is consistent with the $i$th branch of the decision list. Let $S_i$ be the set of states $\mathbf{x}$ such that $\mathbf{t}_i$ is the first event in the decision list for which $\mathbf{x}$ is consistent. That is, for each state $\mathbf{x} \in S_i$, $\mathbf{x}$ is consistent with $\mathbf{t}_i$, but it is *not* consistent with any $\mathbf{t}_j$ with $j < i$.

Recall that, as in Equation (11), our LP construction defines a set of constraints that imply that $\phi \geq \sum_i w_i \; c_i(\mathbf{x}) - b(\mathbf{x})$ for each state $\mathbf{x}$. Instead, we have a separate set of constraints for the states in each subset $S_i$. For each state in $S_i$, we know that action $a_i$ is taken. Hence, we can apply our construction above using $P_{a_i}$ — a transition model which is factored by assumption — in place of the non-factored $P_{\pi^{(t)}}$. Similarly, the reward function becomes $R^{a_i}(\mathbf{x}) + \sum_{i=1}^{r} R_i(\mathbf{x})$ for this subset of states.

The only issue is to guarantee that the cost network constraints derived from this transition model are applied only to states in $S_i$. Specifically, we must guarantee that they are applied only to states consistent with $\mathbf{t}_i$, but not to states that are consistent with some $\mathbf{t}_j$ for $j < i$. To guarantee the first condition, we simply instantiate the variables in $\mathbf{T}_i$ to take the values specified in $\mathbf{t}_i$. That is, our cost network now considers only the variables in

FactoredAPI ($P$, $R$, $\gamma$, $H$, $\mathcal{O}$, $\varepsilon$, $t_{max}$)

*//$P$ is the factored transition model.*
*//$R$ is the set of factored reward functions.*
*//$\gamma$ is the discount factor.*
*//$H$ is the set of basis functions $H = \{h_1, \ldots, h_k\}$.*
*//$\mathcal{O}$ stores the elimination order.*
*//$\varepsilon$ Bellman error precision.*
*//$t_{max}$ maximum number of iterations.*
*//Return the basis function weights $\mathbf{w}$ computed by approximate policy iteration.*

*//Initialize weights*
**Let** $\mathbf{w}^{(0)} = \mathbf{0}$.
*//Cache the backprojections of the basis functions.*
**For** each basis function $h_i \in H$; for each action $a$:
  **Let** $g_i^a = Backproj_a(h_i)$.
*//Main approximate policy iteration loop.*
**Let** $t = 0$.
**Repeat**
  ***//Policy improvement part of the loop.***
    *//Compute decision list policy for iteration $t$ weights.*
    **Let** $\Delta^{(t)} = \text{DecisionListPolicy}(R^a + \gamma \sum_i w_i^{(t)} g_i^a)$.
  ***//Value determination part of the loop.***
    *//Initialize constraints for max-norm projection LP.*
    **Let** $\Omega^+ = \{\}$ and $\Omega^- = \{\}$.
    *//Initialize indicators.*
    **Let** $\mathcal{I} = \{\}$.
    *//For every branch of the decision list policy, generate the relevant set of constraints, and update the indicators to constraint the state space for future branches.*
    **For** each branch $\langle \mathbf{t}_j, a_j \rangle$ in the decision list policy $\Delta^{(t)}$:
      *//Instantiate the variables in $\mathbf{T}_j$ to the assignment given in $\mathbf{t}_j$.*
      **Instantiate** the set of functions $\{h_1 - \gamma g_1^{a_j}, \ldots, h_k - \gamma g_k^{a_j}\}$ with the partial state assignment $\mathbf{t}_j$ and store in $C$.
      **Instantiate** the target functions $R^{a_j}$ with the partial state assignment $\mathbf{t}_j$ and store in $\mathbf{b}$.
      **Instantiate** the indicator functions $\mathcal{I}$ with the partial state assignment $\mathbf{t}_j$ and store in $\mathcal{I}'$.
      *//Generate the factored LP constraints for the current decision list branch.*
      **Let** $\Omega^+ = \Omega^+ \cup \text{FactoredLP}(C, -\mathbf{b} + \mathcal{I}', \mathcal{O})$.
      **Let** $\Omega^- = \Omega^- \cup \text{FactoredLP}(-C, \mathbf{b} + \mathcal{I}', \mathcal{O})$.
      *//Update the indicator functions.*
      **Let** $\mathcal{I}_j(\mathbf{x}) = -\infty \mathbb{1}(\mathbf{x} = \mathbf{t}_j)$ and update the indicators $\mathcal{I} = \mathcal{I} \cup \mathcal{I}_j$.
    *//We can now obtain the new set of weights by solving an LP, which corresponds to the max-norm projection.*
    **Let** $\mathbf{w}^{(t+1)}$ be the solution of the linear program: minimize $\phi$, subject to the constraints $\{\Omega^+, \Omega^-\}$.
    **Let** $t = t + 1$.
**Until** $\text{BellmanErr}(\text{H}\mathbf{w}^{(t)}) \leq \varepsilon$ or $t \geq t_{max}$ or $\mathbf{w}^{(t-1)} = \mathbf{w}^{(t)}$.
**Return** $\mathbf{w}^{(t)}$.

Figure 9: Factored approximate policy iteration with max-norm projection algorithm.

$\{X_1, \ldots, X_n\} - \mathbf{T}_i$, and computes the maximum only over the states consistent with $\mathbf{T}_i = \mathbf{t}_i$. To guarantee the second condition, we ensure that we do not impose any constraints on states associated with previous decisions. This is achieved by adding indicators $\mathcal{I}_j$ for each previous decision $\mathbf{t}_j$, with weight $-\infty$. More specifically, $\mathcal{I}_j$ is a function that takes value $-\infty$ for states consistent with $\mathbf{t}_j$ and zero for other all assignments of $\mathbf{T}_j$. The constraints for the $i$th branch will be of the form:

$$\phi \geq R(\mathbf{x}, a_i) + \sum_l w_l \left(\gamma g_l(\mathbf{x}, a_i) - h(\mathbf{x})\right) + \sum_{j<i} -\infty \mathbb{1}(\mathbf{x} = \mathbf{t}_j), \qquad \forall \mathbf{x} \sim [\mathbf{t}_i], \tag{21}$$

where $\mathbf{x} \sim [\mathbf{t}_i]$ defines the assignments of $\mathbf{X}$ consistent with $\mathbf{t}_i$. The introduction of these indicators causes the constraints associated with $\mathbf{t}_i$ to be trivially satisfied by states in $S_j$ for $j < i$. Note that each of these indicators is a restricted-scope function of $\mathbf{T}_j$ and can be handled in the same fashion as all other terms in the factored LP. Thus, for a decision list of size $L$, our factored LP contains constraints from $2L$ cost networks. The complete approximate policy iteration with max-norm projection algorithm is outlined in Figure 9.

### 6.4 Comparisons

It is instructive to compare our max-norm policy iteration algorithm to the $\mathcal{L}_2$-projection policy iteration algorithm of Koller and Parr (2000) in terms of computational costs per iteration and implementation complexity. Computing the $\mathcal{L}_2$ projection requires (among other things) a series of dot product operations between basis functions and backprojected basis functions $\langle h_i \bullet g_j^\pi \rangle$. These expressions are easy to compute if $P_\pi$ refers to the transition model of a particular action $a$. However, if the policy $\pi$ is represented as a decision list, as is the result of the policy improvement step, then this step becomes much more complicated. In particular, for every branch of the decision list, for every pair of basis functions $i$ and $j$, and for each assignment to the variables in $\text{Scope}[h_i] \cup \text{Scope}[g_j^a]$, it requires the solution of a counting problem which is $\sharp P$-complete in general. Although Koller and Parr show that this computation can be performed using a Bayesian network (BN) inference, the algorithm still requires a BN inference for each one of those assignments at each branch of the decision list. This makes the algorithm very difficult to implement efficiently in practice.

The max-norm projection, on the other hand, relies on solving a linear program at every iteration. The size of the linear program depends on the cost networks generated. As we discuss, two cost networks are needed for each point in the decision list. The complexity of each of these cost networks is approximately the same as only one of the BN inferences in the counting problem for the $\mathcal{L}_2$ projection. Overall, for each branch in the decision list, we have a total of two of these "inferences," as opposed to one for each assignment of $\text{Scope}[h_i] \cup \text{Scope}[g_j^a]$ for every pair of basis functions $i$ and $j$. Thus, the max-norm policy iteration algorithm is substantially less complex computationally than the approach based on $\mathcal{L}_2$-projection. Furthermore, the use of linear programming allows us to rely on existing LP packages (such as CPLEX), which are very highly optimized.

It is also interesting to compare the approximate policy iteration algorithm to the approximate linear programming algorithm we presented in Section 5. In the approximate linear programming algorithm, we never need to compute the decision list policy. The policy is always represented implicitly by the $Q_a$ functions. Thus, this algorithm does not require explicit computation or manipulation of the greedy policy. This difference has two important consequences: one computational and the other in terms of generality.

First, not having to compute or consider the decision lists makes approximate linear programming faster and easier to implement. In this algorithm, we generate a single LP with one cost network for each action and never need to compute a decision list policy. On the other hand, in each iteration, approximate policy iteration needs to generate two LPs for every branch of the decision list of size $L$, which is usually significantly longer than $|A|$, with a total of $2L$ cost networks. In terms of representation, we do not require the policies to be compact; thus, we do not need to make the default action assumption. Therefore, the approximate linear programming algorithm can deal with a more general class of problems, where each action can have its own independent DBN transition model. On the other hand, as described in Section 3.2, approximate policy iteration has stronger guarantees in terms of error bounds. These differences will be highlighted further in our experimental results presented in Section 9.

## 7. Computing Bounds on Policy Quality

We have presented two algorithms for computing approximate solutions to factored MDPs. All these algorithms generate linear value functions which can be denoted by $\mathrm{H}\widehat{\mathbf{w}}$, where $\widehat{\mathbf{w}}$ are the resulting basis function weights. In practice, the agent will define its behavior by acting according to the greedy policy $\widehat{\pi} = \mathit{Greedy}(\mathrm{H}\widehat{\mathbf{w}})$. One issue that remains is how this policy $\widehat{\pi}$ compares to the true optimal policy $\pi^*$; that is, how the *actual* value $\mathcal{V}_{\widehat{\pi}}$ of policy $\widehat{\pi}$ compares to $\mathcal{V}^*$.

In Section 3, we showed some *a priori* bounds for the quality of the policy. Another possible procedure is to compute an *a posteriori* bound. That is, given our resulting weights $\widehat{\mathbf{w}}$, we compute a bound on the loss of acting according to the greedy policy $\widehat{\pi}$ rather than the optimal policy. This can be achieved by using the *Bellman error* analysis of Williams and Baird (1993).

The *Bellman error* is defined as $\mathrm{BellmanErr}(\mathcal{V}) = \|\mathcal{T}^*\mathcal{V} - \mathcal{V}\|_\infty$. Given the greedy policy $\widehat{\pi} = \mathit{Greedy}(\mathcal{V})$, their analysis provides the bound:

$$\|\mathcal{V}^* - \mathcal{V}_{\widehat{\pi}}\|_\infty \leq \frac{2\gamma \mathrm{BellmanErr}(\mathcal{V})}{1-\gamma}. \tag{22}$$

Thus, we can use the Bellman error $\mathrm{BellmanErr}(\mathrm{H}\widehat{\mathbf{w}})$ to evaluate the quality of our resulting greedy policy.

Note that computing the Bellman error involves a maximization over the state space. Thus, the complexity of this computation grows exponentially with the number of state variables. Koller and Parr (2000) suggested that structure in the factored MDP can be exploited to compute the Bellman error efficiently. Here, we show how this error bound can be computed by a set of cost networks using a similar construction to the one in our max-norm projection algorithms. This technique can be used for any $\widehat{\pi}$ that can be represented as a decision list and does not depend on the algorithm used to determine the policy. Thus, we can apply this technique to solutions determined approximate linear programming if the action descriptions permit a decision list representation of the policy.

For some set of weights $\widehat{\mathbf{w}}$, the Bellman error is given by:

FactoredBellmanErr $(P,\ R,\ \gamma,\ H,\ \mathcal{O},\ \widehat{\mathbf{w}})$

*//$P$ is the factored transition model.*
*//$R$ is the set of factored reward functions.*
*//$\gamma$ is the discount factor.*
*//$H$ is the set of basis functions $H = \{h_1, \ldots, h_k\}$.*
*//$\mathcal{O}$ stores the elimination order.*
*//$\widehat{\mathbf{w}}$ are the weights for the linear value function.*
*//Return the Bellman error for the value function $\mathrm{H}\widehat{\mathbf{w}}$.*

*//Cache the backprojections of the basis functions.*
**For** each basis function $h_i \in H$; for each action $a$:
  **Let** $g_i^a = Backproj_a(h_i)$.
*//Compute decision list policy for value function $\mathrm{H}\widehat{\mathbf{w}}$.*
**Let** $\widehat{\Delta} = \text{DecisionListPolicy}(R^a + \gamma \sum_i \widehat{w}_i g_i^a)$.
*//Initialize indicators.*
**Let** $\mathcal{I} = \{\}$.
*//Initialize Bellman error.*
**Let** $\varepsilon = 0$.
*//For every branch of the decision list policy, generate the relevant cost networks, solve it with variable elimination, and update the indicators to constraint the state space for future branches.*
**For** each branch $\langle \mathbf{t}_j, a_j \rangle$ in the decision list policy $\widehat{\Delta}$:
  *//Instantiate the variables in $\mathbf{T}_j$ to the assignment given in $\mathbf{t}_j$.*
  **Instantiate** the set of functions $\{\widehat{w}_1(h_1 - \gamma g_1^{a_j}), \ldots, \widehat{w}_k(h_k - \gamma g_k^{a_j})\}$ with the partial state assignment $\mathbf{t}_j$ and store in $C$.
  **Instantiate** the target functions $R^{a_j}$ with the partial state assignment $\mathbf{t}_j$ and store in $\mathbf{b}$.
  **Instantiate** the indicator functions $\mathcal{I}$ with the partial state assignment $\mathbf{t}_j$ and store in $\mathcal{I}'$.
  *//Use variable elimination to solve first cost network, and update Bellman error, if error for this branch is larger.*
  **Let** $\varepsilon = \max(\varepsilon, \text{VariableElimination}(C - \mathbf{b} + \mathcal{I}', \mathcal{O}))$.
  *//Use variable elimination to solve second cost network, and update Bellman error, if error for this branch is larger.*
  **Let** $\varepsilon = \max(\varepsilon, \text{VariableElimination}(-C + \mathbf{b} + \mathcal{I}', \mathcal{O}))$.
  *//Update the indicator functions.*
  **Let** $\mathcal{I}_j(\mathbf{x}) = -\infty \mathbb{1}(\mathbf{x} = \mathbf{t}_j)$ and update the indicators $\mathcal{I} = \mathcal{I} \cup \mathcal{I}_j$.
**Return** $\varepsilon$.

Figure 10: Algorithm for computing Bellman error for factored value function $\mathrm{H}\widehat{\mathbf{w}}$.

$$\begin{aligned}\text{BellmanErr}(\mathrm{H}\widehat{\mathbf{w}}) &= \left\|\mathcal{T}^*\mathrm{H}\widehat{\mathbf{w}} - \mathrm{H}\widehat{\mathbf{w}}\right\|_\infty ; \\ &= \max\begin{pmatrix} \max_{\mathbf{x}} \sum_i w_i h_i(\mathbf{x}) - R_{\widehat{\pi}}(\mathbf{x}) - \gamma \sum_{\mathbf{x}'} P_{\widehat{\pi}}(\mathbf{x}' \mid \mathbf{x}) \sum_j w_j h_j(\mathbf{x}') \ , \\ \max_{\mathbf{x}} R_{\widehat{\pi}}(\mathbf{x}) + \gamma \sum_{\mathbf{x}'} P_{\widehat{\pi}}(\mathbf{x}' \mid \mathbf{x}) \sum_j w_j h_j(\mathbf{x}') - \sum_i w_i h_i(\mathbf{x}) \end{pmatrix}.\end{aligned}$$

If the rewards $R_{\widehat{\pi}}$ and the transition model $P_{\widehat{\pi}}$ are factored appropriately, then we can compute each one of these two maximizations ($\max_{\mathbf{x}}$) using variable elimination in a cost network as described in Section 4.2.1. However, $\widehat{\pi}$ is a decision list policy and it does not induce a factored transition model. Fortunately, as in the approximate policy iteration algorithm in Section 6, we can exploit the structure in the decision list to perform such maximization efficiently. In particular, as in approximate policy iteration, we will generate two cost networks for each branch in the decision list. To guarantee that our maximization is performed only over states where this branch is relevant, we include the same type of indicator functions, which will force irrelevant states to have a value of $-\infty$, thus guaranteeing that at each point of the decision list policy we obtain the corresponding state with the maximum error. The state with the overall largest Bellman error will be the maximum over the ones generated for each point the in the decision list policy. The complete factored algorithm for computing the Bellman error is outlined in Figure 10.

One last interesting note concerns our approximate policy iteration algorithm with max-norm projection of Section 6. In all our experiments, this algorithm converged, so that $\mathbf{w}^{(t)} = \mathbf{w}^{(t+1)}$ after some iterations. If such convergence occurs, then the objective function $\phi^{(t+1)}$ of the linear program in our last iteration is equal to the Bellman error of the final policy:

**Lemma 7.1** *If approximate policy iteration with max-norm projection converges, so that* $\mathbf{w}^{(t)} = \mathbf{w}^{(t+1)}$ *for some iteration* $t$*, then the max-norm projection error* $\phi^{(t+1)}$ *of the last iteration is equal to the Bellman error for the final value function estimate* $\mathrm{H}\widehat{\mathbf{w}} = \mathrm{H}\mathbf{w}^{(t)}$*:*

$$\text{BellmanErr}(\mathrm{H}\widehat{\mathbf{w}}) = \phi^{(t+1)}.$$

**Proof:** See Appendix A.4. □

Thus, we can bound the loss of acting according to the final policy $\pi^{(t+1)}$ by substituting $\phi^{(t+1)}$ into the Bellman error bound:

**Corollary 7.2** *If approximate policy iteration with max-norm projection converges after* $t$ *iterations to a final value function estimate* $\mathrm{H}\widehat{\mathbf{w}}$ *associated with a greedy policy* $\widehat{\pi} = \text{Greedy}(\mathrm{H}\widehat{\mathbf{w}})$*, then the loss of acting according to* $\widehat{\pi}$ *instead of the optimal policy* $\pi^*$ *is bounded by:*

$$\left\|\mathcal{V}^* - \mathcal{V}_{\widehat{\pi}}\right\|_\infty \leq \frac{2\gamma\phi^{(t+1)}}{1-\gamma},$$

*where* $\mathcal{V}_{\widehat{\pi}}$ *is the* actual *value of the policy* $\widehat{\pi}$*.* □

Therefore, when approximate policy iteration converges we can obtain a bound on the quality of the resulting policy without needing to compute the Bellman error explicitly.

## 8. Exploiting Context-specific Structure

Thus far, we have presented a suite of algorithms which exploit additive structure in the reward and basis functions and sparse connectivity in the DBN representing the transition model. However, there exists another important type of structure that should also be exploited for efficient decision making: *context-specific independence* (CSI). For example, consider an agent responsible for building and maintaining a house, if the painting task can only be completed after the plumbing and the electrical wiring have been installed, then the probability that the painting is done is 0, in all contexts where plumbing or electricity are not done, *independently* of the agents action. The representation we have used so far in this paper would use a table to represent this type of function. This table is exponentially large in the number of variables in the scope of the function, and ignores the context-specific structure inherent in the problem definition.

Boutilier *et al.* (Boutilier *et al.*, 1995; Dearden & Boutilier, 1997; Boutilier, Dean, & Hanks, 1999; Boutilier *et al.*, 2000) have developed a set of algorithms which can exploit CSI in the transition and reward models to perform efficient (approximate) planning. Although this approach is often successful in problems where the *value function* contains sufficient context-specific structure, the approach is not able to exploit the additive structure which is also often present in real-world problems.

In this section, we extend the factored MDP model to include context-specific structure. We present a simple, yet effective extension of our algorithms which can exploit both CSI and additive structure to obtain efficient approximations for factored MDPs. We first extend the factored MDP representation to include context-specific structure and then show how the basic operations from Section 4 required by our algorithms can be performed efficiently in this new representation.

### 8.1 Factored MDPs with Context-specific and Additive Structure

There are several representations for context-specific functions. The most common are decision trees (Boutilier *et al.*, 1995), algebraic decision diagrams (ADDs) (Hoey, St-Aubin, Hu, & Boutilier, 1999), and rules (Zhang & Poole, 1999). We choose to use rules as our basic representation, for two main reasons. First, the rule-based representation allows a fairly simple algorithm for variable elimination, which is a key operation in our framework. Second, rules are not required to be mutually exclusive and exhaustive, a requirement that can be restrictive if we want to exploit additive independence, where functions can be represented as a linear combination of a set of non-mutually exclusive functions.

We begin by describing the rule-based representation (along the lines of Zhang and Poole's presentation (1999)) for the probabilistic transition model, in particular, the CPDs of our DBN model. Roughly speaking, each rule corresponds to some set of CPD entries that are all associated with a particular probability value. These entries with the same value are referred to as *consistent* contexts:

**Definition 8.1** *Let* $\mathbf{C} \subseteq \{\mathbf{X}, \mathbf{X}'\}$ *and* $\mathbf{c} \in \mathrm{Dom}(\mathbf{C})$. *We say that* $\mathbf{c}$ *is* consistent *with* $\mathbf{b} \in \mathrm{Dom}(\mathbf{B})$*, for* $\mathbf{B} \subseteq \{\mathbf{X}, \mathbf{X}'\}$*, if* $\mathbf{c}$ *and* $\mathbf{b}$ *have the same assignment for the variables in* $\mathbf{C} \cap \mathbf{B}$. $\square$

The probability of these consistent contexts will be represented by *probability rules*:

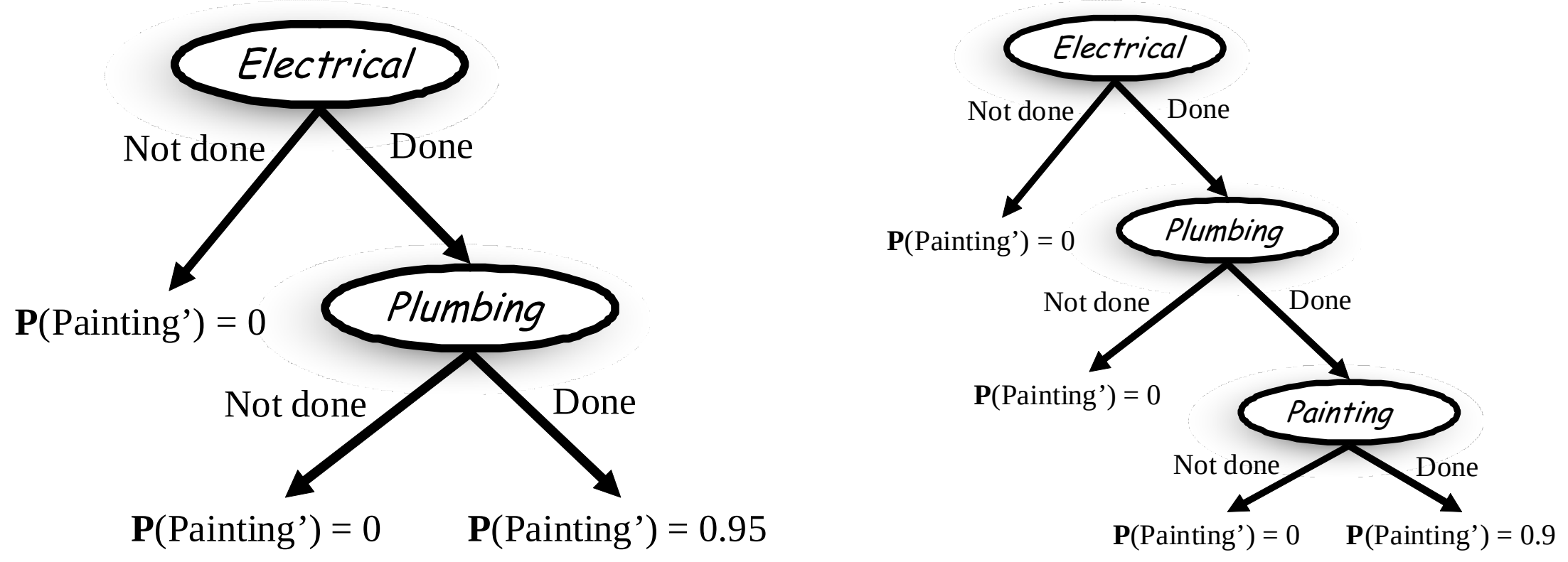

**(a)** **(b)**

$$\eta_1 = \langle \neg\text{Electrical} : 0 \rangle$$
$$\eta_2 = \langle \text{Electrical} \wedge \neg \text{ Plumbing} : 0 \rangle$$
$$\eta_3 = \langle \text{Electrical} \wedge \text{Plumbing} : 0.95 \rangle$$

**(c)**

$$\eta_4 = \langle \neg\text{Electrical} : 0 \rangle$$
$$\eta_5 = \langle \text{Electrical} \wedge \neg\text{Plumbing} : 0 \rangle$$
$$\eta_6 = \langle \text{Electrical} \wedge \text{Plumbing} \wedge \neg\text{Painting} : 0 \rangle$$
$$\eta_7 = \langle \text{Electrical} \wedge \text{Plumbing} \wedge \text{Painting} : 0.9 \rangle$$

**(d)**

Figure 11: Example CPDs for variable the *Painting'* = *true* represented as decision trees: (a) when the action is paint; (b) when the action is not paint. The same CPDs can be represented by probability rules as shown in (c) and (d), respectively.

**Definition 8.2** *A* probability rule $\eta = \langle \mathbf{c} : p \rangle$ *is a function* $\eta : \{\mathbf{X}, \mathbf{X}'\} \mapsto [0, 1]$, *where the* context $\mathbf{c} \in \text{Dom}(\mathbf{C})$ *for* $\mathbf{C} \subseteq \{\mathbf{X}, \mathbf{X}'\}$ *and* $p \in [0, 1]$, *such that* $\eta(\mathbf{x}, \mathbf{x}') = p$ *if* $(\mathbf{x}, \mathbf{x}')$ *is consistent with* $\mathbf{c}$ *and is equal to* 1 *otherwise.* □

In this case, it is convenient to require that the rules be mutually exclusive and exhaustive, so that each CPD entry is uniquely defined by its association with a single rule.

**Definition 8.3** *A* rule-based conditional probability distribution *(rule CPD)* $P_a$ *is a function* $P_a : (\{X_i'\} \cup \mathbf{X}) \mapsto [0, 1]$, *composed of a set of probability rules* $\{\eta_1, \eta_2, \ldots, \eta_m\}$ *whose contexts are mutually exclusive and exhaustive. We define:*

$$P_a(x_i' \mid \mathbf{x}) = \eta_j(\mathbf{x}, \mathbf{x}'),$$

*where* $\eta_j$ *is the unique rule in* $P_a$ *for which* $\mathbf{c}_j$ *is consistent with* $(x_i', \mathbf{x})$. *We require that, for all* $\mathbf{x}$,

$$\sum_{x_i'} P_a(x_i' \mid \mathbf{x}) = 1. \quad \square$$

We can define $Parents_a(X_i')$ to be the union of the contexts of the rules in $P_a(X_i' \mid \mathbf{X})$. An example of a CPD represented by a set of probability rules is shown in Figure 11.

Rules can also be used to represent additive functions, such as reward or basis functions. We represent such context specific value dependencies using *value rules*:

**Definition 8.4** *A* value rule $\rho = \langle \mathbf{c} : v \rangle$ *is a function* $\rho : \mathbf{X} \mapsto \mathbb{R}$ *such that* $\rho(\mathbf{x}) = v$ *when* $\mathbf{x}$ *is consistent with* $\mathbf{c}$ *and* $0$ *otherwise.* □

Note that a value rule $\langle \mathbf{c} : v \rangle$ has a scope $\mathbf{C}$.

It is important to note that value rules are not required to be mutually exclusive and exhaustive. Each value rule represents a (weighted) indicator function, which takes on a value $v$ in states consistent with some context $\mathbf{c}$, and 0 in all other states. In any given state, the values of the zero or more rules consistent with that state are simply added together.

**Example 8.5** *In our construction example, we might have a set of rules:*

$$\begin{aligned} \rho_1 &= \langle \mathit{Plumbing} = \text{done} : 100 \rangle; \\ \rho_2 &= \langle \mathit{Electricity} = \text{done} : 100 \rangle; \\ \rho_3 &= \langle \mathit{Painting} = \text{done} : 100 \rangle; \\ \rho_4 &= \langle \mathit{Action} = \text{plumb} : -10 \rangle; \\ &\vdots \end{aligned}$$

*which, when summed together, define the reward function* $R = \rho_1 + \rho_2 + \rho_3 + \rho_4 + \cdots$. □

In general, our reward function $R^a$ is represented as a *rule-based function*:

**Definition 8.6** *A* rule-based function $f : \mathbf{X} \mapsto \mathbb{R}$ *is composed of a set of rules* $\{\rho_1, \ldots, \rho_n\}$ *such that* $f(\mathbf{x}) = \sum_{i=1}^{n} \rho_i(\mathbf{x})$. □

In the same manner, each one of our basis functions $h_j$ is now represented as a rule-based function.

This notion of a rule-based function is related to the tree-structure functions used by Boutilier *et al.* (2000), but is substantially more general. In the tree-structure value functions, the rules corresponding to the different leaves are mutually exclusive and exhaustive. Thus, the total number of different values represented in the tree is equal to the number of leaves (or rules). In the rule-based function representation, the rules are not mutually exclusive, and their values are added to form the overall function value for different settings of the variables. Different rules are added in different settings, and, in fact, with $k$ rules, one can easily generate $2^k$ different possible values, as is demonstrated in Section 9. Thus, the rule-based functions can provide a compact representation for a much richer class of value functions.

Using this rule-based representation, we can exploit both CSI and additive independence in the representation of our factored MDP and basis functions. We now show how the basic operations in Section 4 can be adapted to exploit our rule-based representation.

## 8.2 Adding, Multiplying and Maximizing Consistent Rules

In our table-based algorithms, we relied on standard sum and product operators applied to tables. In order to exploit CSI using a rule-based representation, we must redefine these standard operations. In particular, the algorithms will need to add or multiply rules that ascribe values to overlapping sets of states.

We will start by defining these operations for rules with the same context:

**Definition 8.7** *Let* $\rho_1 = \langle \mathbf{c} : v_1 \rangle$ *and* $\rho_2 = \langle \mathbf{c} : v_2 \rangle$ *be two rules with context* $\mathbf{c}$. *Define the* rule product *as* $\rho_1 \times \rho_2 = \langle \mathbf{c} : v_1 \cdot v_2 \rangle$, *and the* rule sum *as* $\rho_1 + \rho_2 = \langle \mathbf{c} : v_1 + v_2 \rangle$. □

Note that this definition is restricted to rules with the same context. We will address this issue in a moment. First, we will introduce an additional operation which maximizes a variable from a set of rules, which otherwise share a common context:

**Definition 8.8** *Let* $Y$ *be a variable with* $\mathrm{Dom}[Y] = \{y_1, \ldots, y_k\}$, *and let* $\rho_i$, *for each* $i = 1, \ldots, k$, *be a rule of the form* $\rho_i = \langle \mathbf{c} \wedge Y = y_i : v_i \rangle$. *Then for the rule-based function* $f = \rho_1 + \cdots + \rho_k$, *define the* rule maximization *over* $Y$ *as* $\max_Y f = \langle \mathbf{c} : \max_i v_i \rangle$ . □

After this operation, $Y$ has been maximized out from the scope of the function $f$.

These three operations we have just described can only be applied to sets of rules that satisfy very stringent conditions. To make our set of rules amenable to the application of these operations, we might need to refine some of these rules. We therefore define the following operation:

**Definition 8.9** *Let* $\rho = \langle \mathbf{c} : v \rangle$ *be a rule, and* $Y$ *be a variable. Define the* rule split *Split*$(\rho \angle Y)$ *of* $\rho$ *on a variable* $Y$ *as follows: If* $Y \in \mathrm{Scope}[\mathbf{C}]$, *then* *Split*$(\rho \angle Y) = \{\rho\}$; *otherwise,*

$$\mathit{Split}(\rho \angle Y) = \{\langle \mathbf{c} \wedge Y = y_i : v \rangle \mid y_i \in \mathrm{Dom}[Y]\} \ . \ \square$$

Thus, if we split a rule $\rho$ on variable $Y$ that is not in the scope of the context of $\rho$, then we generate a new set of rules, with one for each assignment in the domain of $Y$.

In general, the purpose of rule splitting is to extend the context $\mathbf{c}$ of one rule $\rho$ coincide with the context $\mathbf{c}'$ of another consistent rule $\rho'$. Naively, we might take all variables in $\mathrm{Scope}[\mathbf{C}'] - \mathrm{Scope}[\mathbf{C}]$ and split $\rho$ recursively on each one of them. However, this process creates unnecessarily many rules: If $Y$ is a variable in $\mathrm{Scope}[\mathbf{C}'] - \mathrm{Scope}[\mathbf{C}]$ and we split $\rho$ on $Y$, then only one of the $|\mathrm{Dom}[Y]|$ new rules generated will remain consistent with $\rho'$: the one which has the same assignment for $Y$ as the one in $\mathbf{c}'$. Thus, only this consistent rule needs to be split further. We can now define the recursive splitting procedure that achieves this more parsimonious representation:

**Definition 8.10** *Let* $\rho = \langle \mathbf{c} : v \rangle$ *be a rule, and* $\mathbf{b}$ *be a context such that* $\mathbf{b} \in \mathrm{Dom}[\mathbf{B}]$. *Define the* recursive rule split *Split*$(\rho \angle \mathbf{b})$ *of* $\rho$ *on a context* $\mathbf{b}$ *as follows:*

1. $\{\rho\}$, *if* $\mathbf{c}$ *is not consistent with* $\mathbf{b}$; *else,*
2. $\{\rho\}$, *if* $\mathrm{Scope}[\mathbf{B}] \subseteq \mathrm{Scope}[\mathbf{C}]$; *else,*
3. $\{\mathit{Split}(\rho_i \angle \mathbf{b}) \mid \rho_i \in \mathit{Split}(\rho \angle Y)\}$, *for some variable* $Y \in \mathrm{Scope}[\mathbf{B}] - \mathrm{Scope}[\mathbf{C}]$ .

□

In this definition, each variable $Y \in \mathrm{Scope}[\mathbf{B}] - \mathrm{Scope}[\mathbf{C}]$ leads to the generation of $k = |\mathrm{Dom}(Y)|$ rules at the step in which it is split. However, only one of these $k$ rules is used in the next recursive step because only one is consistent with $\mathbf{b}$. Therefore, the size of the split set is simply $1 + \sum_{Y \in \mathrm{Scope}[\mathbf{B}] - \mathrm{Scope}[\mathbf{C}]} (|Dom(Y)| - 1)$. This size is independent of the order in which the variables are split within the operation.

Note that only one of the rules in $Split(\rho\angle\mathbf{b})$ is consistent with $\mathbf{b}$: the one with context $\mathbf{c} \wedge \mathbf{b}$. Thus, if we want to add two consistent rules $\rho_1 = \langle \mathbf{c}_1 : v_1 \rangle$ and $\rho_2 = \langle \mathbf{c}_2 : v_2 \rangle$, then all we need to do is replace these rules by the set:

$$Split(\rho_1\angle\mathbf{c}_2) \cup Split(\rho_2\angle\mathbf{c}_1),$$

and then simply replace the resulting rules $\langle \mathbf{c}_1 \wedge \mathbf{c}_2 : v_1 \rangle$ and $\langle \mathbf{c}_2 \wedge \mathbf{c}_1 : v_2 \rangle$ by their sum $\langle \mathbf{c}_1 \wedge \mathbf{c}_2 : v_1 + v_2 \rangle$. Multiplication is performed in an analogous manner.

**Example 8.11** *Consider adding the following set of consistent rules:*

$$\begin{aligned} \rho_1 &= \langle a \wedge b : 5 \rangle, \\ \rho_2 &= \langle a \wedge \neg c \wedge d : 3 \rangle. \end{aligned}$$

*In these rules, the context* $\mathbf{c}_1$ *of* $\rho_1$ *is* $a \wedge b$, *and the context* $\mathbf{c}_2$ *of* $\rho_2$ *is* $a \wedge \neg c \wedge d$.

*Rules* $\rho_1$ *and* $\rho_2$ *are consistent, therefore, we must split them to perform the addition operation:*

$$Split(\rho_1\angle\mathbf{c}_2) = \begin{cases} \langle a \wedge b \wedge c : 5 \rangle, \\ \langle a \wedge b \wedge \neg c \wedge \neg d : 5 \rangle, \\ \langle a \wedge b \wedge \neg c \wedge d : 5 \rangle. \end{cases}$$

*Likewise,*

$$Split(\rho_2\angle\mathbf{c}_1) = \begin{cases} \langle a \wedge \neg b \wedge \neg c \wedge d : 3 \rangle, \\ \langle a \wedge b \wedge \neg c \wedge d : 3 \rangle. \end{cases}$$

*The result of adding rules* $\rho_1$ *and* $\rho_2$ *is*

$$\begin{aligned} &\langle a \wedge b \wedge c : 5 \rangle, \\ &\langle a \wedge b \wedge \neg c \wedge \neg d : 5 \rangle, \\ &\langle a \wedge b \wedge \neg c \wedge d : 8 \rangle, \\ &\langle a \wedge \neg b \wedge \neg c \wedge d : 3 \rangle. \end{aligned}$$

□

### 8.3 Rule-based One-step Lookahead

Using this compact rule-based representation, we are able to compute a one-step lookahead plan efficiently for models with significant context-specific or additive independence.

As in Section 4.1 for the table-based case, the rule-based $Q_a$ function can be represented as the sum of the reward function and the discounted expected value of the next state. Due to our linear approximation of the value function, the expectation term is, in turn, represented as the linear combination of the backprojections of our basis functions. To exploit CSI, we are representing the rewards and basis functions as rule-based functions. To represent $Q_a$ as a rule-based function, it is sufficient for us to show how to represent the backprojection $g_j$ of the basis function $h_j$ as a rule-based function.

Each $h_j$ is a rule-based function, which can be written as $h_j(\mathbf{x}) = \sum_i \rho_i^{(h_j)}(\mathbf{x})$, where $\rho_i^{(h_j)}$ has the form $\left\langle \mathbf{c}_i^{(h_j)} : v_i^{(h_j)} \right\rangle$. Each rule is a restricted scope function; thus, we can simplify the backprojection as:

*RuleBackproj*$_a(\rho)$ , WHERE $\rho$ IS GIVEN BY $\langle \mathbf{c} : v \rangle$, WITH $\mathbf{c} \in \mathrm{Dom}[\mathbf{C}]$.
  **Let** $g = \{\}$.
  **Select** THE SET $\mathcal{P}$ OF RELEVANT PROBABILITY RULES:
    $\mathcal{P} = \{\eta_j \in P(X_i' \mid \mathit{Parents}(X_i')) \mid X_i' \in \mathbf{C}$ AND $\mathbf{c}$ IS CONSISTENT WITH $\mathbf{c}_j\}$.
  **Remove** THE $\mathbf{X}'$ ASSIGNMENTS FROM THE CONTEXT OF ALL RULES IN $\mathcal{P}$.
  *// Multiply consistent rules:*
  **While** THERE ARE TWO CONSISTENT RULES $\eta_1 = \langle \mathbf{c}_1 : p_1 \rangle$ AND $\eta_2 = \langle \mathbf{c}_2 : p_2 \rangle$:
    **If** $\mathbf{c}_1 = \mathbf{c}_2$, REPLACE THESE TWO RULES BY $\langle \mathbf{c}_1 : p_1 p_2 \rangle$;
    **Else** REPLACE THESE TWO RULES BY THE SET: $\mathit{Split}(\eta_1 \angle \mathbf{c}_2) \cup \mathit{Split}(\eta_2 \angle \mathbf{c}_1)$.
  *// Generate value rules:*
  **For** EACH RULE $\eta_i$ IN $\mathcal{P}$:
    **Update** THE BACKPROJECTION $g = g \cup \{\langle \mathbf{c}_i : p_i v \rangle\}$.
  **Return** $g$.

Figure 12: Rule-based backprojection.

$$\begin{aligned} g_j^a(\mathbf{x}) &= \sum_{\mathbf{x}'} P_a(\mathbf{x}' \mid \mathbf{x}) h_j(\mathbf{x}') \ ; \\ &= \sum_{\mathbf{x}'} P_a(\mathbf{x}' \mid \mathbf{x}) \sum_i \rho_i^{(h_j)}(\mathbf{x}'); \\ &= \sum_i \sum_{\mathbf{x}'} P_a(\mathbf{x}' \mid \mathbf{x}) \rho_i^{(h_j)}(\mathbf{x}'); \\ &= \sum_i v_i^{(h_j)} P_a(\mathbf{c}_i^{(h_j)} \mid \mathbf{x}); \end{aligned}$$

where the term $v_i^{(h_j)} P_a(\mathbf{c}_i^{(h_j)} \mid \mathbf{x})$ can be written as a rule function. We denote this backprojection operation by *RuleBackproj*$_a(\rho_i^{(h_j)})$.

The backprojection procedure, described in Figure 12, follows three steps. First, the relevant rules are selected: In the CPDs for the variables that appear in the context of $\rho$, we select the rules consistent with this context, as these are the only rules that play a role in the backprojection computation. Second, we multiply all consistent probability rules to form a local set of mutually-exclusive rules. This procedure is analogous to the addition procedure described in Section 8.2. Now that we have represented the probabilities that can affect $\rho$ by a mutually-exclusive set, we can simply represent the backprojection of $\rho$ by the product of these probabilities with the value of $\rho$. That is, the backprojection of $\rho$ is a rule-based function with one rule for each one of the mutually-exclusive probability rules $\eta_i$. The context of this new value rule is the same as that of $\eta_i$, and the value is the product of the probability of $\eta_i$ and the value of $\rho$.

**Example 8.12** *For example, consider the backprojection of a simple rule,*

$$\rho = \langle \ \mathit{Painting} = \mathrm{done} : 100 \rangle,$$

*through the CPD in Figure 11(c) for the paint action:*

$$\mathit{RuleBackproj}_{\mathrm{paint}}(\rho) \quad = \quad \sum_{\mathbf{x}'} P_{\mathrm{paint}}(\mathbf{x}' \mid \mathbf{x}) \rho(\mathbf{x}');$$

$$= \sum_{Painting'} P_{\text{paint}}(Painting' \mid \mathbf{x}) \rho(Painting');$$
$$= 100 \prod_{i=1}^{3} \eta_i(Painting' = done, \mathbf{x}) \; .$$

*Note that the product of these simple rules is equivalent to the decision tree CPD shown in Figure 11(a). Hence, this product is equal to* 0 *in most contexts, for example, when electricity is not done at time* $t$*. The product in non-zero only in one context: in the context associated with rule* $\eta_3$*. Thus, we can express the result of the backprojection operation by a rule-based function with a single rule:*

$$RuleBackproj_{\text{paint}}(\rho) = \langle Plumbing \wedge Electrical : 95 \rangle.$$

*Similarly, the backprojection of* $\rho$ *when the action is not paint can also be represented by a single rule:*

$$RuleBackproj_{\neg\text{paint}}(\rho) = \langle Plumbing \wedge Electrical \wedge Painting : 90 \rangle. \quad \square$$

Using this algorithm, we can now write the *backprojection* of the rule-based basis function $h_j$ as:

$$g_j^a(\mathbf{x}) = \sum_i RuleBackproj_a(\rho_i^{(h_j)}), \tag{23}$$

where $g_j^a$ is a sum of rule-based functions, and therefore also a rule-based function. For simplicity of notation, we use $g_j^a = RuleBackproj_a(h_j)$ to refer to this definition of backprojection. Using this notation, we can write $Q_a(\mathbf{x}) = R^a(\mathbf{x}) + \gamma \sum_j w_j g_j^a(\mathbf{x})$, which is again a rule-based function.

### 8.4 Rule-based Maximization Over the State Space

The second key operation required to extend our planning algorithms to exploit CSI is to modify the variable elimination algorithm in Section 4.2.1 to handle the rule-based representation. In Section 4.2.1, we showed that the maximization of a linear combination of table-based functions with restricted scope can be performed efficiently using non-serial dynamic programming (Bertele & Brioschi, 1972), or variable elimination. To exploit structure in rules, we use an algorithm similar to variable elimination in a Bayesian network with context-specific independence (Zhang & Poole, 1999).

Intuitively, the algorithm operates by selecting the value rules relevant to the variable being maximized in the current iteration. Then, a local maximization is performed over this subset of the rules, generating a new set of rules without the current variable. The procedure is then repeated recursively until all variables have been eliminated.

More precisely, our algorithm "eliminates" variables one by one, where the elimination process performs a maximization step over the variable's domain. Suppose that we are eliminating $X_i$, whose collected value rules lead to a rule function $f$, and $f$ involves additional variables in some set $\mathbf{B}$, so that $f$'s scope is $\mathbf{B} \cup \{X_i\}$. We need to compute the maximum value for $X_i$ for each choice of $\mathbf{b} \in \text{Dom}[\mathbf{B}]$. We use $MaxOut(f, X_i)$ to denote a procedure that takes a rule function $f(\mathbf{B}, X_i)$ and returns a rule function $g(\mathbf{B})$ such

*MaxOut* $(f, B)$
**Let** $g = \{\}$.
**Add** COMPLETING RULES TO $f$: $\langle B = b_i : 0 \rangle, i = 1, \ldots, k$.
*// Summing consistent rules:*
**While** THERE ARE TWO CONSISTENT RULES $\rho_1 = \langle \mathbf{c}_1 : v_1 \rangle$ AND $\rho_2 = \langle \mathbf{c}_2 : v_2 \rangle$:
**If** $\mathbf{c}_1 = \mathbf{c}_2$, THEN REPLACE THESE TWO RULES BY $\langle \mathbf{c}_1 : v_1 + v_2 \rangle$;
**Else** REPLACE THESE TWO RULES BY THE SET: $Split(\rho_1 \angle \mathbf{c}_2) \cup Split(\rho_2 \angle \mathbf{c}_1)$.
*// Maximizing out variable $B$:*
**Repeat** UNTIL $f$ IS EMPTY:
**If** THERE ARE RULES $\langle \mathbf{c} \wedge B = b_i : v_i \rangle, \forall b_i \in \mathrm{Dom}(B)$ :
**Then** REMOVE THESE RULES FROM $f$ AND ADD RULE $\langle \mathbf{c} : \max_i v_i \rangle$ TO $g$;
**Else** SELECT TWO RULES: $\rho_i = \langle \mathbf{c}_i \wedge B = b_i : v_i \rangle$ AND $\rho_j = \langle \mathbf{c}_j \wedge B = b_j : v_j \rangle$ SUCH THAT $\mathbf{c}_i$ IS CONSISTENT WITH $\mathbf{c}_j$, BUT NOT IDENTICAL, AND REPLACE THEM WITH $Split(\rho_i \angle \mathbf{c}_j) \cup Split(\rho_j \angle \mathbf{c}_i)$ .
**Return** $g$.

Figure 13: Maximizing out variable $B$ from rule function $f$.

that: $g(\mathbf{b}) = \max_{x_i} f(\mathbf{b}, x_i)$. Such a procedure is an extension of the variable elimination algorithm of Zhang and Poole (Zhang & Poole, 1999).

The rule-based variable elimination algorithm maintains a set $\mathcal{F}$ of value rules, initially containing the set of rules to be maximized. The algorithm then repeats the following steps for each variable $X_i$ until all variables have been eliminated:

1. Collect all rules which depend on $X_i$ into $f_i$ — $f_i = \{\langle \mathbf{c} : v \rangle \in \mathcal{F} \mid X_i \in \mathbf{C}\}$ — and remove these rules from $\mathcal{F}$.
2. Perform the local maximization step over $X_i$: $g_i = MaxOut\,(f_i, X_i)$;
3. Add the rules in $g_i$ to $\mathcal{F}$; now, $X_i$ has been "eliminated."

The cost of this algorithm is polynomial in the number of new rules generated in the maximization operation $MaxOut\,(f_i, X_i)$. The number of rules is never larger and in many cases exponentially smaller than the complexity bounds on the table-based maximization in Section 4.2.1, which, in turn, was exponential only in the *induced width* of the cost network graph (Dechter, 1999). However, the computational costs involved in managing sets of rules usually imply that the computational advantage of the rule-based approach over the table-based one will only be significant in problems that possess a fair amount of context-specific structure.

In the remainder of this section, we present the algorithm for computing the local maximization $MaxOut\,(f_i, X_i)$. In the next section, we show how these ideas can be applied to extending the algorithm in Section 4.2.2 to exploit CSI in the LP representation for planning in factored MDPs.

The procedure, presented in Figure 13, is divided into two parts: first, all consistent rules are added together as described in Section 8.2; then, variable $B$ is maximized. This maximization is performed by generating a set of rules, one for each assignment of $B$, whose contexts have the same assignment for all variables except for $B$, as in Definition 8.8. This set is then substituted by a single rule without a $B$ assignment in its context and with value equal to the maximum of the values of the rules in the original set. Note that, to simplify the algorithm, we initially need to add a set of value rules with 0 value, which guarantee that our rule function $f$ is complete (*i.e.*, there is at least one rule consistent with every context).

The correctness of this procedure follows directly from the correctness of the rule-based variable elimination procedure described by Zhang and Poole, merely by replacing summations with product with max, and products with products with sums. We conclude this section with a small example to illustrate the algorithm:

**Example 8.13** *Suppose we are maximizing a for the following set of rules:*

$$\begin{aligned} \rho_1 &= \langle \neg a : 1 \rangle, \\ \rho_2 &= \langle a \wedge \neg b : 2 \rangle, \\ \rho_3 &= \langle a \wedge b \wedge \neg c : 3 \rangle, \\ \rho_4 &= \langle \neg a \wedge b : 1 \rangle. \end{aligned}$$

*When we add completing rules, we get:*

$$\begin{aligned} \rho_5 &= \langle \neg a : 0 \rangle, \\ \rho_6 &= \langle a : 0 \rangle. \end{aligned}$$

*In the first part of the algorithm, we need to add consistent rules: We add $\rho_5$ to $\rho_1$ (which remains unchanged), combine $\rho_1$ with $\rho_4$, $\rho_6$ with $\rho_2$, and then the split of $\rho_6$ on the context of $\rho_3$, to get the following inconsistent set of rules:*

$$\begin{aligned} \rho_2 &= \langle a \wedge \neg b : 2 \rangle, \\ \rho_3 &= \langle a \wedge b \wedge \neg c : 3 \rangle, \\ \rho_7 &= \langle \neg a \wedge b : 2 \rangle, && \text{(from adding } \rho_4 \text{ to the consistent rule from } Split(\rho_1 \angle \mathbf{b})\text{)} \\ \rho_8 &= \langle \neg a \wedge \neg b : 1 \rangle, && \text{(from } Split(\rho_1 \angle \mathbf{b})\text{)} \\ \rho_9 &= \langle a \wedge b \wedge c : 0 \rangle, && \text{(from } Split(\rho_6 \angle a \wedge b \wedge \neg c)\text{).} \end{aligned}$$

*Note that several rules with value 0 are also generated, but not shown here because they are added to other rules with consistent contexts. We can move to the second stage (repeat loop) of MaxOut. We remove $\rho_2$, and $\rho_8$, and maximize a out of them, to give:*

$$\rho_{10} = \langle \neg b : 2 \rangle.$$

*We then select rules $\rho_3$ and $\rho_7$ and split $\rho_7$ on c ($\rho_3$ is split on the empty set and is not changed),*

$$\begin{aligned} \rho_{11} &= \langle \neg a \wedge b \wedge c : 2 \rangle, \\ \rho_{12} &= \langle \neg a \wedge b \wedge \neg c : 2 \rangle. \end{aligned}$$

*Maximizing out a from rules $\rho_{12}$ and $\rho_3$, we get:*

$$\rho_{13} = \langle b \wedge \neg c : 3 \rangle.$$

*We are left with $\rho_{11}$, which maximized over its counterpart $\rho_9$ gives*

$$\rho_{12} = \langle b \wedge \neg c : 2 \rangle.$$

*Notice that, throughout this maximization, we have not split on the variable $C$ when $\neg b \in \mathbf{c}_i$, giving us only 6 distinct rules in the final result. This is not possible in a table-based representation, since our functions would then be over the 3 variables a,b,c, and therefore must have 8 entries.* □

### 8.5 Rule-based Factored LP

In Section 4.2.2, we showed that the LPs used in our algorithms have exponentially many constraints of the form: $\phi \geq \sum_i w_i \; c_i(\mathbf{x}) - b(\mathbf{x}), \forall \mathbf{x}$, which can be substituted by a single, equivalent, non-linear constraint: $\phi \geq \max_{\mathbf{x}} \sum_i w_i \; c_i(\mathbf{x}) - b(\mathbf{x})$. We then showed that, using variable elimination, we can represent this non-linear constraint by an equivalent set of linear constraints in a construction we called the factored LP. The number of constraints in the factored LP is linear in the size of the largest table generated in the variable elimination procedure. This table-based algorithm can only exploit additive independence. We now extend the algorithm in Section 4.2.2 to exploit *both* additive and context-specific structure, by using the rule-based variable elimination described in the previous section.

Suppose we wish to enforce the more general constraint $0 \geq \max_{\mathbf{y}} F^{\mathbf{w}}(\mathbf{y})$, where $F^{\mathbf{w}}(\mathbf{y}) = \sum_j f_j^{\mathbf{w}}(\mathbf{y})$ such that each $f_j$ is a rule. As in the table-based version, the superscript $\mathbf{w}$ means that $f_j$ might depend on $\mathbf{w}$. Specifically, if $f_j$ comes from basis function $h_i$, it is multiplied by the weight $w_i$; if $f_j$ is a rule from the reward function, it is not.

In our rule-based factored linear program, we generate LP variables associated with contexts; we call these *LP rules*. An LP rule has the form $\langle \mathbf{c} : u \rangle$; it is associated with a context $\mathbf{c}$ and a variable $u$ in the linear program. We begin by transforming all our original rules $f_j^{\mathbf{w}}$ into LP rules as follows: If rule $f_j$ has the form $\langle \mathbf{c}_j : v_j \rangle$ and comes from basis function $h_i$, we introduce an LP rule $e_j = \langle \mathbf{c}_j : u_j \rangle$ and the equality constraint $u_j = w_i v_j$. If $f_j$ has the same form but comes from a reward function, we introduce an LP rule of the same form, but the equality constraint becomes $u_j = v_j$.

Now, we have only LP rules and need to represent the constraint: $0 \geq max_{\mathbf{y}} \sum_j e_j(\mathbf{y})$. To represent such a constraint, we follow an algorithm very similar to the variable elimination procedure in Section 8.4. The main difference occurs in the *MaxOut* $(f, B)$ operation in Figure 13. Instead of generating new value rules, we generate new LP rules, with associated new variables and new constraints. The simplest case occurs when computing a split or adding two LP rules. For example, when we add two value rules in the original algorithm, we instead perform the following operation on their associated LP rules: If the LP rules are $\langle \mathbf{c} : u_i \rangle$ and $\langle \mathbf{c} : u_j \rangle$, we replace these by a new rule $\langle \mathbf{c} : u_k \rangle$, associated with a new LP variable $u_k$ with context $\mathbf{c}$, whose value should be $u_i + u_j$. To enforce this value constraint, we simply add an additional constraint to the LP: $u_k = u_i + u_j$. A similar procedure can be followed when computing the split.

More interesting constraints are generated when we perform a maximization. In the rule-based variable elimination algorithm in Figure 13, this maximization occurs when we replace a set of rules:

$$\langle \mathbf{c} \wedge B = b_i : v_i \rangle, \forall b_i \in \mathrm{Dom}(B),$$

by a new rule

$$\left\langle \mathbf{c} : \max_i v_i \right\rangle.$$

Following the same process as in the LP rule summation above, if we are maximizing

$$e_i = \langle \mathbf{c} \wedge B = b_i : u_i \rangle, \forall b_i \in \mathrm{Dom}(B),$$

we generate a new LP variable $u_k$ associated with the rule $e_k = \langle \mathbf{c} : u_k \rangle$. However, we cannot add the nonlinear constraint $u_k = \max_i u_i$, but we can add a set of equivalent linear

constraints

$$u_k \geq u_i, \ \forall i.$$

Therefore, using these simple operations, we can exploit structure in the rule functions to represent the nonlinear constraint $e_n \geq max_{\mathbf{y}} \sum_j e_j(\mathbf{y})$, where $e_n$ is the very last LP rule we generate. A final constraint $u_n = \phi$ implies that we are representing exactly the constraints in Equation (12), without having to enumerate every state.

The correctness of our rule-based factored LP construction is a corollary of Theorem 4.4 and of the correctness of the rule-based variable elimination algorithm (Zhang & Poole, 1999) .

**Corollary 8.14** *The constraints generated by the* rule-based *factored LP construction are equivalent to the non-linear constraint in Equation (12). That is, an assignment to* $(\phi, \mathbf{w})$ *satisfies the rule-based factored LP constraints if and only if it satisfies the constraint in Equation (12).* □

The number of variables and constraints in the rule-based factored LP is linear in the number of rules generated by the variable elimination process. In turn, the number of rules is no larger, and often exponentially smaller, than the number of entries in the table-based approach.

To illustrate the generation of LP constraints as just described, we now present a small example:

**Example 8.15** *Let* $e_1$*,* $e_2$*,* $e_3$*, and* $e_4$ *be the set of LP rules which depend on the variable* $b$ *being maximized. Here, rule* $e_i$ *is associated with the LP variable* $u_i$*:*

$$\begin{aligned} e_1 &= \langle a \wedge b : u_1 \rangle, \\ e_2 &= \langle a \wedge b \wedge c : u_2 \rangle, \\ e_3 &= \langle a \wedge \neg b : u_3 \rangle, \\ e_4 &= \langle a \wedge b \wedge \neg c : u_4 \rangle. \end{aligned}$$

*In this set, note that rules* $e_1$ *and* $e_2$ *are consistent. We combine them to generate the following rules:*

$$\begin{aligned} e_5 &= \langle a \wedge b \wedge c : u_5 \rangle, \\ e_6 &= \langle a \wedge b \wedge \neg c : u_1 \rangle. \end{aligned}$$

*and the constraint* $u_1 + u_2 = u_5$*. Similarly,* $e_6$ *and* $e_4$ *may be combined, resulting in:*

$$e_7 = \langle a \wedge b \wedge \neg c : u_6 \rangle.$$

*with the constraint* $u_6 = u_1 + u_4$*. Now, we have the following three inconsistent rules for the maximization:*

$$\begin{aligned} e_3 &= \langle a \wedge \neg b : u_3 \rangle, \\ e_5 &= \langle a \wedge b \wedge c : u_5 \rangle, \\ e_7 &= \langle a \wedge b \wedge \neg c : u_6 \rangle. \end{aligned}$$

*Following the maximization procedure, since no pair of rules can be eliminated right away, we split* $e_3$ *and* $e_5$ *to generate the following rules:*

$$\begin{aligned} e_8 &= \langle a \wedge \neg b \wedge c : u_3 \rangle, \\ e_9 &= \langle a \wedge \neg b \wedge \neg c : u_3 \rangle, \\ e_5 &= \langle a \wedge b \wedge c : u_5 \rangle. \end{aligned}$$

*We can now maximize $b$ out from $e_8$ and $e_5$, resulting in the following rule and constraints respectively:*

$$e_{10} = \langle a \wedge c : u_7 \rangle,$$
$$u_7 \geq u_5,$$
$$u_7 \geq u_3.$$

*Likewise, maximizing $b$ out from $e_9$ and $e_6$, we get:*

$$e_{11} = \langle a \wedge \neg c : u_8 \rangle,$$
$$u_8 \geq u_3,$$
$$u_8 \geq u_6;$$

*which completes the elimination of variable $b$ in our rule-based factored LP.* □

We have presented an algorithm for exploiting both additive and context-specific structure in the LP construction steps of our planning algorithms. This rule-based factored LP approach can now be applied directly in our approximate linear programming and approximate policy iteration algorithms, which were presented in Sections 5 and 6.

The only additional modification required concerns the manipulation of the decision list policies presented in Section 6.2. Although approximate linear programming does not require any explicit policy representation (or the default action model), approximate policy iteration require us to represent such policy. Fortunately, no major modifications are required in the rule-based case. In particular, the conditionals $\langle \mathbf{t}_i, a_i, \delta_i \rangle$ in the decision list policies are already context-specific rules. Thus, the policy representation algorithm in Section 6.2 can be applied directly with our new rule-based representation. Therefore, we now have a complete framework for exploiting both additive and context-specific structure for efficient planning in factored MDPs.

## 9. Experimental Results

The factored representation of a value function is most appropriate in certain types of systems: Systems that involve many variables, but where the strong interactions between the variables are fairly sparse, so that the decoupling of the influence between variables does not induce an unacceptable loss in accuracy. As argued by Herbert Simon (1981) in “Architecture of Complexity,” many complex systems have a “nearly decomposable, hierarchical structure,” with the subsystems interacting only weakly between themselves. To evaluate our algorithm, we selected problems that we believe exhibit this type of structure.

In this section, we perform various experiments intended to explore the performance of our algorithms. First, we compare our factored approximate linear programming (LP) and approximate policy iteration (PI) algorithms. We also compare to the $\mathcal{L}_2$-projection algorithm of Koller and Parr (2000). Our second evaluation compares a table-based implementation to a rule-based implementation that can exploit CSI. Finally, we present comparisons between our approach and the algorithms of Boutilier *et al.* (2000).

### 9.1 Approximate LP and Approximate PI

In order to compare our approximate LP and approximate PI algorithms, we tested both on the *SysAdmin* problem described in detail in Section 2.1. This problem relates to a system administrator who has to maintain a network of computers; we experimented with various network architectures, shown in Figure 1. Machines fail randomly, and a faulty machine increases the probability that its neighboring machines will fail. At every time step, the SysAdmin can go to one machine and reboot it, causing it to be working in the next time step with high probability. Recall that the state space in this problem grows exponentially in the number of machines in the network, that is, a problem with $m$ machines has $2^m$ states. Each machine receives a reward of 1 when working (except in the ring, where one machine receives a reward of 2, to introduce some asymmetry), a zero reward is given to faulty machines, and the discount factor is $\gamma = 0.95$. The optimal strategy for rebooting machines will depend upon the topology, the discount factor, and the status of the machines in the network. If machine $i$ and machine $j$ are both faulty, the benefit of rebooting $i$ must be weighed against the expected discounted impact of delaying rebooting $j$ on $j$'s successors. For topologies such as rings, this policy may be a function of the status of every single machine in the network.

The basis functions used included independent indicators for each machine, with value 1 if it is working and zero otherwise (*i.e.*, each one is a restricted scope function of a single variable), and the constant basis, whose value is 1 for all states. We selected straightforward variable elimination orders: for the "Star" and "Three Legs" topologies, we first eliminated the variables corresponding to computers in the legs, and the center computer (server) was eliminated last; for "Ring," we started with an arbitrary computer and followed the ring order; for "Ring and Star," the ring machines were eliminated first and then the center one; finally, for the "Ring of Rings" topology, we eliminated the computers in the outer rings first and then the ones in the inner ring.

We implemented the factored policy iteration and linear programming algorithms in Matlab, using CPLEX as the LP solver. Experiments were performed on a Sun UltraSPARC-II, 359 MHz with 256MB of RAM. To evaluate the complexity of the approximate policy iteration with max-norm projection algorithm, tests were performed with increasing the number of states, that is, increasing number of machines on the network. Figure 14 shows the running time for increasing problem sizes, for various architectures. The simplest one is the "Star," where the backprojection of each basis function has scope restricted to two variables and the largest factor in the cost network has scope restricted to two variables. The most difficult one was the "Bidirectional Ring," where factors contain five variables.

Note that the number of states is growing exponentially (indicated by the log scale in Figure 14), but running times increase only logarithmically in the number of states, or polynomially in the number of variables. We illustrate this behavior in Figure 14(d), where we fit a 3rd order polynomial to the running times for the "unidirectional ring." Note that the size of the problem description grows quadratically with the number of variables: adding a machine to the network also adds the possible action of fixing that machine. For this problem, the computation cost of our factored algorithm empirically grows approximately as $O\left((n \cdot |A|)^{1.5}\right)$, for a problem with $n$ variables, as opposed to the exponential complexity — $poly\left(2^n, |A|\right)$ — of the explicit algorithm.

For further evaluation, we measured the error in our approximate value function relative to the true optimal value function $\mathcal{V}^*$. Note that it is only possible to compute $\mathcal{V}^*$ for small problems; in our case, we were only able to go up to 10 machines. For comparison, we also evaluated the error in the approximate value function produced by the $\mathcal{L}_2$-projection algorithm of Koller and Parr (2000). As we discussed in Section 6.4, the $\mathcal{L}_2$ projections in factored MDPs by Koller and Parr are difficult and time consuming; hence, we were only able to compare the two algorithms for smaller problems, where an equivalent $\mathcal{L}_2$-projection can be implemented using an explicit state space formulation. Results for both algorithms are presented in Figure 15(a), showing the relative error of the approximate solutions to the true value function for increasing problem sizes. The results indicate that, for larger problems, the max-norm formulation generates a better approximation of the true optimal value function $\mathcal{V}^*$ than the $\mathcal{L}_2$-projection. Here, we used two types of basis functions: the same single variable functions, and pairwise basis functions. The pairwise basis functions contain indicators for neighboring pairs of machines (*i.e.*, functions of two variables). As expected, the use of pairwise basis functions resulted in better approximations.

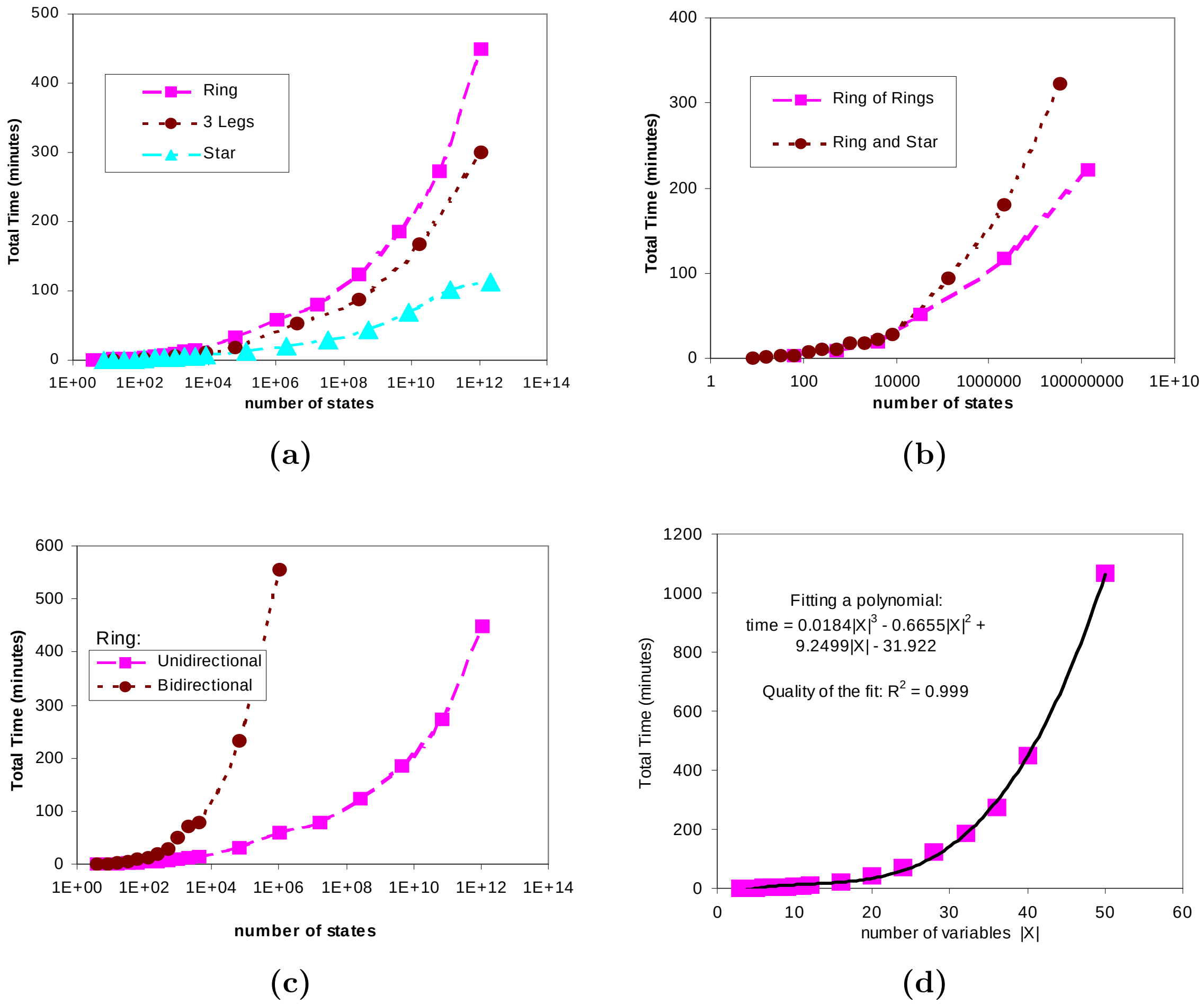

Figure 14: (a)–(c) Running times for policy iteration with max-norm projection on variants of the *SysAdmin* problem; (d) Fitting a polynomial to the running time for the "Ring" topology.

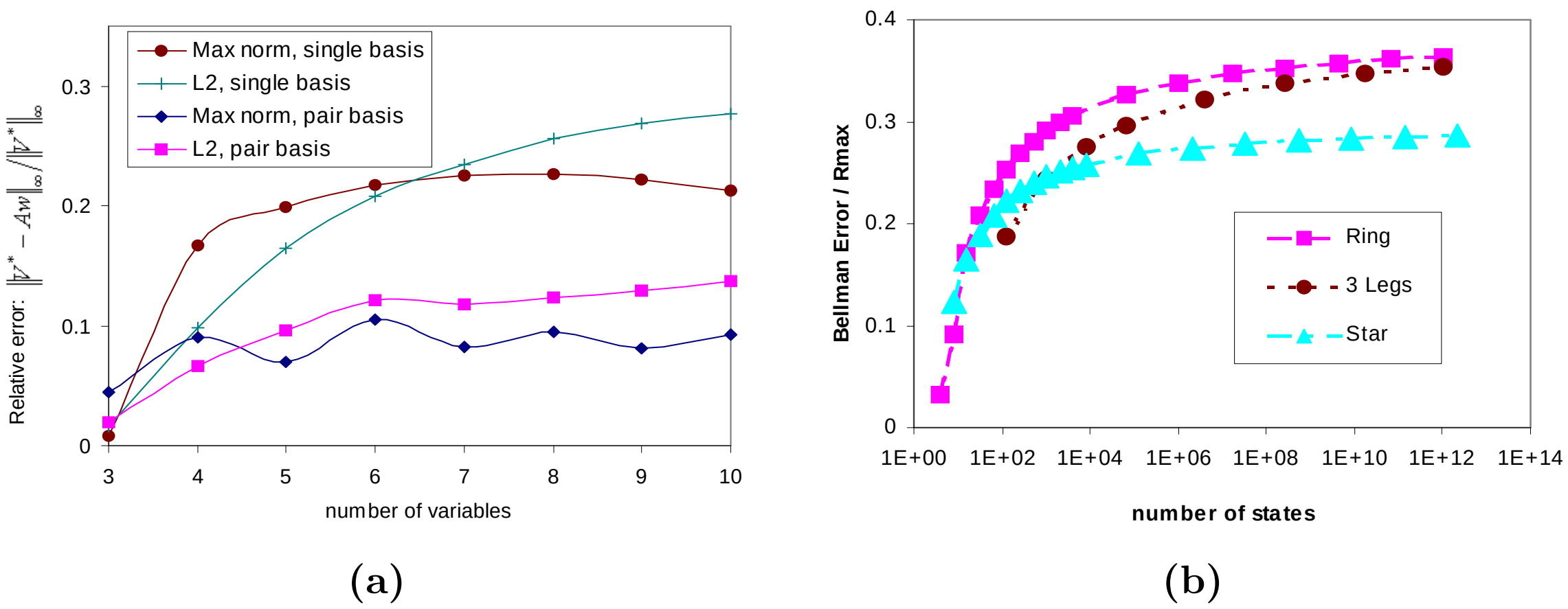

Figure 15: (a) Relative error to optimal value function $\mathcal{V}^*$ and comparison to $\mathcal{L}_2$ projection for "Ring"; (b) For large models, measuring Bellman error after convergence.

For these small problems, we can also compare the actual value of the policy generated by our algorithm to the value of the optimal policy. Here, the value of the policy generated by our algorithm is much closer to the value of the optimal policy than the error implied by the difference between our approximate value function and $\mathcal{V}^*$. For example, for the "Star" architecture with one server and up to 6 clients, our approximation with single variable basis functions had relative error of 12%, but the policy we generated had the same value as the optimal policy. In this case, the same was true for the policy generated by the $\mathcal{L}_2$ projection. In a "Unidirectional Ring" with 8 machines and pairwise basis, the relative error between our approximation and $\mathcal{V}^*$ was about 10%, but the resulting policy only had a 6% loss over the optimal policy. For the same problem, the $\mathcal{L}_2$ approximation has a value function error of 12%, and a true policy loss was 9%. In other words, both methods induce policies that have lower errors than the errors in the approximate value function (at least for small problems). However, our algorithm continues to outperform the $\mathcal{L}_2$ algorithm, even with respect to actual policy loss.

For large models, we can no longer compute the correct value function, so we cannot evaluate our results by computing $\|\mathcal{V}^* - \mathbf{Hw}\|_\infty$. Fortunately, as discussed in Section 7, the Bellman error can be used to provide a bound on the approximation error and can be computed efficiently by exploiting problem-specific structure. Figure 15(b) shows that the Bellman error increases very slowly with the number of states.

It is also valuable to look at the actual decision-list policies generated in our experiments. First, we noted that the lists tended to be short, the length of the final decision list policy grew approximately linearly with the number of machines. Furthermore, the policy itself is often fairly intuitive. In the "Ring and Star" architecture, for example, the decision list says: If the server is faulty, fix the server; else, if another machine is faulty, fix it.

Thus far, we have presented scaling results for running times and approximation error for our approximate PI approach. We now compare this algorithm to the simpler approximate

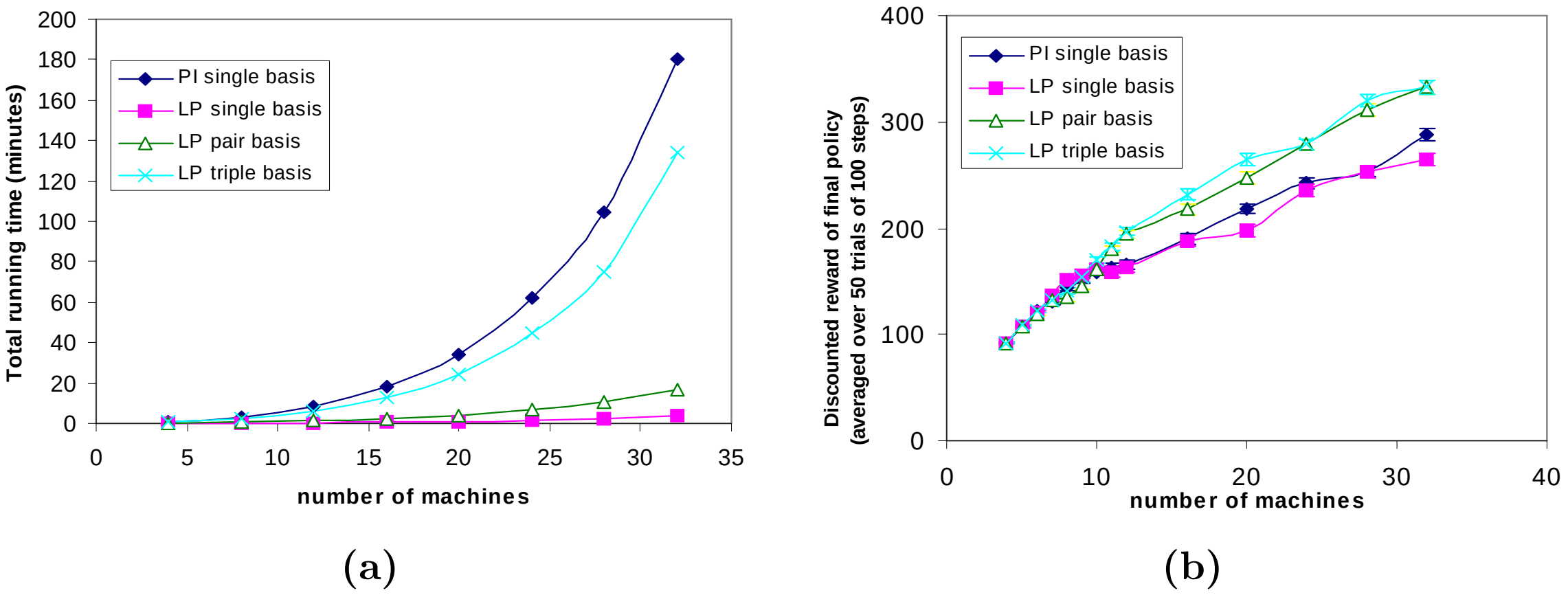

Figure 16: Approximate LP versus approximate PI on the *SysAdmin* problem with a "Ring" topology: (a) running time; (b) estimated value of policy.

LP approach of Section 5. As shown in Figure 16(a), the approximate LP algorithm for factored MDPs is significantly faster than the approximate PI algorithm. In fact, approximate PI with single-variable basis functions variables is more costly computationally than the LP approach using basis functions over consecutive triples of variables. As shown in Figure 16(b), for singleton basis functions, the approximate PI policy obtains slightly better performance for some problem sizes. However, as we increase the number of basis functions for the approximate LP formulation, the value of the resulting policy is much better. Thus, in this problem, our factored approximate linear programming formulation allows us to use more basis functions and to obtain a resulting policy of higher value, while still maintaining a faster running time. These results, along with the simpler implementation, suggest that in practice one may first try to apply the approximate linear programming algorithm before deciding to move to the more elaborate approximate policy iteration approach.

## 9.2 Comparing Table-based and Rule-based Implementations

Our next evaluation compares a table-based representation, which exploits only additive independence, to the rule-based representation presented in Section 8, which can exploit both additive and context-specific independence. For these experiments, we implemented our factored approximate linear programming algorithm with table-based and rule-based representations in C++, using CPLEX as the LP solver. Experiments were performed on a Sun UltraSPARC-II, 400 MHz with 1GB of RAM.

To evaluate and compare the algorithms, we utilized a more complex extension of the *SysAdmin* problem. This problem, dubbed the *Process-SysAdmin* problem, contains three state variables for each machine $i$ in the network: $Load_i$, $Status_i$ and $Selector_i$. Each computer runs processes and receives rewards when the processes terminate. These processes are represented by the $Load_i$ variable, which takes values in $\{Idle, Loaded, Success\}$, and the computer receives a reward when the assignment of $Load_i$ is $Success$. The $Status_i$ variable,

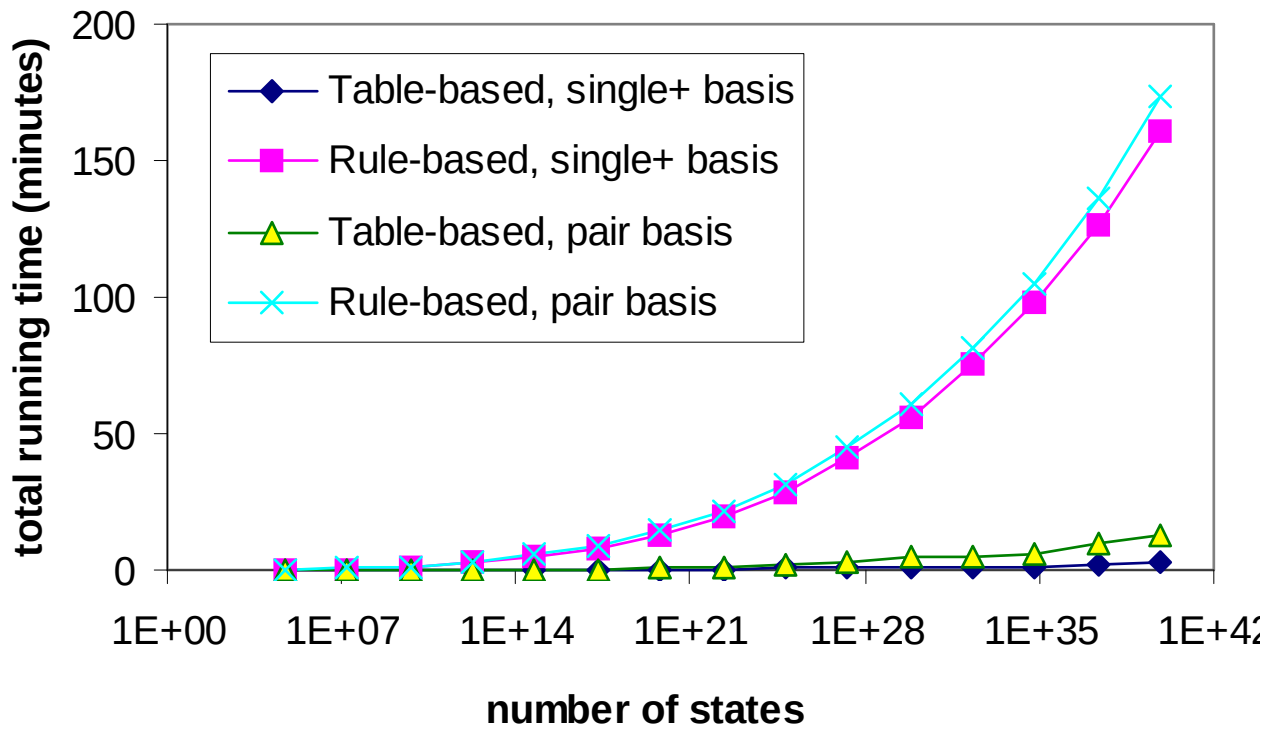

**(a)**

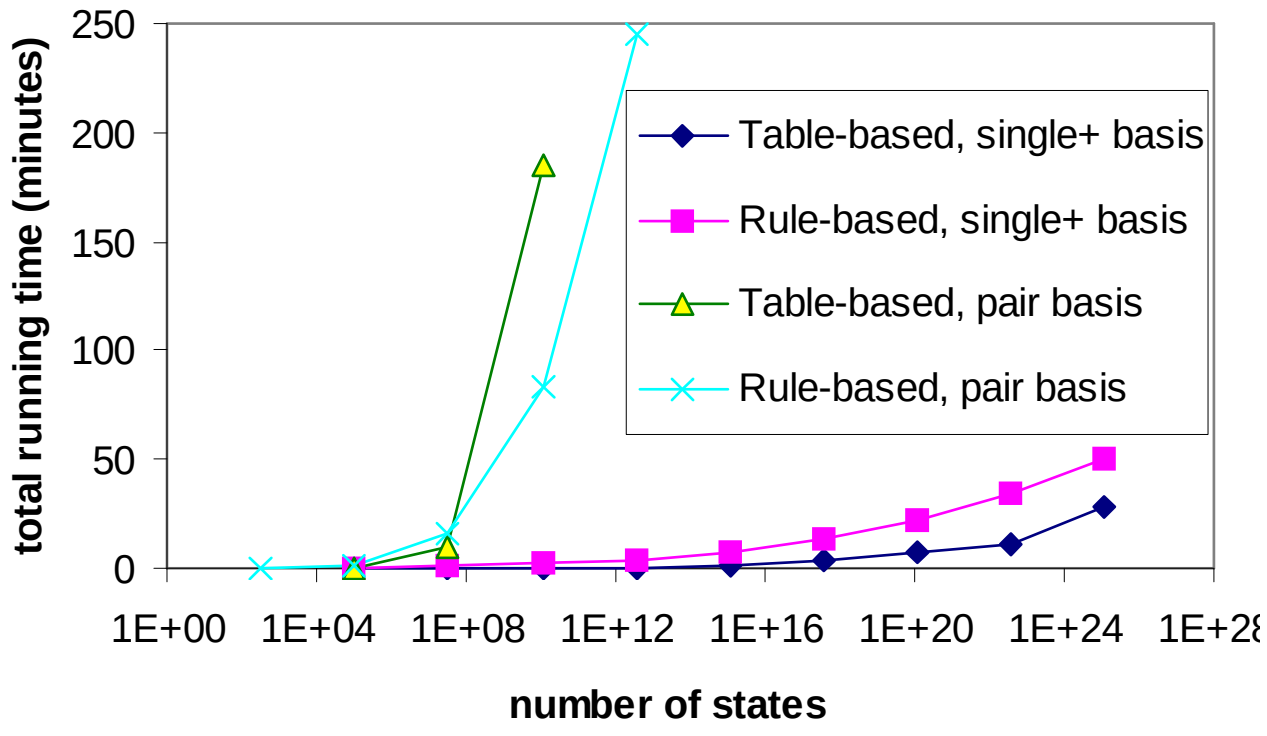

**(b)**

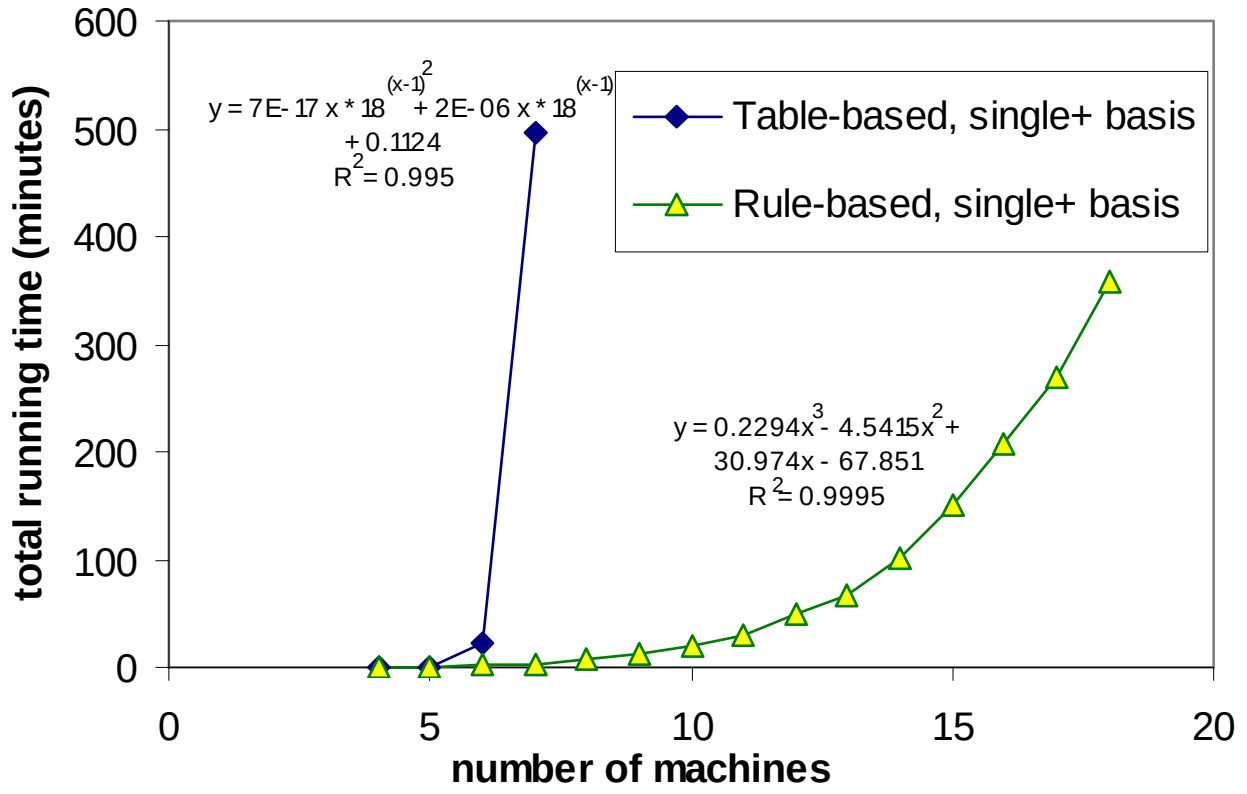

**(c)**

Figure 17: Running time for *Process-SysAdmin* problem for various topologies: (a) "Star"; (b) "Ring"; (c) "Reverse star" (with fit function).

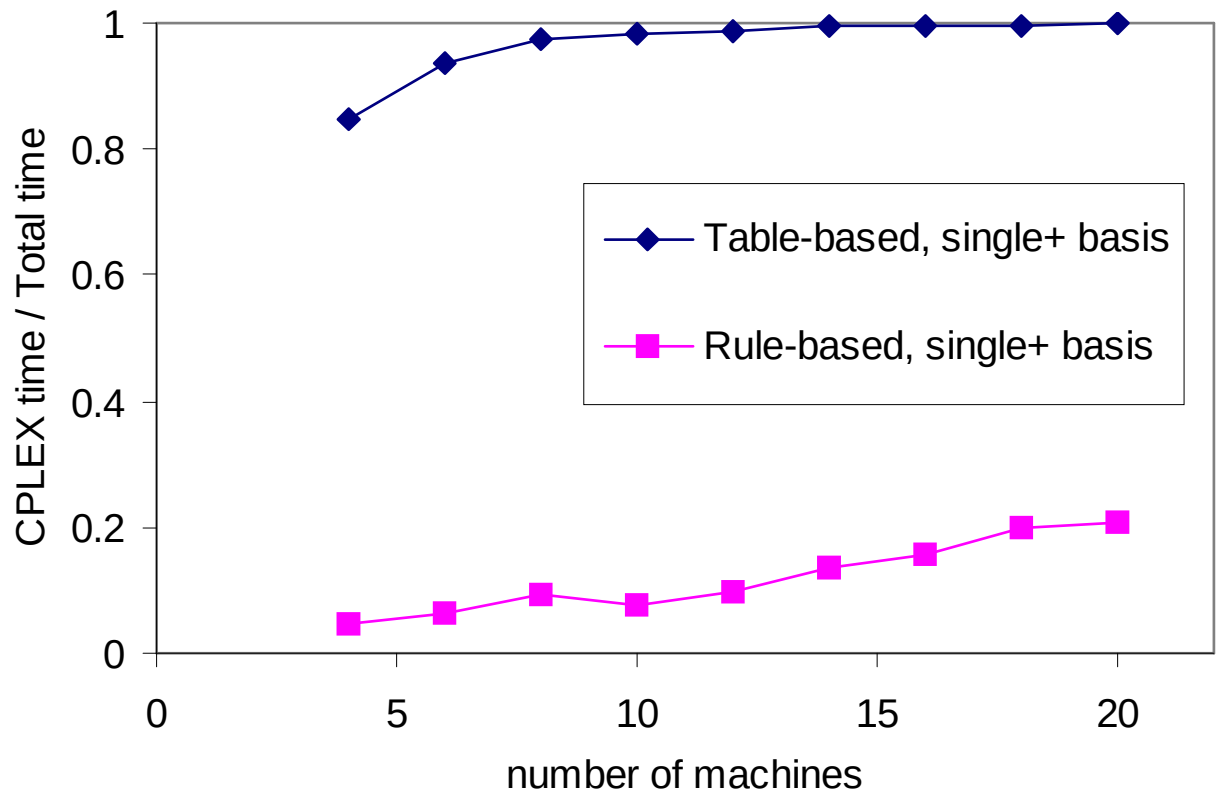

Figure 18: Fraction of total running time spent in CPLEX for the *Process-SysAdmin* problem with a "Ring" topology.

representing the status of machine $i$, takes values in $\{Good,\ Faulty,\ Dead\}$; if its value is *Faulty*, then processes have a smaller probability of terminating and if its value is *Dead*, then any running process is lost and $Load_i$ becomes *Idle*. The status of machine $i$ can become *Faulty* and eventually *Dead* at random; however, if machine $i$ receives a packet from a dead machine, then the probability that $Status_i$ becomes *Faulty* and then *Dead* increases. The $Selector_i$ variable represents this communication by selecting one of the neighbors of $i$ uniformly at random at every time step. The SysAdmin can select at most one computer to reboot at every time step. If computer $i$ is rebooted, then its status becomes *Good* with probability 1, but any running process is lost, *i.e.*, the $Load_i$ variable becomes *Idle*. Thus, in this problem, the SysAdmin must balance several conflicting goals: rebooting a machine kills processes, but not rebooting a machine may cause cascading faults in network. Furthermore, the SysAdmin can only choose one machine to reboot, which imposes the additional tradeoff of selecting only one of the (potentially many) faulty or dead machines in the network to reboot.

We experimented with two types of basis functions: "single+" includes indicators over all of the joint assignments of $Load_i$, $Status_i$ and $Selector_i$, and "pair" which, in addition, includes a set of indicators over $Status_i$, $Status_j$, and $Selector_i = j$, for each neighbor $j$ of machine $i$ in the network. The discount factor was $\gamma = 0.95$. The variable elimination order eliminated all of the $Load_i$ variables first, and then followed the same patterns as in the simple *SysAdmin* problem, eliminating first $Status_i$ and then $Selector_i$ when machine $i$ is eliminated.

Figure 17 compares the running times for the table-based implementation to the ones for the rule-based representation for three topologies: "Star," "Ring," and "Reverse star." The "Reverse star" topology reverses the direction of the influences in the "Star": rather than the central machine influencing all machines in the topology, all machines influence the central one. These three topologies demonstrate three different levels of CSI: In the "Star" topology, the factors generated by variable elimination are small. Thus, although the running times are polynomial in the number of state variables for both methods, the table-based representation is significantly faster than the rule-based one, due to the overhead of managing the rules. The "Ring" topology illustrates an intermediate behavior: "single+" basis functions induce relatively small variable elimination factors, thus the table-based approach is faster. However, with "pair" basis the factors are larger and the rule-based approach starts to demonstrate faster running times in larger problems. Finally, the "Reverse star" topology represents the worst-case scenario for the table-based approach. Here, the scope of the backprojection of a basis function for the central machine will involve all computers in the network, as all machines can potentially influence the central one in the next time step. Thus, the size of the factors in the table-based variable elimination approach are exponential in the number of machines in the network, which is illustrated by the exponential growth in Figure 17(c). The rule-based approach can exploit the CSI in this problem; for example, the status of the central machine $Status_0$ only depends on machine $j$ if the value selector is $j$, *i.e.*, if $Selector_0 = j$. By exploiting CSI, we can solve the same problem in polynomial time in the number of state variables, as seen in the second curve in Figure 17(c).

It is also instructive to compare the portion of the total running time spent in CPLEX for the table-based as compared to the rule-based approach. Figure 18 illustrates this comparison. Note that amount of time spent in CPLEX is significantly higher for the table-based approach. There are two reasons for this difference: first, due to CSI, the LPs generated by the rule-based approach are smaller than the table-based ones; second, rule-based variable elimination is more complex than the table-based one, due to the overhead introduced by rule management. Interestingly, the proportion of CPLEX time increases as the problem size increases, indicating that the asymptotic complexity of the LP solution is higher than that of variable elimination, thus suggesting that, for larger problems, additional large-scale LP optimization procedures, such as constraint generation, may be helpful.

### 9.3 Comparison to Apricodd

The most closely related work to ours is a line of research that began with the work of Boutilier *et al.* (1995). In particular, the approximate Apricodd algorithm of Hoey *et al.* (1999), which uses analytic decision diagrams (ADDs) to represent the value function is a strong alternative approach for solving factored MDPs. As discussed in detail in Section 10, the Apricodd algorithm can successfully exploit context-specific structure in the *value function*, by representing it with the set of mutually-exclusive and exhaustive branches of the ADD. On the other hand, our approach can exploit both additive and context-specific structure in the problem, by using a linear combination of non-mutually-exclusive rules. To better understand this difference, we evaluated both our rule-based approximate linear programming algorithm and Apricodd in two problems, *Linear* and *Expon*, designed by Boutilier *et al.* (2000) to illustrate respectively the best-case and the worst-case behavior of their algorithm. In these experiments, we used the web-distributed version of Apricodd (Hoey, St-Aubin, Hu, & Boutilier, 2002), running it locally on a Linux Pentium III 700MHz with 1GB of RAM.

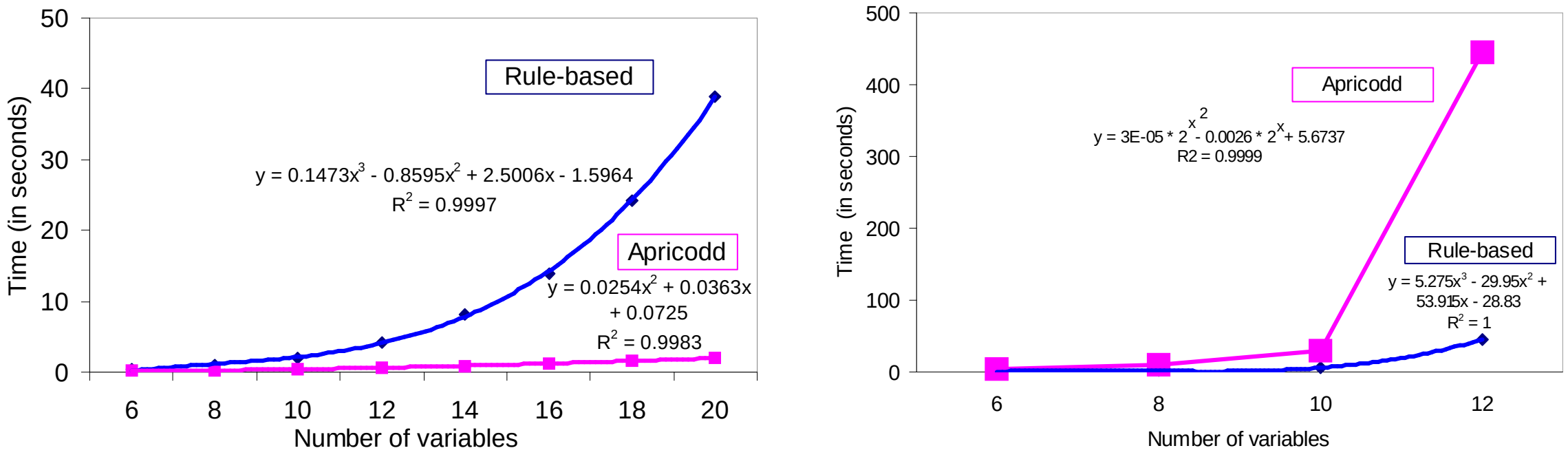

**(a)** **(b)**

Figure 19: Comparing Apricodd to rule-based approximate linear programming on the (a) *Linear* and (b) *Expon* problems.

These two problems involve $n$ binary variables $X_1, \ldots, X_n$ and $n$ deterministic actions $a_1, \ldots, a_n$. The reward is 1 when all variables $X_k$ are *true*, and is 0 otherwise. The problem is discounted by a factor $\gamma = 0.99$. The difference between the *Linear* and the *Expon* problems is in the transition probabilities. In the *Linear* problem, the action $a_k$ sets the variable $X_k$ to *true* and makes all *succeeding* variables, $X_i$ for $i > k$, *false*. If the state space of the *Linear* problem is seen as a binary number, the optimal policy is to set repeatedly the largest bit ($X_k$ variable) which has all preceding bits set to *true*. Using an ADD, the optimal value function for this problem can be represented in linear space, with $n+1$ leaves (Boutilier *et al.*, 2000). This is the "best-case" for Apricodd, and the algorithm can compute this value function quite efficiently. Figure 19(a) compares the running time of Apricodd to that of one of our algorithms with indicator basis functions between pairs of consecutive variables. Note that both algorithms obtain the same policy in polynomial time in the number of variables. However, in such structured problems, the efficient implementation of the ADD package used in Apricodd makes it faster in this problem.

On the other hand, the *Expon* problem illustrates the worst-case for Apricodd. In this problem, the action $a_k$ sets the variable $X_k$ to *true*, if all *preceding* variables, $X_i$ for $i < k$, are *true*, and it makes all preceding variables *false*. If the state space is seen as a binary number, the optimal policy goes through all binary numbers in sequence, by repeatedly setting the largest bit ($X_k$ variable) which has all preceding bits set to *true*. Due to discounting, the optimal value function assigns a value of $\gamma^{2^n - j - 1}$ to the $j$th binary number, so that the value function contains exponentially many different values. Using an ADD, the optimal value function for this problem requires an exponential number of leaves (Boutilier *et al.*, 2000), which is illustrated by the exponential running time in Figure 19(b). However, the same value function can be approximated very compactly as a factored linear value function using $n + 1$ basis functions: an indicator over each variable $X_k$ and the constant base. As shown in Figure 19(b), using this representation, our factored approximate linear programming algorithm computes the value function in polynomial time. Furthermore, the policy obtained by our approach was optimal for this problem. Thus, in this problem, the ability to exploit additive independence allows an efficient polynomial time solution.

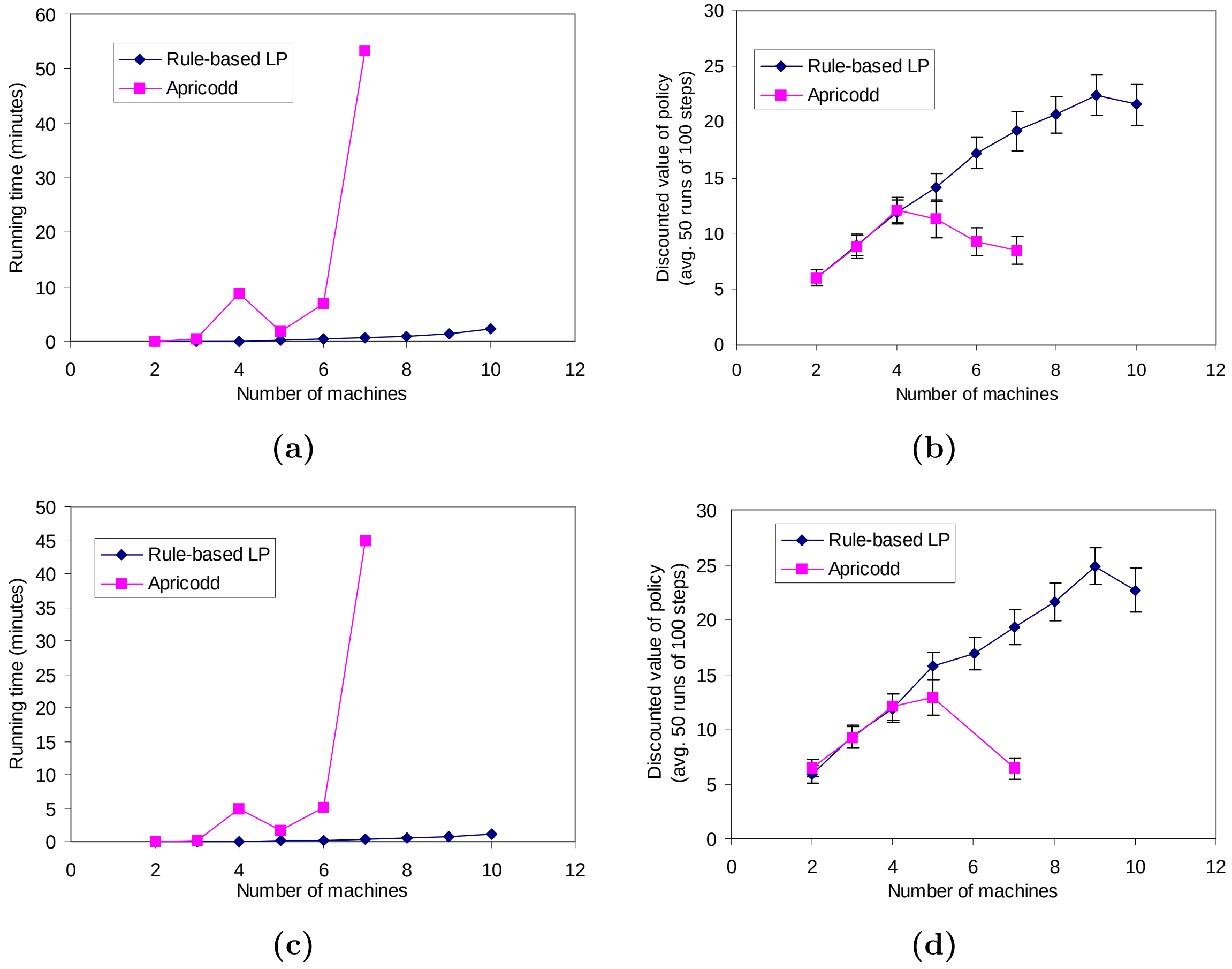

Figure 20: Comparing Apricodd to rule-based approximate linear programming with "single+" basis functions on the *Process-SysAdmin* problem with "Ring" topology (a) running time and (b) value of the resulting policy; and with "Star" topology (c) running time and (d) value of the resulting policy.

We have also compared Apricodd to our rule-based approximate linear programming algorithm on the *Process-SysAdmin* problem. This problem has significant additive structure in the reward function and factorization in the transition model. Although this type of structure is not exploited directly by Apricodd, the ADD approximation steps performed by the algorithm can, in principle, allow Apricodd to find approximate solutions to the problem. We spent a significant amount of time attempting to find the best set of parameters for Apricodd for these problems.[4] We settled on the "sift" method of variable reordering and the "round" approximation method with the "size" (maximum ADD size) criteria. To

4. We are very grateful to Jesse Hoey and Robert St-Aubin for their assistance in selecting the parameters.

allow the value function representation to scale with the problem size, we set the maximum ADD size to $4000 + 400n$ for a network with $n$ machines. (We experimented with a variety of different growth rates for the maximum ADD size; here, as for the other parameters, we selected the choice that gave the best results for Apricodd.) We compared Apricodd with these parameters to our rule-based approximate linear programming algorithm with "single+" basis functions on a Pentium III 700MHz with 1GB of RAM. These results are summarized in Figure 20.

On very small problems (up to 4–5 machines), the performance of the two algorithms is fairly similar in terms of both the running time and the quality of the policies generated. However, as the problem size grows, the running time of Apricodd increases rapidly, and becomes significantly higher than that of our algorithm . Furthermore, as the problem size increases, the quality of the policies generated by Apricodd also deteriorates. This difference in policy quality is caused by the different value function representation used by the two algorithms. The ADDs used in Apricodd represent $k$ different values with $k$ leaves; thus, they are forced to agglomerate many different states and represent them using a single value. For smaller problems, such agglomeration can still represent good policies. Unfortunately, as the problem size increases and the state space grows exponentially, Apricodd's policy representation becomes inadequate, and the quality of the policies decreases. On the other hand, our linear value functions can represent exponentially many values with only $k$ basis functions, which allows our approach to scale up to significantly larger problems.

## 10. Related Work

The most closely related work to ours is a line of research that began with the work of Boutilier *et al.* (1995). We address this comparison separately below, but we begin this section with some broader background references.

### 10.1 Approximate MDP Solutions

The field of MDPs, as it is popularly known, was formalized by Bellman (1957) in the 1950's. The importance of value function approximation was recognized at an early stage by Bellman himself (1963). In the early 1990's the MDP framework was recognized by AI researchers as a formal framework that could be used to address the problem of planning under uncertainty (Dean, Kaelbling, Kirman, & Nicholson, 1993).

Within the AI community, value function approximation developed concomitantly with the notion of value function representations for Markov chains. Sutton's seminal paper on temporal difference learning (1988), which addressed the use of value functions for prediction but not planning, assumed a very general representation of the value function and noted the connection to general function approximators such as neural networks. However, the stability of this combination was not directly addressed at that time.

Several important developments gave the AI community deeper insight into the relationship between function approximation and dynamic programming. Tsitsiklis and Van Roy (1996a) and, independently, Gordon (1995) popularized the analysis of approximate MDP methods via the contraction properties of the dynamic programming operator and function approximator. Tsitsiklis and Van Roy (1996b) later established a general convergence result for linear value function approximators and $TD(\lambda)$, and Bertsekas and Tsitsiklis (1996) unified a large body of work on approximate dynamic programming under the name of *Neuro-dynamic Programming*, also providing many novel and general error analyses.

Approximate linear programming for MDPs using linear value function approximation was introduced by Schweitzer and Seidmann (1985), although the approach was somewhat deprecated until fairly recently due the lack of compelling error analyses and the lack of an effective method for handling the large number of constraints. Recent work by de Farias and Van Roy (2001a, 2001b) has started to address these concerns with new error bounds and constraint sampling methods. Our approach, rather than sampling constraints, utilizes structure in the model and value function to represent all of the constraints compactly.

### 10.2 Factored Approaches

Tatman and Shachter (1990) considered the additive decomposition of value nodes in influence diagrams. A number of approaches to factoring of general MDPs have been explored in the literature. Techniques for exploiting reward functions that decompose additively were studied by Meuleau *et al.* (1998), and by Singh and Cohn (1998).

The use of factored representations such as dynamic Bayesian networks was pioneered by Boutilier *et al.* (1995) and has developed steadily in recent years. These methods rely on the use of context-specific structures such as decision trees or analytic decision diagrams (ADDs) (Hoey *et al.*, 1999) to represent both the transition dynamics of the DBN and the value function. The algorithms use dynamic programming to partition the state space, representing the partition using a tree-like structure that branches on state variables and assigns values at the leaves. The tree is grown dynamically as part of the dynamic programming process and the algorithm creates new leaves as needed: A leaf is split by the application of a DP operator when two states associated with that leaf turn out to have different values in the backprojected value function. This process can also be interpreted as a form of model minimization (Dean & Givan, 1997).

The number of leaves in a tree used to represent a value function determines the computational complexity of the algorithm. It also limits the number of distinct values that can be assigned to states: since the leaves represent a partitioning of the state space, every state maps to exactly one leaf. However, as was recognized early on, there are trivial MDPs which require exponentially large value functions. This observation led to a line of approximation algorithms aimed at limiting the tree size (Boutilier & Dearden, 1996) and, later, limiting the ADD size (St-Aubin, Hoey, & Boutilier, 2001). Kim and Dean (2001) also explored techniques for discovering tree-structured value functions for factored MDPs. While these methods permit good approximate solutions to some large MDPs, their complexity is still determined by the number of leaves in the representation and the number of distinct values than can be assigned to states is still limited as well.

Tadepalli and Ok (1996) were the first to apply linear value function approximation to Factored MDPs. Linear value function approximation is a potentially more expressive approximation method because it can assign unique values to every state in an MDP without requiring storage space that is exponential in the number of state variables. The expressive power of a tree with $k$ leaves can be captured by a linear function approximator with $k$ basis functions such that basis function $h_i$ is an indicator function that tests if a state belongs in the partition of leaf $i$. Thus, the set of value functions that can be represented by a tree with $k$ leaves is a subset of the set of value functions that can be represented by a value function with $k$ basis functions. Our experimental results in Section 9.3 highlight this difference by showing an example problem that requires exponentially many leaves in the value function, but that can be approximated well using a linear value function.

The main advantage of tree-based value functions is that their structure is determined dynamically during the solution of the MDP. In principle, as the value function representation is derived automatically from the model description, this approach requires less insight from the user. In problems for which the value function can be well approximated by a relatively small number of values, this approach provides an excellent solution to the problem. Our method of linear value function approximation aims to address what we believe to be the more common case, where a large range of distinct values is required to achieve a good approximation.

Finally, we note that Schuurmans and Patrascu (2001), based on our earlier work on max-norm projection using cost networks and linear programs, independently developed an alternative approach to approximate linear programming using a cost network. Our method embeds a cost network inside a single linear program. By contrast, their method is based on a constraint generation approach, using a cost network to detect constraint violations. When constraint violations are found, a new constraint is added, repeatedly generating and attempting to solve LPs until a feasible solution is found. Interestingly, as the approach of Schuurmans and Patrascu uses multiple calls to variable elimination in order to speed up the LP solution step, it will be most successful when the time spent solving the LP is significantly larger than the time required for variable elimination. As suggested in Section 9.2, the LP solution time is larger for the table-based approach. Thus, Schuurmans and Patrascu's constraint generation method will probably be more successful in table-based problems than in rule-based ones.

## 11. Conclusions

In this paper, we presented new algorithms for approximate linear programming and approximate dynamic programming (value and policy iteration) for factored MDPs. Both of these algorithms leverage on a novel LP decomposition technique, analogous to variable elimination in cost networks, which reduces an exponentially large LP to a provably equivalent, polynomial-sized one.

Our approximate dynamic programming algorithms are motivated by error analyses showing the importance of minimizing $\mathcal{L}_\infty$ error. These algorithms are more efficient and substantially easier to implement than previous algorithms based on the $\mathcal{L}_2$-projection. Our experimental results suggest that they also perform better in practice.

Our approximate linear programming algorithm for factored MDPs is simpler, easier to implement and more general than the dynamic programming approaches. Unlike our policy iteration algorithm, it does not rely on the default action assumption, which states that actions only affect a small number of state variables. Although this algorithm does not have the same theoretical guarantees as max-norm projection approaches, empirically it seems to be a favorable option. Our experiments suggest that approximate policy iteration tends to generate better policies for the same set of basis functions. However, due to the computa-tional advantages, we can add more basis functions to the approximate linear programming algorithm, obtaining a better policy and still maintaining a much faster running time than approximate policy iteration.

Unlike previous approaches, our algorithms can exploit both additive and context-specific structure in the factored MDP model. Typical real-world systems possess both of these types of structure. thus, this feature of our algorithms will increase the applicability of factored MDPs to more practical problems. We demonstrated that exploiting context-specific independence, by using a rule-based representation instead of the standard table-based one, can yield exponential improvements in computational time when the problem has significant amounts of CSI. However, the overhead of managing sets of rules make it less well-suited for simpler problems. We also compared our approach to the work of Boutilier *et al.* (2000), which exploits only context-specific structure. For problems with significant context-specific structure in the value function, their approach can be faster due to their efficient handling of the ADD representation. However, there are problems with significant context-specific structure in the problem representation, rather than in the value function, which require exponentially large ADDs. In some such problems, we demonstrated that by using a linear value function our algorithm can obtain a polynomial-time near-optimal approximation of the true value function.

The success of our algorithm depends on our ability to capture the most important structure in the value function using a linear, factored approximation. This ability, in turn, depends on the choice of the basis functions and on the properties of the domain. The algorithms currently require the designer to specify the factored basis functions. This is a limitation compared to the algorithms of Boutilier *et al.* (2000), which are fully automated. However, our experiments suggest that a few simple rules can be quite successful for designing a basis. First, we ensure that the reward function is representable by our basis. A simple basis that, in addition, contained a separate set of indicators for each variable often did quite well. We can also add indicators over pairs of variables; most simply, we can choose these according to the DBN transition model, where an indicator is added between variables $X_i$ and each one of the variables in $\mathit{Parents}(X_i)$, thus representing one-step influences. This procedure can be extended, adding more basis functions to represent more influences as required. Thus, the structure of the DBN gives us indications of how to choose the basis functions. Other sources of prior knowledge can also be included for further specifying the basis.

Nonetheless, a general algorithm for choosing good factored basis functions still does not exist. However, there are some potential approaches: First, in problems with CSI, one could apply the algorithms of Boutilier *et al.* for a few iterations to generate partial tree-structured solutions. Indicators defined over the variables in backprojection of the leaves could, in turn, be used to generate a basis set for such problems. Second, the Bellman error computation, which can be performed efficiently as shown in Section 7, does not only provide a bound on the quality of the policy, but also the actual state where the error is largest. This knowledge can be used to create a mechanism to incrementally increase the basis set, adding new basis functions to tackle states with high Bellman error.

There are many other possible extensions to this work. We have already pursued extensions to collaborative multiagent systems, where multiple agents act simultaneously to maximize the global reward (Guestrin *et al.*, 2001b), and factored POMDPs, where the full state is not observed directly, but indirectly through observation variables (Guestrin, Koller, & Parr, 2001c). However, there are other settings that remain to be explored. In particular, we hope to address the problem of learning a factored MDP and planning in a competitive multiagent system.

Additionally, in this paper we have tackled problems where the induced width of the cost network is sufficiently low or that possess sufficient context-specific structure to allow for the exact solution of our factored LPs. Unfortunately, some practical problems may have prohibitively large induced width. We plan to leverage on ideas from loopy belief propagation algorithms for approximate inference in Bayesian networks (Pearl, 1988; Yedidia, Freeman, & Weiss, 2001) to address this issue.

We believe that the methods described herein significantly further extend the efficiency, applicability and general usability of factored models and value functions for the control of practical dynamic systems.

## Acknowledgements

We are very grateful to Craig Boutilier, Dirk Ormoneit and Uri Lerner for many useful discussions, and to the anonymous reviewers for their detailed and thorough comments. We also would like to thank Jesse Hoey, Robert St-Aubin, Alan Hu, and Craig Boutilier for distributing their algorithm and for their very useful assistance in using Apricodd and in selecting its parameters. This work was supported by the DoD MURI program, administered by the Office of Naval Research under Grant N00014-00-1-0637, by Air Force contract F30602-00-2-0598 under DARPA's TASK program, and by the Sloan Foundation. The first author was also supported by a Siebel Scholarship.

## Appendix A. Proofs

### A.1 Proof of Lemma 3.3

There exists a setting to the weights — the all zero setting — that yields a bounded max-norm projection error $\beta_P$ for any policy ($\beta_P \leq R_{max}$). Our max-norm projection operator chooses the set of weights that minimizes the projection error $\beta^{(t)}$ for each policy $\pi^{(t)}$. Thus, the projection error $\beta^{(t)}$ must be at least as low as the one given by the zero weights $\beta_P$ (which is bounded). Thus, the error remains bounded for all iterations. □

### A.2 Proof of Theorem 3.5

First, we need to bound our approximation of $\mathcal{V}_{\pi^{(t)}}$:

$$\begin{aligned}\left\|\mathcal{V}_{\pi^{(t)}} - \mathrm{H}\mathbf{w}^{(t)}\right\|_\infty &\leq \left\|\mathcal{T}_{\pi^{(t)}}\mathrm{H}\mathbf{w}^{(t)} - \mathrm{H}\mathbf{w}^{(t)}\right\|_\infty + \left\|\mathcal{V}_{\pi^{(t)}} - \mathcal{T}_{\pi^{(t)}}\mathrm{H}\mathbf{w}^{(t)}\right\|_\infty \quad \text{; (triangle inequality;)}\\ &\leq \left\|\mathcal{T}_{\pi^{(t)}}\mathrm{H}\mathbf{w}^{(t)} - \mathrm{H}\mathbf{w}^{(t)}\right\|_\infty + \gamma\left\|\mathcal{V}_{\pi^{(t)}} - \mathrm{H}\mathbf{w}^{(t)}\right\|_\infty \quad \text{; (}\mathcal{T}_{\pi^{(t)}}\text{ is a contraction.)}\end{aligned}$$

Moving the second term to the right hand side and dividing through by $1-\gamma$, we obtain:

$$\left\|\mathcal{V}_{\pi^{(t)}} - \mathrm{H}\mathbf{w}^{(t)}\right\|_\infty \leq \frac{1}{1-\gamma}\left\|\mathcal{T}_{\pi^{(t)}}\mathrm{H}\mathbf{w}^{(t)} - \mathrm{H}\mathbf{w}^{(t)}\right\|_\infty = \frac{\beta^{(t)}}{1-\gamma}. \tag{24}$$

For the next part of the proof, we adapt a lemma of Bertsekas and Tsitsiklis (1996, Lemma 6.2, p.277) to fit into our framework. After some manipulation, this lemma can be reformulated as:

$$\|\mathcal{V}^* - \mathcal{V}_{\pi^{(t+1)}}\|_\infty \leq \gamma \|\mathcal{V}^* - \mathcal{V}_{\pi^{(t)}}\|_\infty + \frac{2\gamma}{1-\gamma} \left\|\mathcal{V}_{\pi^{(t)}} - \mathrm{H}\mathbf{w}^{(t)}\right\|_\infty . \tag{25}$$

The proof is concluded by substituting Equation (24) into Equation (25) and, finally, induction on $t$. □

### A.3 Proof of Theorem 4.4

First, note that the equality constraints represent a simple change of variable. Thus, we can rewrite Equation (12) in terms of these new LP variables $u^{f_i}_{\mathbf{z}_i}$ as:

$$\phi \geq \max_{\mathbf{x}} \sum_i u^{f_i}_{\mathbf{z}_i}, \tag{26}$$

where any assignment to the weights $\mathbf{w}$ implies an assignment for each $u^{f_i}_{\mathbf{z}_i}$. After this stage, we only have LP variables.

It remains to show that the factored LP construction is equivalent to the constraint in Equation (26). For a system with $n$ variables $\{X_1, \ldots, X_n\}$, we assume, without loss of generality, that variables are eliminated starting from $X_n$ down to $X_1$. We now prove the equivalence by induction on the number of variables.

The base case is $n = 0$, so that the functions $c_i(\mathbf{x})$ and $b(\mathbf{x})$ in Equation (12) all have empty scope. In this case, Equation (26) can be written as:

$$\phi \geq \sum_i u^{e_i}. \tag{27}$$

In this case, no transformation is done on the constraint, and equivalence is immediate.

Now, we assume the result holds for systems with $i-1$ variables and prove the equivalence for a system with $i$ variables. In such a system, the maximization can be decomposed into two terms: one with the factors that *do not* depend on $X_i$, which are irrelevant to the maximization over $X_i$, and another term with all the factors that depend on $X_i$. Using this decomposition, we can write Equation (26) as:

$$\begin{aligned} \phi &\geq \max_{x_1,\ldots,x_i} \sum_j u^{e_j}_{\mathbf{z}_j}; \\ &\geq \max_{x_1,\ldots,x_{i-1}} \left[ \sum_{l\,:\,x_i \notin \mathbf{z}_l} u^{e_l}_{\mathbf{z}_l} + \max_{x_i} \sum_{j\,:\,x_i \in \mathbf{z}_j} u^{e_j}_{\mathbf{z}_j} \right]. \end{aligned} \tag{28}$$

At this point we can define new LP variables $u^e_{\mathbf{z}}$ corresponding to the second term on the right hand side of the constraint. These new LP variables must satisfy the following constraint:

$$u^e_{\mathbf{z}} \geq \max_{x_i} \sum_{j=1}^{\ell} u^{e_j}_{(\mathbf{z},x_i)[\mathbf{Z}_j]}. \tag{29}$$

This new non-linear constraint is again represented in the factored LP construction by a set of equivalent linear constraints:

$$u^e_{\mathbf{z}} \geq \sum_{j=1}^{\ell} u^{e_j}_{(\mathbf{z},x_i)[\mathbf{Z}_j]}, \forall \mathbf{z}, x_i. \tag{30}$$

The equivalence between the non-linear constraint Equation (29) and the set of linear constraints in Equation (30) can be shown by considering binding constraints. For each new LP variable created $u^e_{\mathbf{z}}$, there are $|X_i|$ new constraints created, one for each value $x_i$ of $X_i$. For any assignment to the LP variables in the right hand side of the constraint in Equation (30), only one of these $|X_i|$ constraints is relevant. That is, one where $\sum_{j=1}^{\ell} u^{e_j}_{(\mathbf{z},x_i)[\mathbf{Z}_j]}$ is maximal, which corresponds to the maximum over $X_i$. Again, if for each value of $\mathbf{z}$ more than one assignment to $X_i$ achieves the maximum, then any of (and only) the constraints corresponding to those maximizing assignments could be binding. Thus, Equation (29) and Equation (30) are equivalent.

Substituting the new LP variables $u^e_{\mathbf{z}}$ into Equation (28), we get:

$$\phi \geq \max_{x_1,\ldots,x_{i-1}} \sum_{l\,:\,x_i \notin \mathbf{z}_l} u^{e_l}_{\mathbf{z}_l} + u^e_{\mathbf{z}},$$

which does not depend on $X_i$ anymore. Thus, it is equivalent to a system with $i-1$ variables, concluding the induction step and the proof. □

### A.4 Proof of Lemma 7.1

First note that at iteration $t+1$ the objective function $\phi^{(t+1)}$ of the max-norm projection LP is given by:

$$\phi^{(t+1)} = \left\| \mathrm{H}\mathbf{w}^{(t+1)} - \left( R_{\pi^{(t+1)}} + \gamma P_{\pi^{(t+1)}} \mathrm{H}\mathbf{w}^{(t+1)} \right) \right\|_\infty .$$

However, by convergence the value function estimates are equal for both iterations:

$$\mathbf{w}^{(t+1)} = \mathbf{w}^{(t)}.$$

So we have that:

$$\phi^{(t+1)} = \left\| \mathrm{H}\mathbf{w}^{(t)} - \left( R_{\pi^{(t+1)}} + \gamma P_{\pi^{(t+1)}} \mathrm{H}\mathbf{w}^{(t)} \right) \right\|_\infty .$$

In operator notation, this term is equivalent to:

$$\phi^{(t+1)} = \left\| \mathrm{H}\mathbf{w}^{(t)} - \mathcal{T}_{\pi^{(t+1)}} \mathrm{H}\mathbf{w}^{(t)} \right\|_\infty .$$

Note that, $\pi^{(t+1)} = \mathit{Greedy}(\mathrm{H}\mathbf{w}^{(t)})$ by definition. Thus, we have that:

$$\mathcal{T}_{\pi^{(t+1)}} \mathrm{H}\mathbf{w}^{(t)} = \mathcal{T}^* \mathrm{H}\mathbf{w}^{(t)}.$$

Finally, substituting into the previous expression, we obtain the result:

$$\phi^{(t+1)} = \left\| \mathrm{H}\mathbf{w}^{(t)} - \mathcal{T}^* \mathrm{H}\mathbf{w}^{(t)} \right\|_\infty . \quad \square$$

## References

Arnborg, S., Corneil, D. G., & Proskurowski, A. (1987). Complexity of finding embeddings in a K-tree. *SIAM Journal of Algebraic and Discrete Methods*, *8*(2), 277 – 284.

Becker, A., & Geiger, D. (2001). A sufficiently fast algorithm for finding close to optimal clique trees. *Artificial Intelligence*, *125*(1-2), 3–17.

Bellman, R., Kalaba, R., & Kotkin, B. (1963). Polynomial approximation – a new computational technique in dynamic programming. *Math. Comp.*, *17*(8), 155–161.

Bellman, R. E. (1957). *Dynamic Programming.* Princeton University Press, Princeton, New Jersey.

Bertele, U., & Brioschi, F. (1972). *Nonserial Dynamic Programming.* Academic Press, New York.

Bertsekas, D., & Tsitsiklis, J. (1996). *Neuro-Dynamic Programming.* Athena Scientific, Belmont, Massachusetts.

Boutilier, C., Dean, T., & Hanks, S. (1999). Decision theoretic planning: Structural assumptions and computational leverage. *Journal of Artificial Intelligence Research*, *11*, 1 – 94.

Boutilier, C., & Dearden, R. (1996). Approximating value trees in structured dynamic programming. In *Proc. ICML*, pp. 54–62.

Boutilier, C., Dearden, R., & Goldszmidt, M. (1995). Exploiting structure in policy construction. In *Proc. IJCAI*, pp. 1104–1111.

Boutilier, C., Dearden, R., & Goldszmidt, M. (2000). Stochastic dynamic programming with factored representations. *Artificial Intelligence*, *121*(1-2), 49–107.

Cheney, E. W. (1982). *Approximation Theory* (2nd edition). Chelsea Publishing Co., New York, NY.

de Farias, D., & Van Roy, B. (2001a). The linear programming approach to approximate dynamic programming. *Submitted to Operations Research.*

de Farias, D., & Van Roy, B. (2001b). On constraint sampling for the linear programming approach to approximate dynamic programming. *To appear in Mathematics of Operations Research.*

Dean, T., Kaelbling, L. P., Kirman, J., & Nicholson, A. (1993). Planning with deadlines in stochastic domains. In *Proceedings of the Eleventh National Conference on Artificial Intelligence (AAAI-93)*, pp. 574–579, Washington, D.C. AAAI Press.

Dean, T., & Kanazawa, K. (1989). A model for reasoning about persistence and causation. *Computational Intelligence*, *5*(3), 142–150.

Dean, T., & Givan, R. (1997). Model minimization in Markov decision processes. In *Proceedings of the Fourteenth National Conference on Artificial Intelligence (AAAI-97)*, pp. 106–111, Providence, Rhode Island, Oregon. AAAI Press.

Dearden, R., & Boutilier, C. (1997). Abstraction and approximate decision theoretic planning. *Artificial Intelligence*, *89*(1), 219–283.

Dechter, R. (1999). Bucket elimination: A unifying framework for reasoning. *Artificial Intelligence*, *113*(1–2), 41–85.

Gordon, G. (1995). Stable function approximation in dynamic programming. In *Proceedings of the Twelfth International Conference on Machine Learning*, pp. 261–268, Tahoe City, CA. Morgan Kaufmann.

Guestrin, C. E., Koller, D., & Parr, R. (2001a). Max-norm projections for factored MDPs. In *Proceedings of the Seventeenth International Joint Conference on Artificial Intelligence (IJCAI-01)*, pp. 673 – 680, Seattle, Washington. Morgan Kaufmann.

Guestrin, C. E., Koller, D., & Parr, R. (2001b). Multiagent planning with factored MDPs. In *14th Neural Information Processing Systems (NIPS-14)*, pp. 1523–1530, Vancouver, Canada.

Guestrin, C. E., Koller, D., & Parr, R. (2001c). Solving factored POMDPs with linear value functions. In *Seventeenth International Joint Conference on Artificial Intelligence (IJCAI-01) workshop on Planning under Uncertainty and Incomplete Information*, pp. 67 – 75, Seattle, Washington.

Guestrin, C. E., Venkataraman, S., & Koller, D. (2002). Context specific multiagent coordination and planning with factored MDPs. In *The Eighteenth National Conference on Artificial Intelligence (AAAI-2002)*, pp. 253–259, Edmonton, Canada.

Hoey, J., St-Aubin, R., Hu, A., & Boutilier, C. (1999). SPUDD: Stochastic planning using decision diagrams. In *Proceedings of the Fifteenth Conference on Uncertainty in Artificial Intelligence (UAI-99)*, pp. 279–288, Stockholm, Sweden. Morgan Kaufmann.

Hoey, J., St-Aubin, R., Hu, A., & Boutilier, C. (2002). Stochastic planning using decision diagrams – C implementation. http://www.cs.ubc.ca/spider/staubin/Spudd/.

Howard, R. A., & Matheson, J. E. (1984). Influence diagrams. In Howard, R. A., & Matheson, J. E. (Eds.), *Readings on the Principles and Applications of Decision Analysis*, pp. 721–762. Strategic Decisions Group, Menlo Park, California.

Keeney, R. L., & Raiffa, H. (1976). *Decisions with Multiple Objectives: Preferences and Value Tradeoffs*. Wiley, New York.

Kim, K.-E., & Dean, T. (2001). Solving factored Mdps using non-homogeneous partitioning. In *Proceedings of the Seventeenth International Joint Conference on Artificial Intelligence (IJCAI-01)*, pp. 683 – 689, Seattle, Washington. Morgan Kaufmann.

Kjaerulff, U. (1990). Triangulation of graphs – algorithms giving small total state space. Tech. rep. TR R 90-09, Department of Mathematics and Computer Science, Strandvejen, Aalborg, Denmark.

Koller, D., & Parr, R. (1999). Computing factored value functions for policies in structured MDPs. In *Proceedings of the Sixteenth International Joint Conference on Artificial Intelligence (IJCAI-99)*, pp. 1332 – 1339. Morgan Kaufmann.

Koller, D., & Parr, R. (2000). Policy iteration for factored MDPs. In *Proceedings of the Sixteenth Conference on Uncertainty in Artificial Intelligence (UAI-00)*, pp. 326 – 334, Stanford, California. Morgan Kaufmann.

Meuleau, N., Hauskrecht, M., Kim, K., Peshkin, L., Kaelbling, L., Dean, T., & Boutilier, C. (1998). Solving very large weakly-coupled Markov decision processes. In *Proceedings of the 15th National Conference on Artificial Intelligence*, pp. 165–172, Madison, WI.

Pearl, J. (1988). *Probabilistic Reasoning in Intelligent Systems: Networks of Plausible Inference*. Morgan Kaufmann, San Mateo, California.

Puterman, M. L. (1994). *Markov decision processes: Discrete stochastic dynamic programming*. Wiley, New York.

Reed, B. (1992). Finding approximate separators and computing tree-width quickly. In *24th Annual Symposium on Theory of Computing*, pp. 221–228. ACM.

Schuurmans, D., & Patrascu, R. (2001). Direct value-approximation for factored MDPs. In *Advances in Neural Information Processing Systems (NIPS-14)*, pp. 1579–1586, Vancouver, Canada.

Schweitzer, P., & Seidmann, A. (1985). Generalized polynomial approximations in Markovian decision processes. *Journal of Mathematical Analysis and Applications*, *110*, 568 – 582.

Simon, H. A. (1981). *The Sciences of the Artificial* (second edition). MIT Press, Cambridge, Massachusetts.

Singh, S., & Cohn, D. (1998). How to dynamically merge Markov decision processes. In Jordan, M. I., Kearns, M. J., & Solla, S. A. (Eds.), *Advances in Neural Information Processing Systems*, Vol. 10. The MIT Press.

St-Aubin, R., Hoey, J., & Boutilier, C. (2001). APRICODD: Approximate policy construction using decision diagrams. In *Advances in Neural Information Processing Systems 13: Proceedings of the 2000 Conference*, pp. 1089–1095, Denver, Colorado. MIT Press.

Stiefel, E. (1960). Note on Jordan elimination, linear programming and Tchebycheff approximation. *Numerische Mathematik*, *2*, 1 – 17.

Sutton, R. S. (1988). Learning to predict by the methods of temporal differences. *Machine Learning*, *3*, 9–44.

Tadepalli, P., & Ok, D. (1996). Scaling up average reward reinforcmeent learning by approximating the domain models and the value function. In *Proceedings of the Thirteenth International Conference on Machine Learning*, Bari, Italy. Morgan Kaufmann.

Tatman, J. A., & Shachter, R. D. (1990). Dynamic programming and influence diagrams. *IEEE Transactions on Systems, Man and Cybernetics*, *20*(2), 365–379.

Tsitsiklis, J. N., & Van Roy, B. (1996a). Feature-based methods for large scale dynamic programming. *Machine Learning*, *22*, 59–94.

Tsitsiklis, J. N., & Van Roy, B. (1996b). An analysis of temporal-difference learning with function approximation. Technical report LIDS-P-2322, Laboratory for Information and Decision Systems, Massachusetts Institute of Technology.

Van Roy, B. (1998). *Learning and Value Function Approximation in Complex Decision Processes*. Ph.D. thesis, Massachusetts Institute of Technology.

Williams, R. J., & Baird, L. C. I. (1993). Tight performance bounds on greedy policies based on imperfect value functions. Tech. rep., College of Computer Science, Northeastern University, Boston, Massachusetts.

Yedidia, J., Freeman, W., & Weiss, Y. (2001). Generalized belief propagation. In *Advances in Neural Information Processing Systems 13: Proceedings of the 2000 Conference*, pp. 689–695, Denver, Colorado. MIT Press.

Zhang, N., & Poole, D. (1999). On the role of context-specific independence in probabilistic reasoning. In *Proceedings of the Sixteenth International Joint Conference on Artificial Intelligence (IJCAI-99)*, pp. 1288–1293. Morgan Kaufmann.